\documentclass[10pt,twocolumn,letterpaper]{article}

\usepackage{iccv}
\usepackage{times}
\usepackage{epsfig}
\usepackage{graphicx}
\usepackage{amsmath}
\usepackage{amssymb}
\usepackage[accsupp]{axessibility}  % Improves PDF readability for those with disabilities.

\usepackage[pagebackref=true,breaklinks=true,letterpaper=true,colorlinks,bookmarks=false]{hyperref}

\iccvfinalcopy % *** Uncomment this line for the final submission

\def\iccvPaperID{12539} % *** Enter the ICCV Paper ID here

\ificcvfinal\fi

\usepackage{multirow}
\usepackage{xspace}

\newcommand{\Tokyo}{Tokyo~24/7\xspace}

\renewcommand{\paragraph}[1]{\vspace{.25\baselineskip}\noindent{\bf #1}\xspace}

\def\roxf{$\mathcal{R}$Oxf\xspace}
\def\rpar{$\mathcal{R}$Par\xspace}

\def\clahe{$\mathcal{C}$}
\def\diverse{$\mathcal{D}$\xspace}
\newcommand{\hedn}{HED\textsuperscript{N}}
\newcommand{\hednx}{\hedn\xspace}
\newcommand{\rcfn}{RCF\textsuperscript{N}}

\newcommand{\nddata}{\xspace{\color{red}\bf !}\xspace}

\newcommand{\veryshortarrow}[1][3pt]{\mathrel{%
   \hbox{\rule[\dimexpr\fontdimen22\textfont2-.2pt\relax]{#1}{.4pt}}%
   \mkern-4mu\hbox{\usefont{U}{lasy}{m}{n}\symbol{41}}}}

\makeatletter
\DeclareRobustCommand\onedot{\futurelet\@let@token\@onedot}
\def\@onedot{\ifx\@let@token.\else.\null\fi\xspace}

\def\eg{\emph{e.g}\onedot} 
\def\ie{\emph{i.e}\onedot}

\makeatother
\begin{document}

%%%%%%%%% TITLE
\title{\!\!Dark~Side~Augmentation:~Generating~Diverse~Night~Examples~for~Metric~Learning}

\author{Albert Mohwald \qquad Tomas Jenicek \qquad Ondřej Chum\\
VRG, Faculty of Electrical Engineering, Czech Technical University in Prague\\
% {\tt\small \{mohwaalb,jenicto2\}@fel.cvut.cz, chum@cmp.felk.cvut.cz}
{\tt\small mohwaalb@fel.cvut.cz} \qquad {\tt\small jenicto2@fel.cvut.cz} \qquad {\tt\small chum@cmp.felk.cvut.cz}
}

\maketitle
% Remove page # from the first page of camera-ready.
\ificcvfinal\thispagestyle{empty}\fi

%%%%%%%%% ABSTRACT
\begin{abstract}

Image retrieval methods based on CNN descriptors rely on metric learning from a large number of diverse examples of positive and negative image pairs. Domains, such as night-time images, with limited availability and variability of training data suffer from poor retrieval performance even with methods performing well on standard benchmarks. We propose to train a GAN-based synthetic-image generator, translating available day-time image examples into night images. Such a generator is used in metric learning as a form of augmentation, supplying training data to the scarce domain. 
Various types of generators are evaluated and analyzed. We contribute with a novel light-weight GAN architecture that enforces the consistency between the original and translated image through edge consistency. The proposed architecture also allows a simultaneous training of an edge detector that operates on both night and day images. 
To further increase the variability in the training examples and to maximize the generalization of the trained model, we propose a novel method of diverse anchor mining. 
% Instead of random selection of training examples for each epoch, pseudo-random sampling is used, preventing similar examples from the modes of training data distribution to be selected.
 
The proposed method improves over the state-of-the-art results on a standard Tokyo 24/7 day-night retrieval benchmark while preserving the performance on Oxford and Paris datasets. This is achieved without the need of training image pairs of matching day and night images. The source code is available at \href{https://github.com/mohwald/gandtr}{https://github.com/mohwald/gandtr}.
% We further demonstrate that edge consistency between the original and translated image is the only constraint that is necessary in the context of subsequent image retrieval. 

\end{abstract}

%%%%%%%%% BODY TEXT
\section{Introduction}
\label{sec:intro}

\begin{figure}[t]
   \newcommand{\outscale}{0.19}
\newcommand{\toplab}[1]{\parbox{\outscale\linewidth}{\centering \scriptsize #1}}
\newcommand{\subfloat}[1]{\parbox{\outscale\linewidth}{\includegraphics[width=\linewidth]{#1}}}
\newcommand{\hspacing}{\hspace{1.5em}}

  \centering
   \toplab{Original}  \toplab{CycleGAN}  \toplab{CyEDA}  \toplab{\rcfn GAN}  \toplab{\hedn GAN}\\
    \subfloat{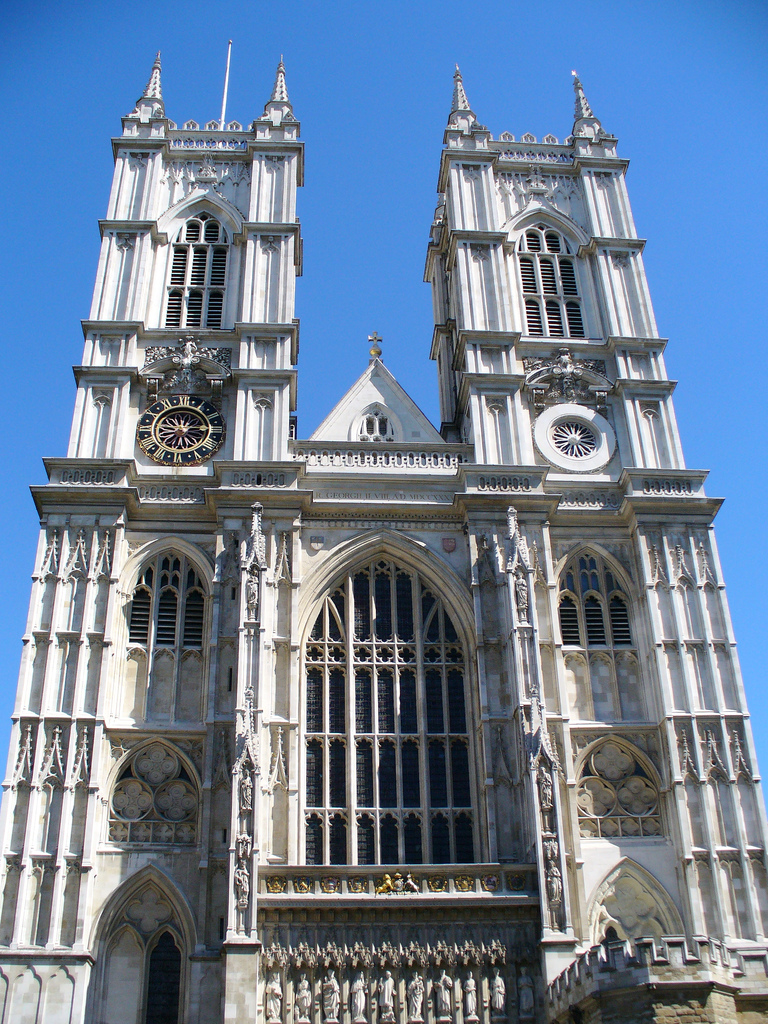} 
    \subfloat{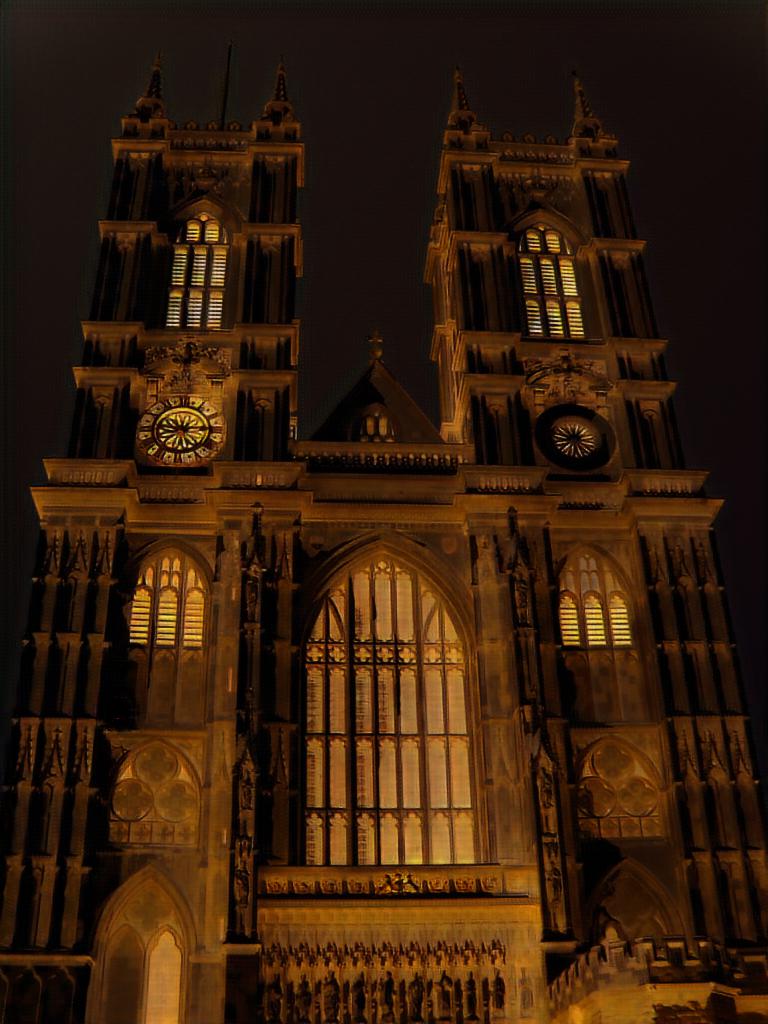} 
    \subfloat{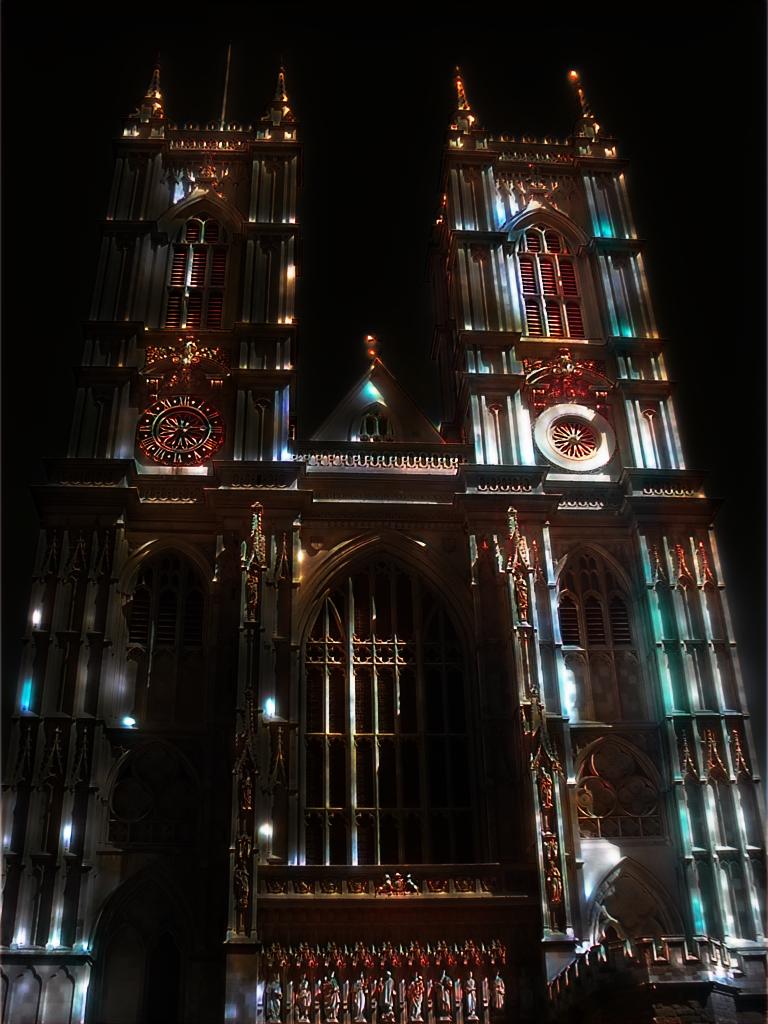} 
    \subfloat{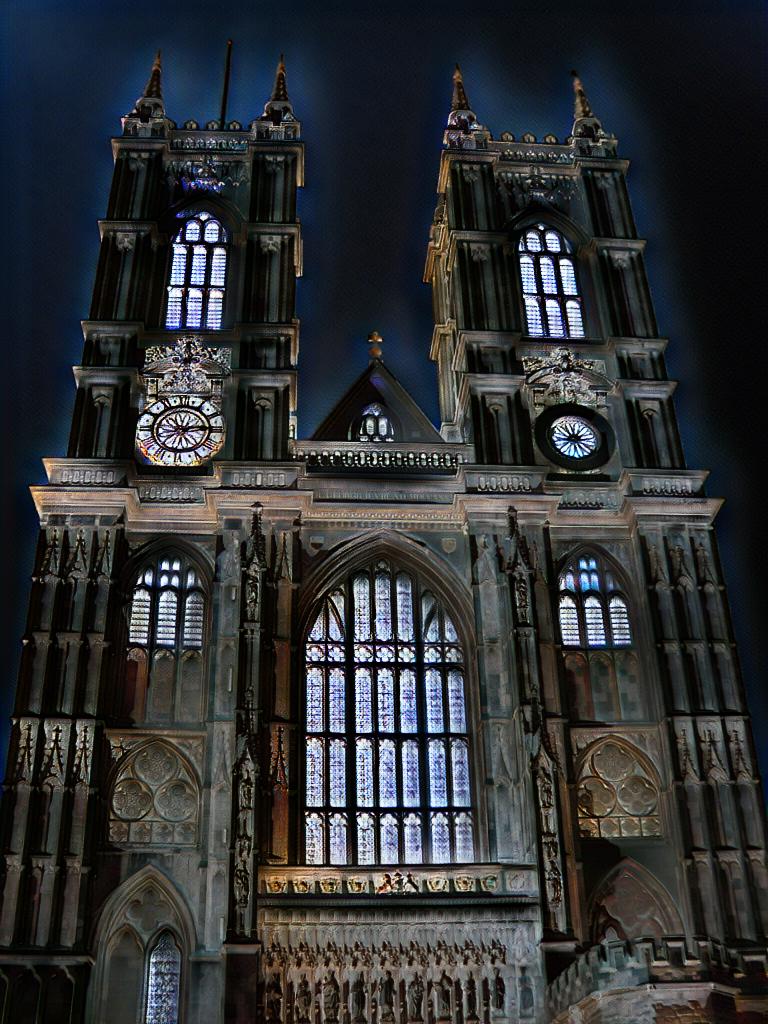} 
    \subfloat{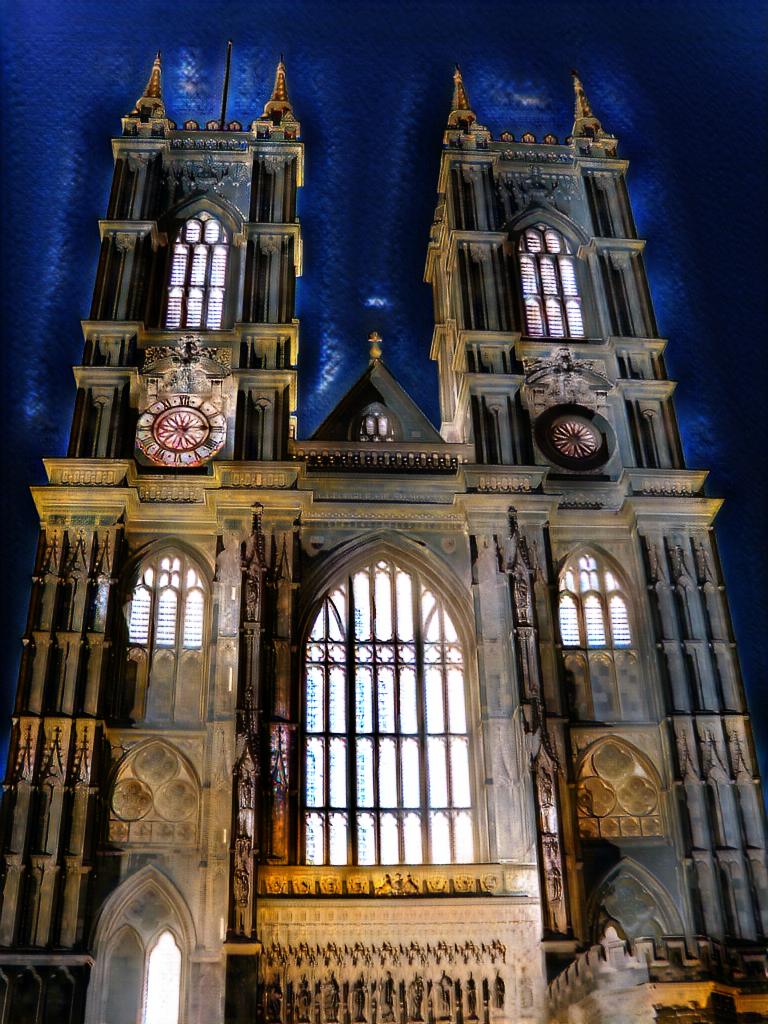} \\
    \subfloat{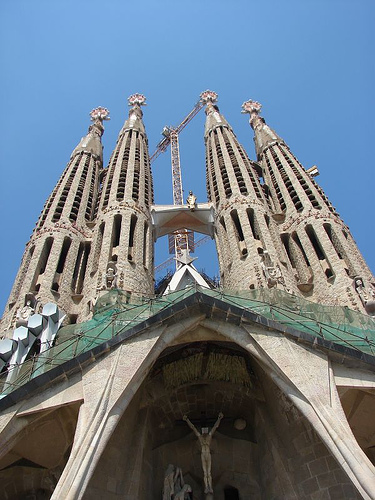} 
    \subfloat{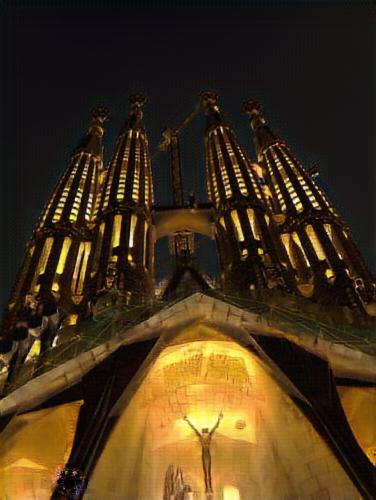} 
    \subfloat{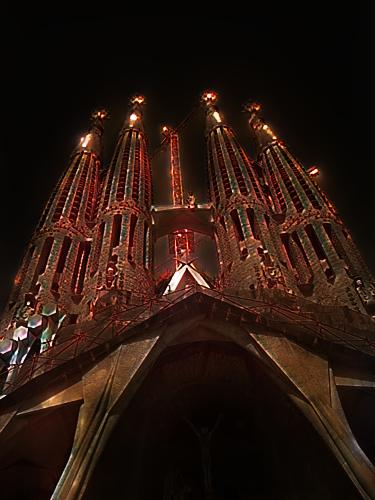} 
    \subfloat{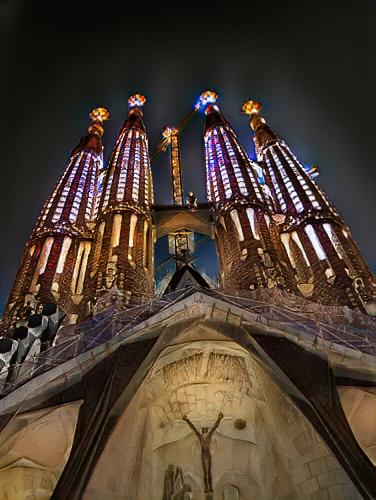} 
    \subfloat{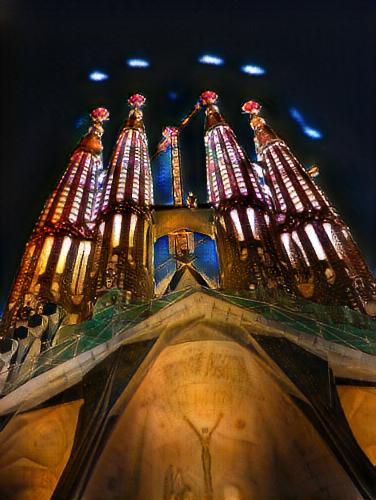} \\
    \subfloat{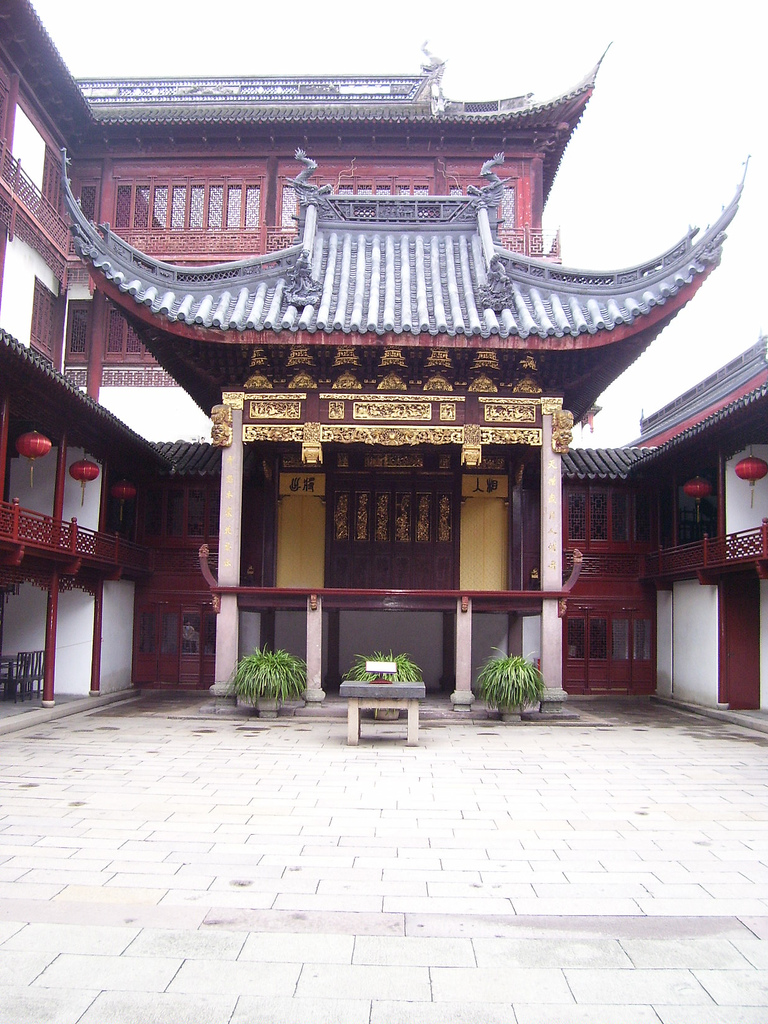} 
    \subfloat{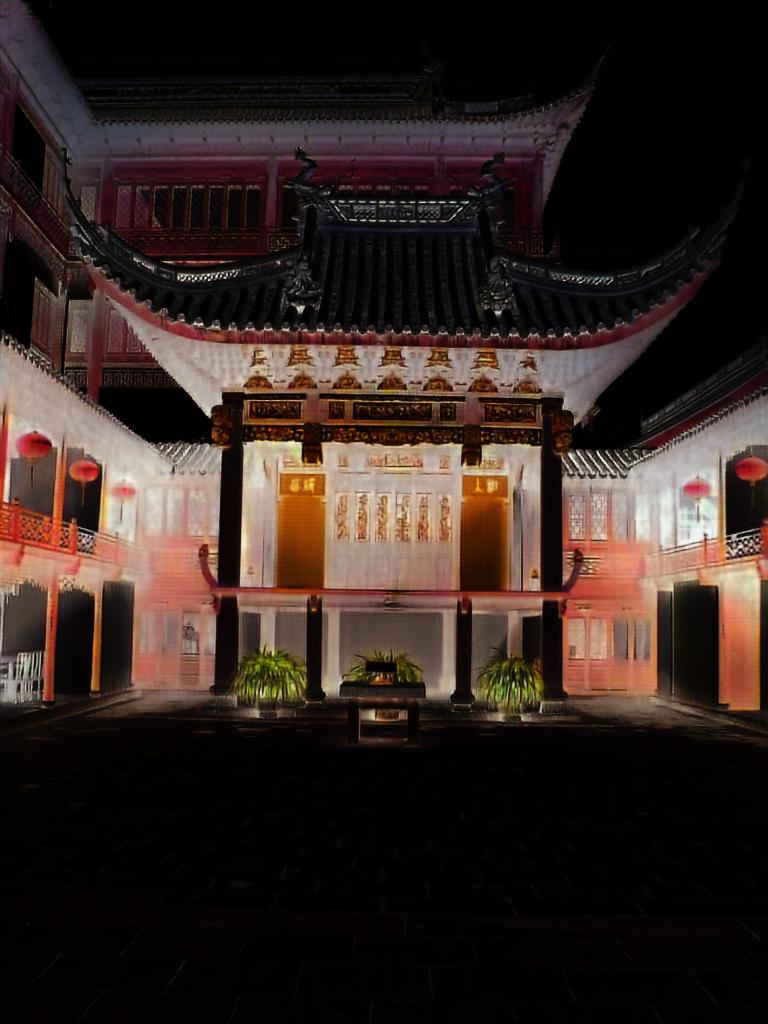} 
    \subfloat{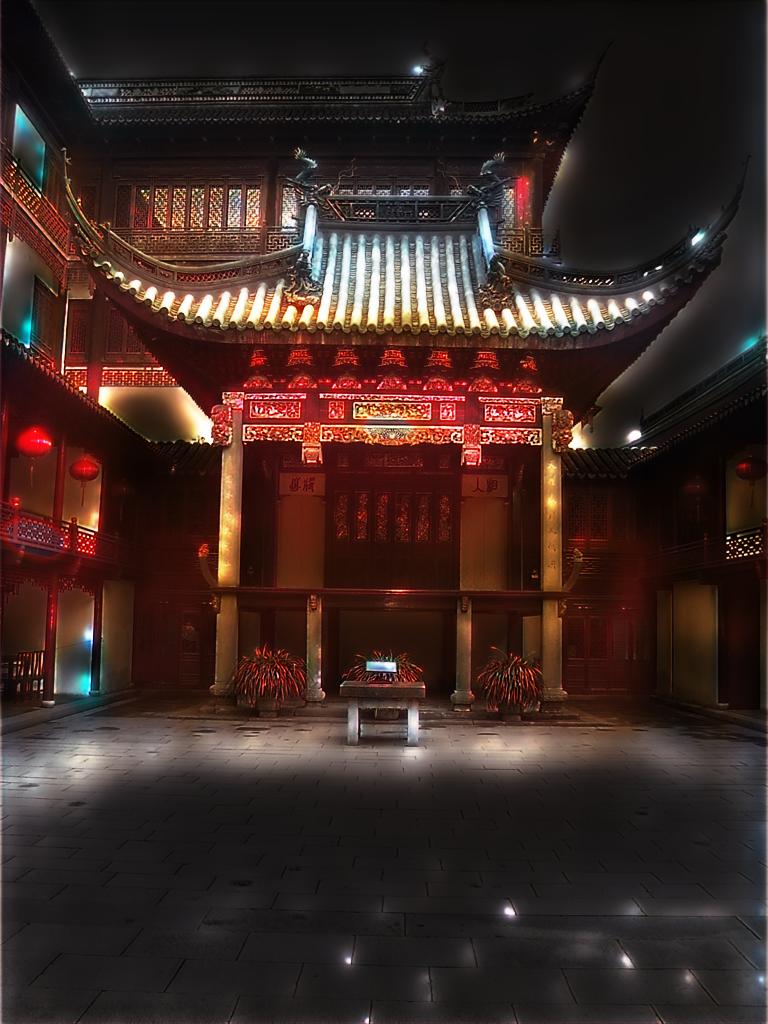} 
    \subfloat{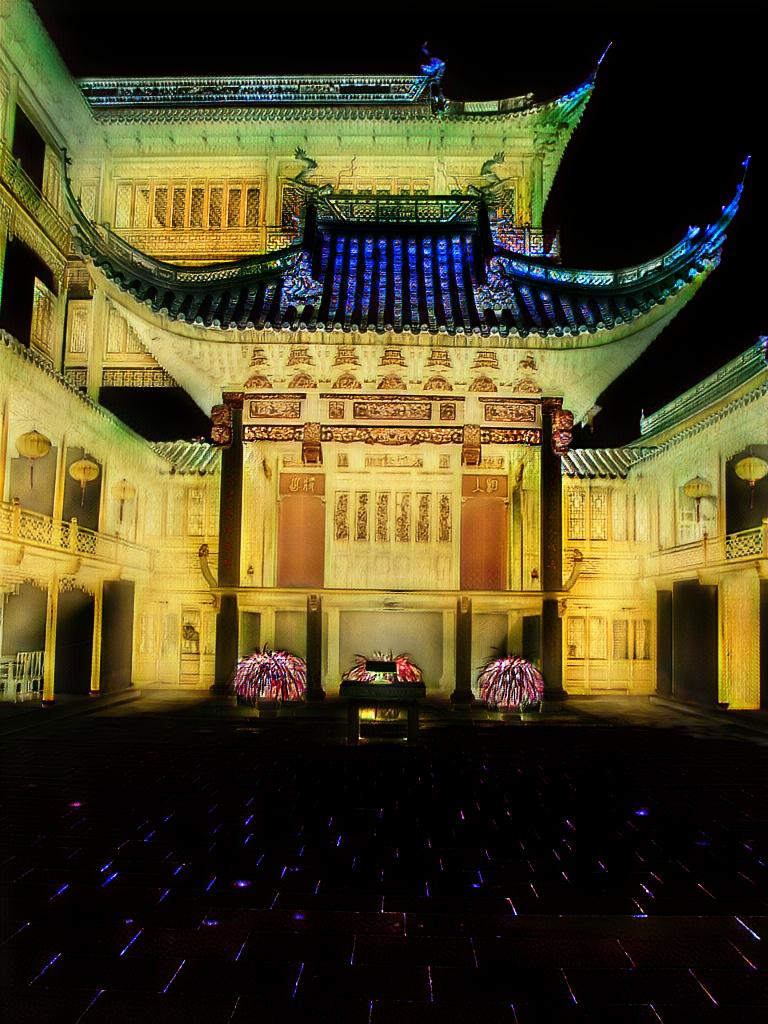} 
    \subfloat{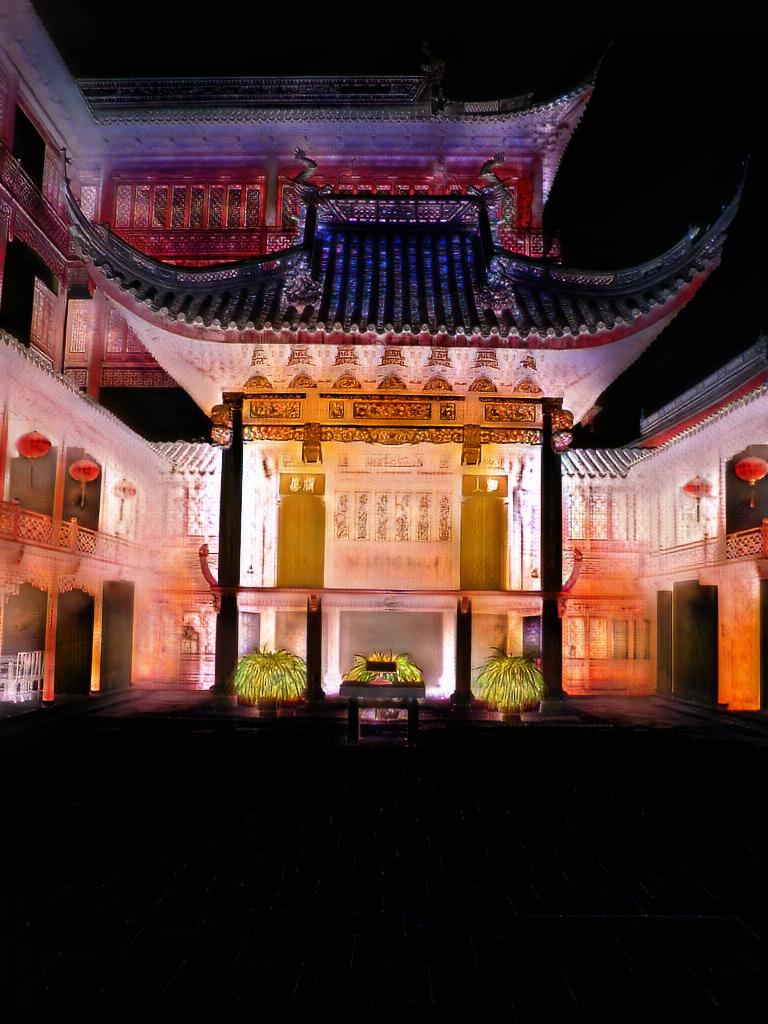} \\
    \vspace{1em}
    \caption{Examples of day-to-night translations with various generators. Each row consists of (left to right) the source image, and images translated by: CycleGAN, CyEDA, and the proposed RCF\textsuperscript{N}GAN and HED\textsuperscript{N}GAN. All models are trained on the \textit{SfM} dataset, except for CyEDA where a model pre-trained on BDD100k is used.}
   \label{fig:translated}
   
\newcommand{\motivationtwoscale}{0.23}
  \centering
    \begin{scriptsize}\hspace{0.2em} Input \hspace{5.5em} RCF  \hspace{5.5em} HED \hspace{5.5em} \hedn \end{scriptsize} \\[0.2em]
   \includegraphics[width=\motivationtwoscale\linewidth]{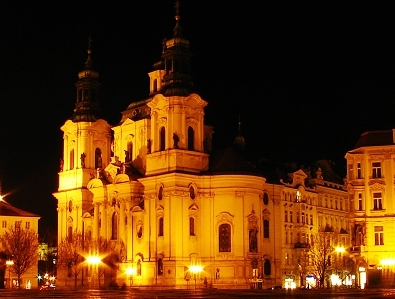} 
   \includegraphics[width=\motivationtwoscale\linewidth]{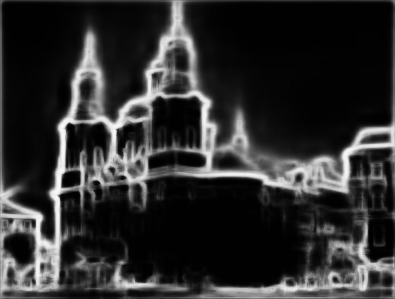}
   \includegraphics[width=\motivationtwoscale\linewidth]{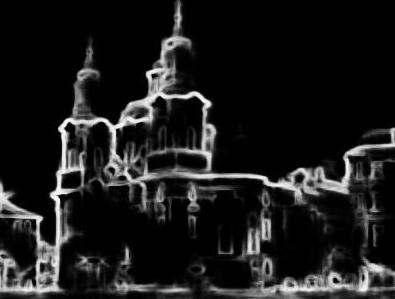} 
   \includegraphics[width=\motivationtwoscale\linewidth]{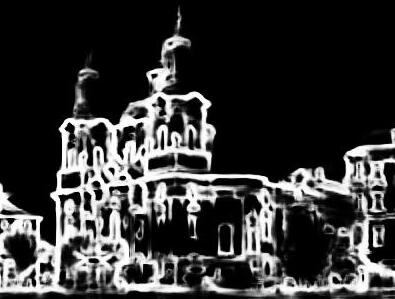}\\
    \includegraphics[width=\motivationtwoscale\linewidth]{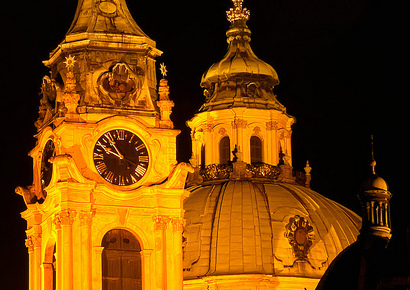} 
    \includegraphics[width=\motivationtwoscale\linewidth]{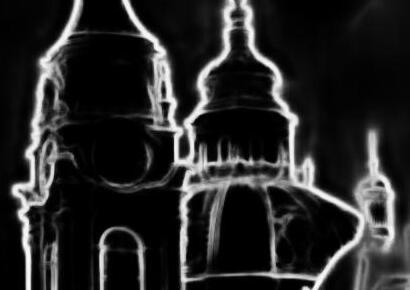} 
    \includegraphics[width=\motivationtwoscale\linewidth]{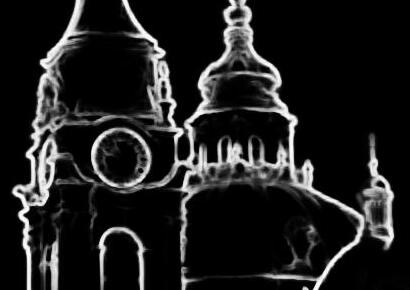} 
    \includegraphics[width=\motivationtwoscale\linewidth]{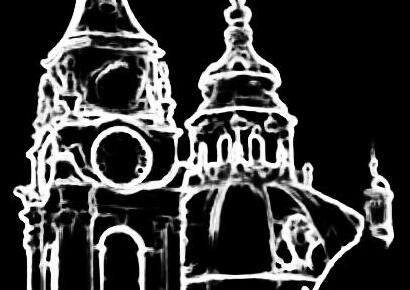}\\
    \includegraphics[width=\motivationtwoscale\linewidth]{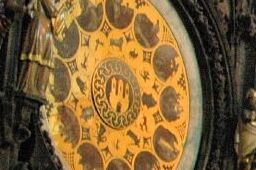}
    \includegraphics[width=\motivationtwoscale\linewidth]{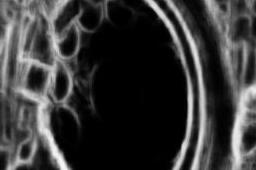} 
    \includegraphics[width=\motivationtwoscale\linewidth]{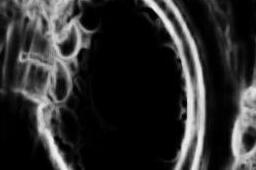} 
    \includegraphics[width=\motivationtwoscale\linewidth]{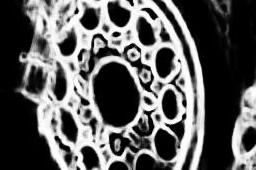}\\
    \caption{Comparison of edges extracted from real night images by RCF~\cite{liu2019richer}, HED~\cite{xie2015holistically}, and proposed \hedn, which is trained jointly with the edge-consistency based \hedn GAN generator. RCF and HED were trained mainly on day images and do not detect some edges in the night.}
   \label{fig:edgemaps}   
\mbox{}\vspace{-3em}   
\end{figure}

Large-scale instance-level image retrieval is commonly used \eg as a first step in
visual place recognition and visual localization. 
As other computer vision problems, image retrieval is dominated by methods based on deep learning models.
Fast and memory efficient approaches learn global image descriptors via metric learning. A large number and variety of corresponding image pairs is required to train well-performing global image descriptors.

One of the recent challenges in retrieval and visual localization is insensitivity to severe illumination changes, such as day and night~\cite{Sattler2018CVPR,Sattler2012BMVC}. 
Methods trained mostly on the day domain perform well on that domain, while having relatively poor results when night images are observed. The goal is an approach that performs well on all domains; improving results when night images are involved, while at the same time, the performance on the day domain should be preserved. The requirement of unharmed performance on the original (source) domain is, however, often neglected in the domain adaptation methods (\eg \cite{Lengyel-ICCV21}, as shown by our experiments).

To achieve illumination invariant image retrieval, alignment of the day and night domains by metric learning was proposed by~\cite{jenicek}. To achieve this, a large number of matching night to day image pairs is required. The acquisition of such pairs is a non trivial task, and suffers form significantly lower variability compared to day-to-day corresponding image pairs.  
Photo-sharing sites are a popular source of landmark image training data. A number of training and testing datasets were crawled from such sites, for example revisited Oxford and Paris~\cite{Radenovic-CVPR18}, Google landmarks v1~\cite{noh2017large}, \textit{SfM}~\cite{Radenovic-ECCV16} and \textit{SfM N/D}~\cite{jenicek}. For day images, these datasets exhibit sufficient visual variability to train image to descriptor mappings that generalize well to unseen scenes. In contrary, not only there are significantly less images taken during the night (\eg in the Aachen Day-Night dataset~\cite{Sattler2018CVPR}, there are 30 times more day images than night images), but their variability is also lower, as only parts of the scenes are visually interesting (\eg lit) during the night time. Therefore, only a small fraction of the scene reconstructed from the daytime images is photographed during night. This has been shown by day and night 3D reconstructions in~\cite{radenovic2016dusk}. 

As our {\em main contribution}, we propose to replace night training examples by synthetic images derived from day images by a generative adversarial network (GAN). In particular,
a standard \textit{SfM} dataset~\cite{radenovic2018fine} with high variability of homogeneous (mostly day to day) matching image pairs is used and one of the matching images is transformed by a GAN into a night image, see Figure~\ref{fig:translated}. This alleviates the necessity of obtaining night to day matching image pairs, and also significantly increases the variability of the training pairs. 
Even though night images are required to train the GAN, (i) a much lower number of night images is needed compared to performing the metric learning, (ii) these do not have to be paired with matching day counterpart. 
We compare various existing image translation methods that do not require pixel-aligned or visually related training data, in particular CycleGAN~\cite{cycleGAN}, DRIT~\cite{lee2020drit}, CUT~\cite{park2020contrastive}, and CyEDA~\cite{beh2022cyeda}. 

Inspired by the relative success of edge-based approaches to illumination invariance, as a {\em second contribution}, we propose a novel consistency enforcement through the edge consistency. Specifically, differentiable edge detector HED~\cite{xie2015holistically} is used to extract edges from the original and the translated image and their dissimilarity is penalized. The proposed method has a number of advantages: (i) it is an order of magnitude faster to train than CycleGAN while providing similar retrieval results, (ii) it provides insight into the importance and sufficiency of edges in night vision, (iii) it allows for simultaneous training of an edge detector (HED\textsuperscript{N}) that detects edges well in both day and night images. In this setup, HED~\cite{xie2015holistically} is compared to the more recent edge detector RCF~\cite{liu2019richer}.

Training data from automated 3D reconstructions are popular, as they are very clean and available without any human annotation. On the downside, the data distribution has strong modes that correspond to canonical views of popular landmarks.
As a {\em third contribution}, we propose to further increase the variability in the training examples, by a novel method of diverse anchor mining. Instead of a random selection of training examples for each epoch, pseudo-random importance sampling is used, preventing over-using training data from the modes of the training distribution (avoiding using multiple similar examples in the training).

We explore the idea of using diverse synthetic data for metric learning, compare different generators, including a newly proposed one, and study the contribution of individual aspects of the proposed method on global descriptors. We evaluate the performance of our models on image retrieval and visual localization datasets. The contribution is applicable to other methods as well, which we demonstrate by applying the proposed method to HOW~\cite{tolias2020learning}, a model that uses local descriptors for retrieval.

%-------------------------------------------------------------------------
\section{Related Work} \label{sec:related}

In this section, we first review relevant approaches to day-night image retrieval and discuss their relation to our work. 
Then, we summarize the image-to-image translation and how it is utilized in the data augmentation task, and finally we outline how other works are tackling data augmentation for visual recognition and image retrieval through day--night domain adaptation.

\paragraph{Day-time image retrieval.} In GeM~\cite{radenovic2018fine}, a CNN backbone produces global descriptors which can be compared by L2 norm to measure similarity between images. An alternative approach to use the CNN backbone to produce a set of local features was proposed in DELF~\cite{noh2017large}, HOW~\cite{tolias2020learning}, and FIRe~\cite{superfeatures}. In HOW, the last feature map is treated as a set of local features, from which the strongest features are used for image retrieval via ASMK~\cite{TAJ13}. FIRe follows the same pipeline, but adds a transformer-based head on top of the convolutional backbone.

The two paradigms can be combined, such as in DELG~\cite{cao2020unifying} where both global and local descriptors are utilized, each produced by a separate head. In DOLG~\cite{yang2021dolg}, a slightly different approach is adopted, using the interactions between global and local descriptors to produce a stronger global descriptor. This principle is applied to vision transformers~\cite{dosovitskiy2020image} in ViTGaL~\cite{phan2022patch} where a cross-attention between the CLS token and spatial token embeddings is performed at the end of the network. An alternative approach using a transformer-based backbone is taken by DToP~\cite{song2023boosting}, where local and global representations are produced by separate branches and the final representation is a concatenation of the two. The application of the proposed method to any of these methods is straightforward; we demonstrate its effectiveness on GeM~\cite{radenovic2018fine} and HOW~\cite{tolias2020learning}.

\paragraph{Real day-night training data.} The closest method of CNN-based illumination-invariant image retrieval is the work of~\cite{jenicek}. The alignment of the day and night domains consists of two steps -- a photometric normalization of input images, e.g. using CLAHE~\footnote{Contrast Limited Adaptive Histogram Equalization, see~\cite{szeliski2010computer} or~\cite{jenicek} for a detailed description}, and the exploitation of {\em matching pairs} of day and night images from a 3D reconstruction~\cite{Radenovic-ECCV16} for training.
In our work, we remove the requirement of obtaining matching night and day images. Further, we show that generating synthetic night images increases the variability in the training data which is reflected in a better retrieval performance.

\paragraph{Edges and color invariants.}
In EdgeMAC~\cite{Radenovic-ECCV18}, metric learning is performed on edge-detector responses (edgemaps), without the need of night training images. It was shown that edges are preserved in the presence of a significant change in illumination such as day and night images,  and even for images where colors and textures are corrupted. However, EdgeMAC was experimentally shown~\cite{Radenovic-ECCV18,jenicek} to perform poorly on standard datasets, as too much relevant information is lost in the process of turning images to edgemaps.

For edge detection, EdgeMAC exploits Dollár~\cite{dollar2013structured} which utilizes random decision forests. HED~\cite{xie2015holistically} proposed a CNN-based edge detector where detections from multiple intermediate feature maps are fused together, combining detections with different receptive fields. RCF~\cite{liu2019richer} follows the same architecture, but proposes to exploit detections from every intermediate feature map, which yields better results at the cost of an increased inference time.

Recently, a zero-shot day-night domain adaptation was proposed by~\cite{Lengyel-ICCV21}. A layer with trainable parameters performing color-invariant edge extraction is preceded to the backbone network. The method achieves interesting results without any night images during training. However, the retrieval results are significantly lower than results of the method proposed in this paper. Further, our experiments show that while improving the retrieval results in day-night settings, application of the method~\cite{Lengyel-ICCV21} substantially harms retrieval in the original day domain.

\paragraph{Image-to-image Translation}.
In image-to-image translation, popularized by the pix2pix~\cite{pix2pix} and CycleGAN~\cite{cycleGAN} models, an image in one visual domain is transformed into another domain, preserving the image content but modifying the image style.
In CycleGAN~\cite{cycleGAN}, the training images from the two domains do not need to be paired. There are two generators trained there, each translating in one direction between the two domains. This enables to constrain the input image and output image after two consecutive translations in the opposing directions to be identical.

In a more recent method DRIT~\cite{lee2020drit}, a similar architecture is presented, but with the focus on generating diverse output images. This is achieved by using a single latent space for all encoders and decoders of generators and splitting this latent space into a content and style space while introducing a new cross-cycle consistency loss.

An uni-directional image translation, training a generator for a single direction between the domains, is proposed in CUT~\cite{park2020contrastive}.
The consistency is enforced by patch-wise contrastive loss between corresponding patches from the original and target domain (positive pair) and other patches in the original domain (negative pairs). 

An edge-like consistency for CycleGAN has been proposed in CyEDA~\cite{beh2022cyeda}. Instead of comparing images in pixel-space, consistence of gradients (Sobel responses) is enforced. In our proposed method, sparse edge detections are used instead, and the edge detector is improved during the GAN training. Additionally, CyEDA generator introduces a blending mask mixing the input and output images as a form of skip-connection, which often fails in our setup, see Figure~\ref{fig:translated} top row for an example.

Another line of work focuses on architectures specifically for the task of object detection. In~\cite{lin2020multimodal}, cycle structure-consistency is enforced by measuring the difference between corresponding segmentation maps. Another approach is taken by~\cite{shen2019towards} and consequent work of~\cite{bhattacharjee2020dunit}, where the translation is performed directly for individual objects, implicitly ensuring a structure consistency in the context of object detection. Despite the impressive results in case of object-rich images, such an approach is not suitable for landmark recognition.

\paragraph{Data augmentation}.
To increase the number of training examples, various types of augmentations~\cite{deepLearning} were introduced in computer vision problems.
Image-to-image translation can be also used as a form of data augmentation for training, which is exercised by the proposed method.
Recently, \cite{mueller2019image} demonstrated that synthetic data helps in feature matching, visual localization, and image retrieval. 

In~\cite{arruda2019cross}, augmentation by image-to-image translation is applied to a car detection problem. Their training data contain car annotations in day images only, so they propose to translate day images into the night domain, exploiting the day annotations with the generated night data.
This is similar to the work of~\cite{lin2020gan}, with the difference that they design a custom GAN architecture in order to preserve objects in the images, aiding the consequent augmentation for a vehicle detection.
The same idea of translating day images with annotations into the night domain is used by~\cite{sun2019see} for image segmentation of vehicle-mounted camera images.

\paragraph{Image translation at the test time}.
Visual localization is approached by translating night images into day images during the inference in ToDayGAN~\cite{toDayGAN}.
Images from a camera mounted on a car are used, hence the variability in the images is relatively low. It is not clear whether such an approach would be applicable to general scenes\footnote{The official GitHub implementation of~\cite{toDayGAN} %https://github.com/AAnoosheh/ToDayGAN 
proclaims ``sensitivity to intrinsic camera characteristics''.}. 
We argue that generating a synthetic image commits to one of possible appearances, which, even if photo-realistic, is not guaranteed to be similar to the reality. Therefore, the applicability of image translation at the test time is limited. Instead, we attempt to learn an embedding that deals with all possible appearances, by using the image translation during the training.
This is supported by~\cite{sun2019see} for the task of image segmentation with vehicle-mounted camera, where translating night images into day during inference yields significantly worse results compared to translating day images into night as a training data augmentation.

%-------------------------------------------------------------------------

\begin{figure}[t]
  \parbox{\linewidth}{\footnotesize \bf \hedn GAN training} %\vspace{0.4em}
  \centering
  \includegraphics[width=0.99\linewidth]{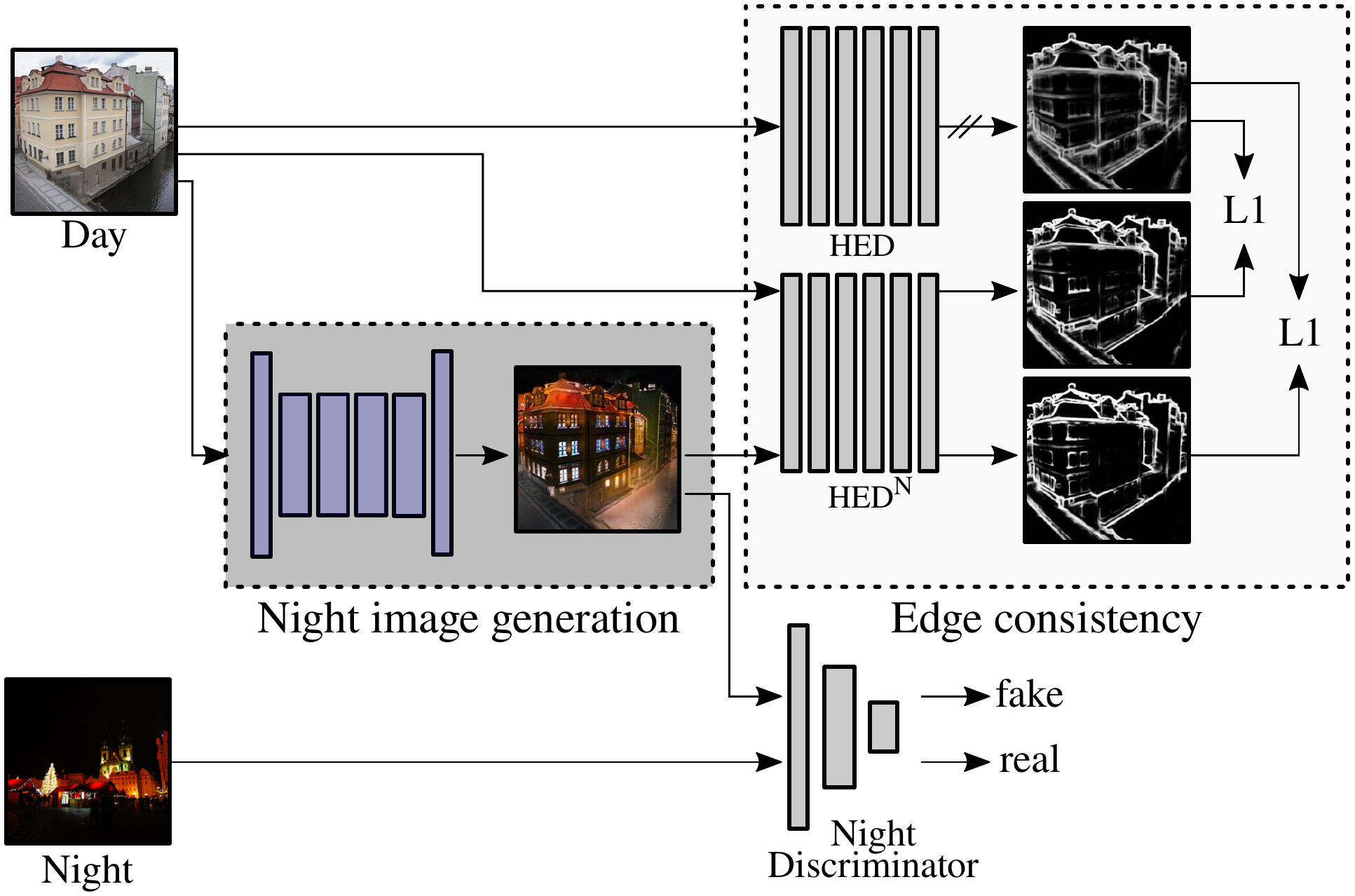}
    \caption{One training step with unpaired day and night images (left block) of our \hedn GAN architecture. The day $\veryshortarrow$ night generator translates the input day image (top left) into a fake night image (center), enforcing the edge consistency by L1 loss between HED and \hedn outputs (top right). The night discriminator predicts whether the generated night image (center) and the input night image (bottom left) are real or fake. \hedn edge detector (student) is trained by HED edge detector (teacher, not trained) to output night image edgemaps while preserving day image edgemaps.}
   \label{fig:hedgan-model}
\end{figure}
\begin{figure}[t]
  \parbox{\linewidth}{\footnotesize \bf Metric learning\vspace{4pt}}
  \centering
  \includegraphics[width=0.99\linewidth]{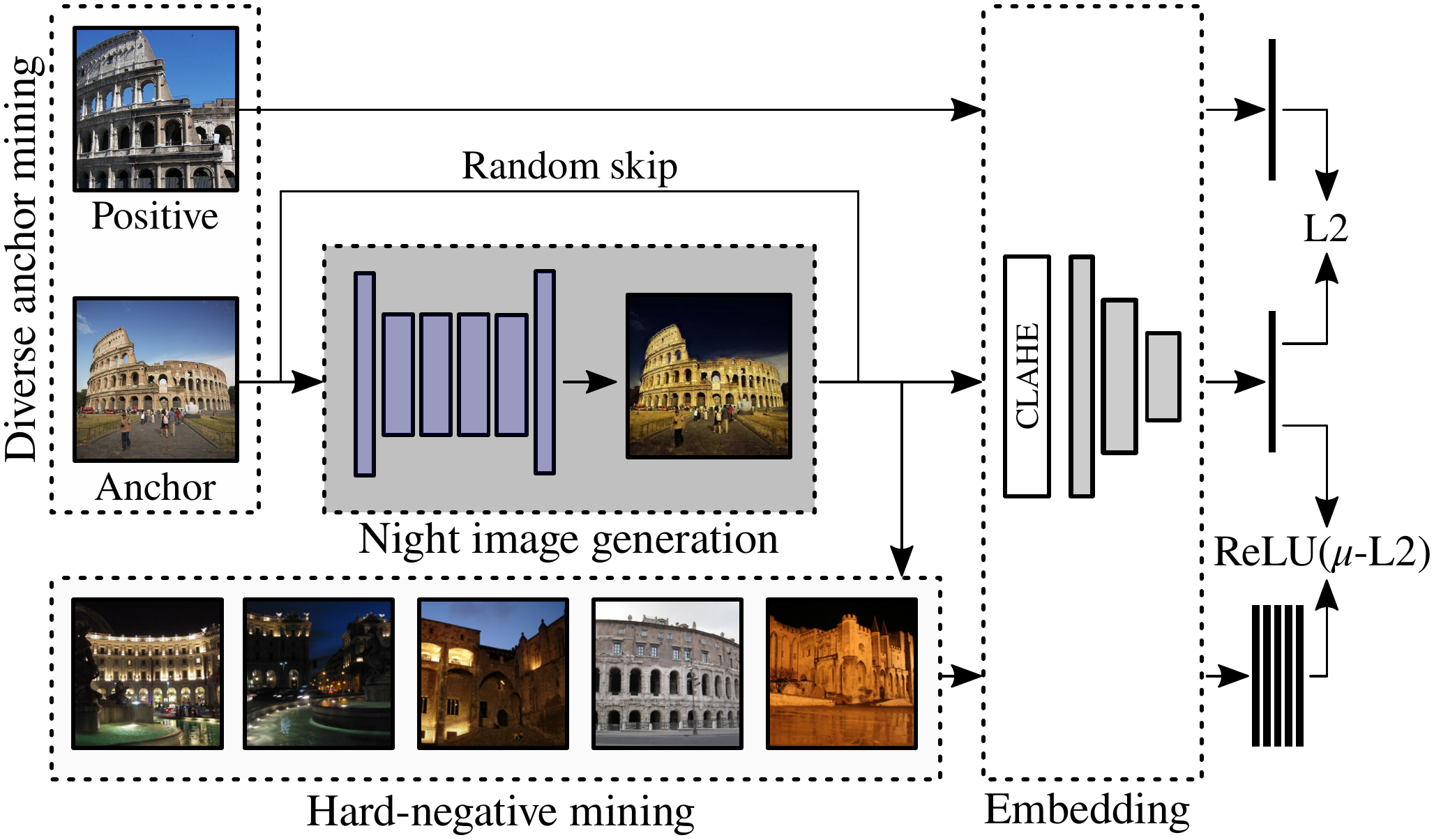}
    \caption[Finetuning embedding network]{Training data generation and photometric normalization during embedding network fine-tuning. A mined diverse anchor (center left day image) is randomly translated into a night image (gray block, trained generator from Figure~\ref{fig:hedgan-model}). The randomly translated image is used to mine a set of five negative images. The contrastive loss is applied on the global descriptors of the positive pair (L2) and of the negative pairs (ReLU($\mu$ - L2)).
%   An input image is translated with the augmentations network into night image by $0.15$ probability or directly skips the augmentation network otherwise. Then, the image descriptor $\textbf{f}$, representing the input image, is extracted from the input image by the embedding network.
  % In fine--tuning the embedding network, the truncated part is denoted with a green color, added parts are denoted with a red color, compared to the original model
   }
   \label{fig:augmentation-model}
\end{figure}

\section{Method}
\label{sec:method}

In the proposed method, a GAN generator is first trained on unpaired day-night images. The trained generator is exploited in the consequent metric learning to generate day-night training examples from labelled day pairs. For metric learning, a standard global descriptor constrastive learning framework is followed~\cite{jenicek} with three introduced changes: (a) diverse anchor images are mined, (b) anchor day images are translated to a night domain, and (c) negative mining is preformed after the optional translation step.

\subsection{\hedn GAN training}

It was shown previously~\cite{Radenovic-ECCV16,jenicek} that edges provide information that survives illumination changes between day and night. We propose a simple uni-directional image translator, named \hedn GAN, that attempts to generate images from the target domain and for which edgemaps are similar to edgemaps of the corresponding source images. For this purpose, a differentiable edge detector HED~\cite{xie2015holistically} is utilized.

The method is trained on examples from the day and night domain, where two unpaired images are randomly sampled in each iteration, one from each domain. The architecture consists of three models -- the generator, discriminator, and edge detector, as depicted in Figure~\ref{fig:hedgan-model}. In each iteration, a day image is translated via the generator into the night domain, resulting in a fake night image that is aligned with the real day image. An edge detection is performed on both the real day image (generator input) and the fake night image (generator output), and the resulting two edgemaps are compared pixel-wise, constraining the edges to be consistent between the two images. The discriminator is applied on the fake night image, training the generator adversarially, and ensuring that the images outputted are indistinguishable from the true night images. At the same time, the discriminator is trained on both sampled images -- the fake night image and a randomly sampled real night image.

HED detector~\cite{xie2015holistically} is trained mainly on day images which causes it to miss edges in night images, negatively influencing the generator performance in our setup. 
Therefore \hednx detector is trained jointly with the generator training, so that \hednx (trained student) and HED (frozen teacher) have similar responses on both real day and generated night images. HED with \hednx is also used to measure the similarity of the generator input and output, enforcing the edge cosistency between them. We also test a variant where the HED edge detector is not trained, named HEDGAN. The alternative edge detector RCF~\cite{liu2019richer} is evaluated in \rcfn GAN method.

\paragraph{Learning details}.
In all experiments, each input image is first randomly downscaled by a scale from $0.8$ to $1.0$ and then random-cropped to the final size of 256x256 px.
In a single iteration, two unpaired images from the two domains, day and night, are processed.
Each training epoch consists of 10000 iterations.

In the \hedn GAN architecture, ResNet generator~\cite{cycleGAN}, patchGAN discriminator~\cite{pix2pix}, and HED edge detector~\cite{xie2015holistically}\footnote{Re-implementation \href{https://github.com/sniklaus/pytorch-hed}{https://github.com/sniklaus/pytorch-hed} is used.} are exploited.
All three networks are trained simultaneously with batch size of 10. The training step of the generator and discriminator is the same as in CycleGAN~\cite{cycleGAN}. The generator and the discriminator use batch normalization and network weights are initalized following~\cite{he2015delving}, both of which led to an increased stability of the generator during training in our case. 
The two HED edge detectors, student and teacher, are initialized with the weights from~\cite{pytorch-hed}; the teacher weights are not updated during training while the student weights are optimized with Adam optimizer~\cite{kingma2014adam} with learning rate $10^{-6}$, $\beta_1 = 0.9$, $\beta_2 = 0.999$, and weight decay of $0.0002$. The edgemaps of the student and the teacher are compared pixel-wise via L1 distance -- in the optimization step of the edge detector, the L1 loss is applied on output values before the sigmoid function, while in the optimization step of the generator, it is after the sigmoid function.

For the other three tested architectures, CycleGAN~\cite{cycleGAN}, DRIT~\cite{lee2020drit}, and CUT~\cite{park2020contrastive}, their original implementations are used, in which the networks are trained for 100, 300, and 50 epochs, respectively. In all three architectures, the learning rate is linearly decayed to zero over the second half of epochs. 
For CyEDA~\cite{beh2022cyeda}, its pre-trained models on GTA~\cite{richter2017playing} and BDD100k~\cite{yu2020bdd100k} are evaluated as well as a variant trained on \textit{SfM}120k~\cite{radenovic2018fine} using its original training implementation.

\subsection{Metric learning}

The learning of global image descriptors is cast as metric learning via Siamese network, the architecture is visualized in Figure~\ref{fig:augmentation-model}. We follow the procedure used in~\cite{jenicek} including the same hyper-parameter settings.
First, to bring the appearance of images from different domains close together, a non-linear photometric CLAHE~\cite{szeliski2010computer} normalization. CLAHE is performed on a grid 8x8 with clip limit of~1.
The training is initialized with ImageNet pretrained network~\cite{russakovsky2015imagenet}, followed by fine-tuning on the \textit{SfM} dataset~\cite{Radenovic-ECCV16,jenicek}. 
For the embedding network architecture, VGG-16~\cite{simonyan2014vgg} or ResNet-101~\cite{he2016deep} backbone is used, followed by GeM pooling and L2 normalization, as described in~\cite{radenovic2018fine}. The network is trained for 40 epochs with 2000 iterations each epoch.

To show wide applicability of the proposed method, we also train a retrieval method based on aggregated local features -- HoW~\cite{tolias2020learning}. The network is trained in a metric learning framework with a contrastive loss on global descriptors; the procedure of~\cite{tolias2020learning} is followed for both training and inference.

\paragraph{Night image generation.}
A trained day$\rightarrow$night generator is used to generate day-night examples from day-only pairs. Before each epoch, 25\% of anchors are translated from day to night domain, while the corresponding positive and negative images are left unchanged. In the Retrieval-SfM N/D dataset~\cite{jenicek}, the same ratio of night anchors is used. 
The generator weights are not updated during training.

\paragraph{Hard negative mining.}
In each iteration of the fine-tuning, the embedding network takes 7-tuple of images -- one image is the \textit{anchor}, one image is \textit{positive}, and the remaining five images are \textit{negative} examples, following~\cite{Radenovic-ECCV16,jenicek}.
For each anchor, negative examples are mined from different 3D reconstructions, so that the distance between their descriptors and the anchor image descriptor is minimal. The negative mining takes place after the eventual anchor translation into the night domain. 

\paragraph{Diverse anchor mining}.
In prior approaches of metric learning on the \textit{SfM}~dataset~\cite{Radenovic-ECCV16,jenicek}, positive pairs for each epoch are selected from the set of all positive pairs in the dataset at random. 
Such a choice may lead to a repeated selection of similar anchor images (the same scene with a near-by view point) within an epoch.
To vary the training examples, we propose to iteratively select diverse anchor images from a random pool of anchor images.
The first anchor is selected at random. Before the next anchor is selected, the remainder of the pool is ordered by the minimal distance to already selected anchor images. The distance is measured as a Euclidean distance of image descriptors extracted by the network in its current state. 
New anchor image is selected at random from images between the 20th and 80th percentile of the pool ordering. Dropping 20\% of the closest images encourages the diversity in anchors, dropping  20\% of the most distant images prevents selecting images with outlying descriptors.
In the training, 2000 anchor images are selected from a pool of 10000 anchors.

%-------------------------------------------------------------------------
\section{Experiments}
\label{sec:experiments}
The proposed method is experimentally evaluated on various standard datasets, including day-night datasets as well as mostly homogeneous day datasets. The ablation study shows contributions of individual steps for different settings and implementation choices. The impact of the choice of the training dataset and the amount of night images used in training is discussed.

\subsection{Datasets}

% A) Retrieval

For training, we use the Retrieval-SfM dataset~\cite{radenovic2018fine}. 
Three standard datasets are used for evaluation of image retrieval: Tokyo~24/7~\cite{Torii-CVPR2015,jenicek}, revisited Oxford and Paris~\cite{Radenovic-CVPR18}.
The night-time performance is also assessed on visual localization on two datasets: Aachen Day-Night v1.1~\cite{zhang2021reference} and RobotCar Seasons~\cite{Sattler2018CVPR, Maddern2017IJRR}.

\paragraph{Retrieval-SfM} \cite{radenovic2018fine} (\textit{SfM}) contains 98045 images from reconstructed 3D models.
This dataset was used in the prior work of \cite{radenovic2018fine,jenicek} to finetune a CNN for image retrieval.
In this work, we also use the \textit{SfM} dataset for the metric learning and day$\rightarrow$night generator training.
For the generator training, the generated day-night annotations from~\cite{radenovic2016dusk} were used and images with dimensions less than 512 px were removed, resulting in 86385 day and 10039 night images.

\paragraph{Retrieval-SfM N/D} \cite{jenicek} (\textit{SfM-N/D}) was introduced by~\cite{jenicek} aiming to construct positive pairs with different lighting conditions. The \textit{SfM} dataset was enriched by additional day-night positive pairs by using the information in the original 3D models. Note that this dataset is included for comparison only, it is \emph{not} required to achieve results claimed as contribution of this paper (marked \nddata in methods).

\paragraph{Tokyo 24/7} \cite{Torii-CVPR2015} (\textit{Tokyo}) is a collection of 1125 smartphone-camera pictures capturing each of 375 scenes from 125 distinct locations in day, night and sunset light conditions.
In this work, we use \textit{Tokyo} to evaluate retrieval performance with the same evaluation protocol as proposed in~\cite{jenicek} -- each image is used as a query, images from the same scene but different lighting conditions are counted as positive, while images from different locations are considered as negative.

\paragraph{Oxford and Paris} \cite{Radenovic-CVPR18} (\roxf and \rpar) are standard image retrieval datasets (in their revisited version) and are used to evaluate retrieval performance on mostly homogeneous dataset with day-time images.

% B) Localization

\paragraph{Aachen Day-Night v1.1} \cite{Sattler2012BMVC, Sattler2018CVPR, zhang2021reference} (\textit{Aachen}) contains images of the old inner city of Aachen in Germany. The database consists of 6697 daytime images taken by handheld cameras and the query set contains 824 day-time and 191 night-time query images taken by three mobile phones. Performance is reported for the night-time queries only.

\paragraph{RobotCar Seasons} \cite{Maddern2017IJRR, Sattler2018CVPR} (\textit{RobotCar}) consists of images captured from 3 vehicle-mounted cameras: 26121 database images and 11934 query images taken under different conditions. Performance is reported for the night-time evaluation protocol of visuallocalization.net benchmark~\cite{Sattler2018CVPR} which contains images taken under \textit{night} (1314 images) and \textit{night-rain} (1320 images) conditions.

\setlength{\tabcolsep}{5.4pt}
\begin{table}[t] \centering
\newcommand{\firstsup}[1]{{\bf\color{red} #1}}
\newcommand{\secondsup}[1]{{\bf #1}}
\begin{tabular}{|l||r||r|r|r|}
    \multicolumn{5}{l}{VGG-16 backbone}\\\hline % 89e8
    Method & Avg & Tokyo & \roxf & \rpar \\\hline
    GeM~\cite{radenovic2018fine} & 69.9 & 79.4 & \secondsup{60.9} & 69.3 \\\hline
    GeM N/D~\cite{jenicek}~\nddata & 71.1 & 83.5 & 60.0 & 69.8 \\\hline
    CIConv~\cite{Lengyel-ICCV21} & - & 83.3 & - & - \\\hline
%    EdgeMAC~\cite{Radenovic-ECCV18} & 45.6 & 75.9 & 17.3 & 43.5 \\\hline
    CLAHE \clahe~\cite{jenicek} & 71.6 & 84.1 & 60.8 & 69.8 \\\hline
%	CLAHE \clahe \diverse (our \diverse) & 72.2 & 85.9 & 60.3 & 70.5 \\\hline
    CLAHE N/D \clahe~\cite{jenicek}~\nddata & 72.4 & 87.0 & 60.2 & 70.0 \\\hline
%	CLAHE N/D \clahe \diverse (our \diverse) \nddata & 73.0 & 87.7 & 60.8 & \secondsup{70.7} \\\hline
	\hedn GAN \clahe \diverse (ours) & \secondsup{73.4} & \secondsup{88.9} & \firstsup{61.1} & \secondsup{70.3} \\\hline
	CycleGAN \clahe \diverse (ours) & \firstsup{74.0} & \firstsup{90.2} & 60.7 & \firstsup{71.0} \\\hline
%    RCFGAN \clahe\diverse & 73.4 & 89.0 & 60.5 & 70.9 \\\hline
%
% ----------
%
    \multicolumn{5}{l}{}\\[-0.2em]
    \multicolumn{5}{l}{ResNet-101 backbone}\\\hline
    Method & Avg & Tokyo & \roxf & \rpar \\\hline
    GeM~\cite{radenovic2018fine} & 75.7 & 85.0 & 65.3 & \firstsup{76.7} \\\hline
    CIConv~\cite{Lengyel-ICCV21} & 75.0 & 88.3 & 62.0 & 74.7 \\\hline
%    GeM N/D \nddata & 77.0 & 88.6 & 65.7 & 76.8 \\\hline
%	CyEDA BDD100k \clahe\diverse & 77.8 & 90.3 & 65.7 & 77.3 \\\hline
%	CyEDA GTA \clahe\diverse & 78.2 & 91.2 & 65.8 & 77.6 \\\hline
%    HEDGAN \clahe \diverse (ours) & 78.0 & 91.2 & 66.5 & 76.4 \\\hline
	\hedn GAN \clahe \diverse (ours) & \firstsup{78.4} & \firstsup{92.2} & \secondsup{66.3} & \secondsup{76.6} \\\hline
	CycleGAN \clahe \diverse (ours) & \firstsup{78.4} & \secondsup{92.0} & \firstsup{66.8} & 76.4 \\\hline
%
% ----------
%
    \multicolumn{5}{l}{}\\[-0.2em]
    \multicolumn{5}{l}{HOW ResNet-18 backbone}\\\hline
    Method & Avg & Tokyo & \roxf & \rpar \\\hline
    HOW~\cite{tolias2020learning} & 80.8 & 87.8 & \secondsup{75.1} & 79.4 \\\hline
    HOW N/D~\nddata & \secondsup{82.0} & 89.2 & \firstsup{75.5} & \firstsup{81.4} \\\hline
    \hedn GAN \clahe \diverse (ours) & \secondsup{82.0} & \secondsup{91.6} & 74.6 & 79.7 \\\hline
    CycleGAN \clahe \diverse (ours) & \firstsup{82.4} & \firstsup{92.9} & 74.6 & \secondsup{79.8} \\\hline

\end{tabular}\\[3pt]
    \caption{Comparison in terms of mAP on \Tokyo, \roxf Medium and \rpar Medium datasets and their average. Methods marked by \nddata use paired day-night training data. The best score for each backbone architecture (in separate tables) is emphasized by red bold, second best by bold.}
\label{tab:results}
\end{table}

\setlength{\tabcolsep}{6.8pt}
\begin{table}[t] \centering
\newcommand{\firstsup}[1]{{\bf\color{red} #1}}
\newcommand{\secondsup}[1]{{\bf #1}}
\begin{tabular}{|l||r||r|r|r|}\hline
	Method & Avg & Tokyo & \roxf & \rpar \\\hline
    DOLG~\cite{yang2021dolg} & - & - & 81.5 & 91.0 \\\hline
    DOLG & 82.6 & 75.4 & 82.4 & 91.0 \\\hline
    ViTGaL~\cite{phan2022patch} & - & - & 82.4 & 91.4 \\\hline
    ViTGaL & 83.6 & 79.8 & 79.6 & 91.4 \\\hline
%	HOW~\cite{tolias2020learning} & 83.8 & 93.1 & 78.3 & 80.1 \\\hline
%	FIRe~\cite{superfeatures} ResNet-50 & \firstsup{86.4} & \firstsup{92.1} & 81.8 & 85.3 \\\hline
%	HOW HED\textsuperscript{N}GAN \clahe\diverse (temp \#s) & 82.1 & 94.4 & 73.9 & 77.9 \\\hline

\end{tabular}\\[3pt]
    \caption{Performance of the SoTA methods with publicly available code. DOLG~\cite{yang2021dolg} uses the convolutional backbone ResNet-101, ViTGaL~\cite{phan2022patch} uses transformer backbone XCiT-S24. All methods are trained on GLDv2 dataset~\cite{weyand2020google} which overlaps with test datasets \roxf and \rpar. For each method, we provide the results as reported in their paper (marked by the paper reference) and as reproduced by their publicly available code. Note the poor perfomance on the \Tokyo dataset.
    %Comparison in terms of mAP on \Tokyo, \roxf Medium and \rpar Medium datasets and their average.
    }
\label{tab:sota}
\end{table}

\subsection{Results}

We provide the results for embedding networks with the VGG-16 as well as ResNet-101 backbone. GAN is trained on the \textit{SfM} dataset (using unpaired day and night images), the embedding network is fine-tuned also on the \textit{SfM} dataset (using matching day images pairs), and the final retrieval performance is evaluated on \textit{Tokyo}, \roxf and \rpar datasets. Our method is trained on the GAN-augmented \textit{SfM} dataset, contains CLAHE normalization step, diverse anchor mining, and night examples are generated from day anchors with probability of 25\%. The full version of tables can be found in the Supplementary Material.

We compare with the baselines GeM~\cite{radenovic2018fine}, CLAHE~\cite{jenicek}, CLAHE N/D~\cite{jenicek} (day-night training pairs used) and CIConv~\cite{Lengyel-ICCV21}.
The baseline methods in Table~\ref{tab:results} are referred to by their original name and their reference. For GeM~\cite{radenovic2018fine}, the results of the publicly available github pytorch models\footnote{\href{https://github.com/filipradenovic/cnnimageretrieval-pytorch}{https://github.com/filipradenovic/cnnimageretrieval-pytorch}} are reported.
Methods proposed in this work are referred to as a combination of the retrieval training data (CycleGAN, \hedn GAN), whether CLAHE photometric normalization~\cite{jenicek} was used (marked \clahe), and whether diversity mining (contribution of this paper) was used (marked \diverse). For comparison, results by recent state-of-the-art (day time) retrieval methods are shown in Table~\ref{tab:sota}.

We trained all methods for 40 epochs, starting from ImageNet-pretrained backbones. This differs from~\cite{jenicek}, where the ImageNet-pretrained embedding network was trained for 20 epochs -- pre-fine-tuned for 10 epochs and then fine-tuned for 10 epochs in the final configuration\footnote{Please note that this difference, apart from providing a fair comparison, has preserved or slightly increased the performance of the re-trained baseline methods.}. Otherwise the setup from~\cite{jenicek} was followed precisely for all methods trained on \textit{SfM} and \textit{SfM-N/D} datasets (methods starting with GeM or CLAHE and GeM N/D \nddata or CLAHE N/D \clahe~\nddata respectively).

\setlength{\tabcolsep}{6.8pt}
\begin{table}[t] \centering
    \newcommand{\firstsup}[1]{{\bf #1}}
    \newcommand{\sla}{\hspace{-1pt}/\hspace{0.9pt}}
    \newcommand{\ns}[1]{{\footnotesize #1}}
    \setlength{\tabcolsep}{0.4em}
    {\small 
    
    \begin{tabular}{|l||r|r|r|} % 0f01
    \multicolumn{4}{l}{VGG-16 backbone}\\\hline % 89e8
        Method & \ns{top-1} & \ns{w/o global map} & \ns{with global map} \\\hline\hline
        GeM~\cite{radenovic2018fine} & \ns{0 \sla 0 \sla 16.2} & \ns{59.2 \sla 73.8 \sla 87.4} & \ns{62.3 \sla 76.4 \sla 91.1} \\\hline
        CycleGAN \clahe \diverse & \ns{0 \sla 0 \sla \firstsup{18.8}} & \ns{61.3 \sla 78.0 \sla \firstsup{90.6}} & \ns{62.8 \sla 78.5 \sla 92.1} \\\hline
        \hedn GAN \clahe \diverse & \ns{0 \sla 0 \sla 18.3} & \ns{\firstsup{62.3} \sla \firstsup{79.1} \sla 90.1 }& \ns{62.8 \sla 78.5 \sla 92.1} \\\hline
    \multicolumn{4}{l}{}\\[-0.2em]
    \multicolumn{4}{l}{ResNet-101 backbone}\\\hline
        Method & \ns{top-1} & \ns{w/o global map} & \ns{with global map} \\\hline\hline
        GeM~\cite{radenovic2018fine} & \ns{0 \sla 0.5 \sla 16.8} & \ns{60.7 \sla 75.4 \sla 88.0} & \ns{62.8 \sla 79.1 \sla 90.6} \\\hline
        CycleGAN \clahe \diverse & \ns{0 \sla 0.5 \sla \firstsup{20.4}} & \ns{63.4 \sla 77.0 \sla 92.1} & \ns{\firstsup{66.0} \sla 80.6 \sla \firstsup{95.8}} \\\hline
        \hedn GAN \clahe \diverse & \ns{0 \sla 0.5 \sla 19.4} & \ns{\firstsup{64.9} \sla \firstsup{79.6} \sla \firstsup{94.2}} & \ns{65.4 \sla 80.6 \sla 94.2} \\\hline
    %     AP-GeM & 0 / 0.0 / 17.3 & 63.9 / 79.6 / 94.8 & 68.6 / 83.8 / 96.9 \\\hline
    \end{tabular}}\\[1pt]
    \begin{center}\small
        Aachen v1.1 dataset - night
    \end{center}
    %     \caption{\look{Visual localization comparison on Aachen dataset. Only night conditions were evaluated. All scores are reported as the percentage of localized night query images within their respective accuracy thresholds (0.25m, 2°) / (0.5m, 5°) / (5m, 10°). In the first column, the pose is approximated only from the first retrieved image; in the second column, the pose is estimated from local SfM pipeline using only retrieved images; in the third column, the pose is estimated from global SfM 3D map.}}
    % \label{tab:aachen}
    % \end{table}
    
    % \begin{table}[t] \centering
    % \newcommand{\firstsup}[1]{{\bf #1}}
    % \newcommand{\secondsup}[1]{{\bf #1}}
    % \newcommand{\ns}[1]{{\footnotesize #1}}
    % \setlength{\tabcolsep}{0.4em}
    {\small
    \begin{tabular}{|l||r|r|r|} % 0f01
    \multicolumn{4}{l}{VGG-16 backbone}\\\hline % 89e8
        Method & \ns{top-1} & \ns{w/o global map} & \ns{with global map} \\\hline\hline
        GeM~\cite{radenovic2018fine}	& \ns{0 \sla 0.6 \sla 7.1} & \ns{5.9 \sla 11.4 \sla 16.1} & \ns{8.2 \sla 14.6 \sla 20.3} \\\hline
        CycleGAN \clahe \diverse & \ns{0.1 \sla 1.3 \sla \firstsup{17.4}} & \ns{\firstsup{12.2} \sla 21.9 \sla 29.2} & \ns{\firstsup{14.5} \sla 26.1 \sla 35.6} \\\hline
        \hedn GAN \clahe \diverse & \ns{0.1 \sla 1.3 \sla 16.3} & \ns{12.1 \sla \firstsup{22.7} \sla \firstsup{31.1}} & \ns{14.0 \sla \firstsup{27.1} \sla \firstsup{37.8}} \\\hline
    \multicolumn{4}{l}{}\\[-0.2em]
    \multicolumn{4}{l}{ResNet-101 backbone}\\\hline
        Method & \ns{top-1} & \ns{w/o global map} & \ns{with global map} \\\hline\hline
        GeM~\cite{radenovic2018fine} & \ns{0.1 \sla 0.9 \sla 10.1} & \ns{7.6 \sla 14.6 \sla 20.3} & \ns{10.3 \sla 18.1 \sla 25.2} \\\hline
        CycleGAN \clahe \diverse & \ns{\firstsup{0.5} \sla \firstsup{2.2} \sla \firstsup{22.6}} & \ns{\firstsup{14.1} \sla \firstsup{26.2} \sla \firstsup{36.0}} & \ns{\firstsup{16.3} \sla 29.2 \sla \firstsup{40.9}} \\\hline
        \hedn GAN \clahe \diverse & \ns{0.3 \sla 1.7 \sla 21.4} & \ns{13.4 \sla 24.7 \sla 34.3} & \ns{15.9 \sla \firstsup{29.7} \sla 40.0} \\\hline
    %     AP-GeM & 0.5 / 3.0 / 28.4 & 19.1 / 36.4 / 53.4 & 21.1 / 40.3 / 58.3 \\\hline
    \end{tabular}}\\[1pt]
    \begin{center}\small
        RobotCar Seasons dataset - all night
    \end{center}
        \caption{Visual localization evaluation on \textit{Aachen} and \textit{RobotCar} datasets -- only results for night conditions are reported. Scores in each table cell correspond to the percentage of localized query images within their respective accuracy thresholds: (0.25m, 2°) / (0.5m, 5°) / (5m, 10°). The columns correspond to three paradigms from Kapture pipeline~\cite{humenberger2022investigating}: pose approximation from top-1 retrieved image, and pose estimation from top 20 images without global map and with global map. The best score for each dataset and backbone architecture (in separate tables) is emphasized by by bold.}
    \label{tab:localization}
\end{table}

\paragraph{Beyond global descriptors.}
The proposed method of generating night data was also applied to HoW~\cite{tolias2020learning}, a retrieval method based on aggregated local features. The results are summerized in the bottom of Table~\ref{tab:results}, showing improvement over the baseline and similar or better results compared to a version of the HOW network trained on \textit{SfM-N/D}.

\paragraph{Beyond image retrieval.}
Visual localization is exploited to measure the night-time performance on a related task. We utilize the Kapture pipeline~\cite{humenberger2022investigating}, in particular its benchmark for image retrieval in the context of visual localization. The percentage of images localized within each of the three thresholds for three paradigms is reported, following visuallocalization.net benchmark~\cite{Sattler2018CVPR}. Three localization paradigms from Kapture pipeline~\cite{humenberger2022investigating} are followed - Paradigm 1: pose approximation by returning the pose of the top-1 retrieved image. Paradigm 2a\,\&\,2b: pose estimation from top 20 retrieved images without (2a) and with (2b) a global map respectively. Three methods from Table~\ref{tab:results} were evaluated: baseline GeM~\cite{radenovic2018fine}, and our CycleGAN and \hedn GAN models. The results in Table~\ref{tab:localization} show a consistent performance improvement over the baseline across all setups. It should be noted that the results are not comparable to SotA methods trained for visual localization.

\paragraph{Night edge detection.}
The qualitative comparison of RCF, HED, and \hednx detectors is shown in Figure~\ref{fig:edgemaps}. To provide quantitative evaluation, we train EdgeMAC~\cite{Radenovic-ECCV18} edge embedding with HED, \hedn, and \rcfn detectors, increasing the image size to 362 and not binarizing the edgemaps to improve performance. The results are summarized in Table~\ref{tab:edge}. The embeddings based on the proposed \hednx outperform those based on both \rcfn and HED when used alone or in an ensemble with an image embedding (\eg GeM). Similarly to~\cite{jenicek}, ensembles of edge and image embeddings obtain better performance at the cost of double dimensionality. The best performance is obtained for an ensemble of our \hednx edge embedding and \hedn GAN \clahe \diverse or CycleGAN \clahe \diverse image embedding.

\paragraph{Discussion.}
Results in Table~\ref{tab:results} show that using GAN training data is superior even to using the paired day-night images from the \textit{SfM-N/D} dataset (CLAHE N/D \clahe~\cite{jenicek}). For CIConv~\cite{Lengyel-ICCV21} with the ResNet-101 backbone, we have used the publicly available model trained by the authors and evaluated it on Oxford and Paris benchmarks.\footnote{For VGG-16 backbone, there is no publicly available model, therefore only results on \textit{Tokyo} dataset published in~\cite{Lengyel-ICCV21} are reported}
We observe drop in both these benchmarks compared to GeM~\cite{Radenovic-ECCV16}. Our method outperforms~\cite{Lengyel-ICCV21} on both backbone architectures; in fact, our VGG-16 model ``\hedn GAN \clahe \diverse'' outperforms CIConv~\cite{Lengyel-ICCV21} ResNet-101 model on the \textit{Tokyo} dataset, despite the weaker backbone architecture.

Further, an alternative approach~\cite{toDayGAN} of using the generator in inference rather than during training was evaluated. On \textit{Tokyo} dataset, all images with the night class label (\ie using oracle, as this label is not exploited in any other method) were translated to the day domain prior to retrieval. Two models, the one provided by~\cite{toDayGAN} and our generator, were tested. However, in both cases the performance was substantially worse than the retrieval baseline, specifically 39 and 52 mAP for \cite{toDayGAN} and our generator respectively.

\setlength{\tabcolsep}{5.4pt}
\begin{table}[t] \centering
\newcommand{\doubledim}{{\color{red}\bf \textdaggerdbl}\xspace}
\newcommand{\firstsup}[1]{{\bf #1}}
\newcommand{\secondsup}[1]{#1}
\begin{tabular}{|l||r||r|r|r|} \hline
	Method & Avg & Tokyo & \roxf & \rpar \\\hline\hline
    EdgeMAC~\cite{Radenovic-ECCV18} & 45.6 & 75.9 & 17.3 & 43.5 \\\hline
    HEDMAC & 56.8 & 79.5 & 38.3 & 52.5 \\\hline
    \hedn MAC & \firstsup{59.2} & 81.9 & \firstsup{38.4} & \firstsup{57.2} \\\hline
    % RCFMAC & 47.4 & 68.4 & 28.9 & 44.8 \\\hline
    \rcfn MAC & 58.5 & \firstsup{88.9} & 35.1 & 51.4 \\\hline\hline
    HEDMAC+GeM \doubledim & 72.0 & 84.8 & 60.9 & 70.3 \\\hline
    \hedn MAC+GeM \doubledim & 72.6 & 85.7 & \firstsup{61.1} & \secondsup{70.9} \\\hline
%     HEDMAC+GAN \doubledim & \secondsup{73.8} & \secondsup{90.9} & 60.1 & \secondsup{70.5} \\\hline
    \hedn MAC+\textsuperscript{N}GAN \doubledim & \secondsup{74.4} & \secondsup{91.4} & \secondsup{60.6} & \secondsup{71.3} \\\hline
    \hedn MAC+GAN \doubledim & \firstsup{74.7} & \firstsup{91.8} & \secondsup{60.4} & \firstsup{71.9} \\\hline
    
\end{tabular}\\[3pt]
    \caption{The effect of our trained \hedn~detector (from \hedn GAN) on the EdgeMAC~\cite{Radenovic-ECCV18} method. HEDMAC and \hedn MAC is a variant of EdgeMAC method with the HED~\cite{xie2015holistically} edge detector with either original or our weights, respectively. \rcfn MAC is a variant of EdgeMAC with the RCF~\cite{liu2019richer} edge detector with our weights. In the bottom block, ensembles of EdgeMAC variants with chosen methods from Table~\ref{tab:results} are reported. GeM is from~\cite{radenovic2018fine}, \textsuperscript{N}GAN corresponds to \hedn GAN \clahe \diverse, and GAN to CycleGAN \clahe \diverse, all from Table~\ref{tab:results}. Ensembles have double the dimensionality (1024) and are marked with \doubledim. The best score for each dataset in each block is in bold.}
\label{tab:edge}
\end{table}

\setlength{\tabcolsep}{6.8pt}
\begin{table}[t] \centering
\newcommand{\firstsup}[1]{{\bf\color{red} #1}}
\newcommand{\secondsup}[1]{{\bf #1}}
\begin{tabular}{|l||r||r|r|r|} \hline % 0f01
	Method & Avg & Tokyo & \roxf & \rpar \\\hline\hline
%
%	GeM~\cite{radenovic2018fine} & 70.0 & 80.4 & 59.9 & 69.8 \\\hline
%	GeM~\cite{radenovic2018fine} \diverse & 70.3 & 81.2 & 59.8 & 69.9 \\\hline
	CLAHE \clahe & 71.9 & 85.4 & 60.0 & 70.1 \\\hline
	CLAHE \clahe \diverse & 72.2 & 85.9 & 60.3 & 70.5 \\\hline
%
%	GeM N/D \nddata & 71.5 & 84.0 & 60.4 & 70.0 \\\hline
%	GeM N/D \diverse & 71.5 & 84.1 & 60.3 & 70.1 \\\hline
	CLAHE N/D \clahe~\nddata & 72.5 & 87.5 & 59.9 & 70.1 \\\hline
	CLAHE N/D \clahe \diverse \nddata & {\bf 73.0} & {\bf 87.7} & {\bf 60.8} & {\bf 70.7} \\\hline\hline
	\hedn GAN & 72.7 & 88.0 & 60.2 & 70.0 \\\hline
%	\hedn GAN \diverse & 73.0 & 88.7 & 60.2 & 70.1 \\\hline
	\hedn GAN \clahe & 73.2 & 88.7 & 60.5 & 70.4 \\\hline
	\hedn GAN \clahe \diverse & {\bf 73.4} & {\bf 88.8} & {\bf 60.7} & {\bf 70.6} \\\hline
%
%	CycleGAN & 73.0 & 88.8 & 59.6 & 70.5 \\\hline
%	CycleGAN \diverse & 73.3 & 89.1 & 59.9 & 70.7 \\\hline
%	CycleGAN \clahe & 73.6 & 89.6 & 60.5 & 70.9 \\\hline
%	CycleGAN \clahe \diverse & {\bf 74.0} & {\bf 90.2} & 60.7 & {\bf 71.0} \\\hline
\end{tabular}\\[3pt]
    \caption{The effect of diverse anchors (\diverse). Methods CLAHE \clahe~\cite{jenicek} and CLAHE N/D \clahe~\cite{jenicek} from Table~\ref{tab:results} are reported in the top block. Please note that we re-train the models for this ablation, so we obtain a slightly higher performance. In the bottom block, the effect of CLAHE (\clahe) and diverse anchors (\diverse) is reported on the proposed method \hedn GAN. The best score for each dataset in each block is in bold.}
\label{tab:diverse}
\end{table}

\setlength{\tabcolsep}{4pt}

\begin{table}[t] \centering
\begin{tabular}{|l||r|r|r|r|r|r|} \hline % 71f8
    \multirow{2}{*}{Method} & \multicolumn{2}{c|}{\tiny \{day, sunset\}} & \multicolumn{2}{c|}{\tiny \{sunset, night\}} & \multicolumn{2}{c|}{\tiny \{day, night\}}\\
	& D$\veryshortarrow$S & S$\veryshortarrow$D & \hspace{0.1em} S$\veryshortarrow$N & N$\veryshortarrow$S & D$\veryshortarrow$N & N$\veryshortarrow$D \\\hline\hline
	CLAHE \clahe & 97.7 & 98.2 & 80.1 & 81.3 & 70.9 & 76.1 \\\hline
%	CLAHE \clahe \diverse & 97.5 & 98.4 & 80.2 & 81.3 & 72.4 & 77.0 \\\hline\hline
%
	CLAHE N/D \clahe~\nddata & 97.5 & 98.2 & 80.3 & 86.2 & 73.0 & 81.3 \\\hline\hline
%	CLAHE N/D \clahe \diverse \nddata & 97.4 & 97.8 & 80.8 & 86.5 & 74.0 & 82.2 \\\hline\hline
%
    \hedn GAN \clahe & 97.1 & 98.2 & 84.5 & 86.9 & 77.1 & 80.3 \\\hline
    \hedn GAN \clahe \diverse & 97.5 & 98.0 & 84.3 & 88.3 & 77.9 & 81.1 \\\hline
%	CycleGAN \clahe & 97.6 & 98.2 & 85.7 & 88.5 & 78.8 & 81.4 \\\hline
%	CycleGAN \clahe \diverse & 97.8 & 98.2 & 86.8 & 89.3 & 80.0 & 82.6 \\\hline
\end{tabular}\\[3pt]
    \caption{Retrieval performance (mAP) on \textit{Tokyo} for a combination of three different subsets of the dataset -- day (D), sunset (S), and night (N). Images from the first class are always queries and from the second class are positives (query$\rightarrow$positive); the last image of the scene from the unused class is excluded from the evaluation. Scores for selected methods from Tab.~\ref{tab:diverse} are reported.}
\label{tab:titech_classes}
\end{table}

\subsection{Diverse anchors}

The effect of diverse anchor mining (\diverse) is ablated in Table~\ref{tab:diverse}. Results show consistent improvement of diverse anchor mining across both baselines and our method on day-night retrieval (\textit{Tokyo} dataset), while having little impact on Oxford and Paris datasets. In Table~\ref{tab:titech_classes}, the results are broken down into combinations of query and result domains (for example, column D$\veryshortarrow$S means querying with a Day image where the Sunset image of the same scene is the only positive). Experiments further show that the proposed \hedn GAN \clahe \diverse method substantially improves the performance when querying or retrieving night images. We interpret this observation as the embedding learned by the proposed method being better at discriminating night images. A similar trend with a smaller improvement can be seen when training with paired day-night data or when diverse anchors are used. We also observe consistent gains when the image photometric normalization CLAHE, proposed by~\cite{jenicek}, is used.

\subsection{Impact of training data}

In this part, we study how sensitive the retrieval performance is to the choice of the night image generator and the amount of night images used to train the embedding network.

\paragraph{Night image generator.}
Different night image generators were compared: CycleGAN, DRIT~\cite{lee2020drit}, CUT~\cite{park2020contrastive}, CyEDA~\cite{beh2022cyeda}, and a proposed edge-consistency \rcfn GAN, HEDGAN, and \hedn GAN. In our experiments, the best-performing generator (Top GAN) is CycleGAN, see Table~\ref{tab:traindata}. Despite different levels of a photo-realistic perception, ranging from DRIT to rather abstract \hedn GAN, the difference in performance is not so severe, all generators outperforming training with real night images \textit{SfM-N/D}. 
Despite similar scores, the generator training times differ greatly, as illustrated in Table~\ref{tab:resources}. The lightest model to train is HEDGAN, followed by \hedn GAN which both train a degree of magnitude faster than CycleGAN and DRIT.

To some observers, the translated night images (see Figure~\ref{fig:translated}) may appear as intensity inverted images. Using images with inverted intensity channel in the LAB colour space improves the results (avg 71.2) over the baseline (avg 70.0), however, such a simple colour augmentation is far below the results achieved by the proposed trained generators or training on the \textit{SfM-N/D} dataset.  

\paragraph{Night data amount.}
We tested the sensitivity of using a different ratio of day and night training data in the embedding learning. We observe that using 25\% or 50\% of night images for learning the embedding does not make a significant difference, with a minor increase of scores on the \textit{Tokyo} dataset. 
In all experiments, we report results for the variant with 25\% of night images in order to stay consistent with the prior work of~\cite{jenicek}.
We also observe that adding true night images from \textit{SfM-N/D} (GAN + \textit{SfM-N/D}) has a minimal impact on the retrieval results.

The experiments show that photo-realistic appearance of the synthetic training images is {\em not} important for retrieval performance,
and that using more powerful generator would not improve the performance, since even adding true night data does not.
However, (local) similarity to real night images encouraged by the discriminator is important (which is not the case for a simple intensity inversion augmentation). 

\setlength{\tabcolsep}{5pt}
\newcommand{\firstsup}[1]{{\bf\color{red} #1}}
\newcommand{\secondsup}[1]{{\bf #1}}
\newcommand{\hlinex}{\hline}
\begin{table}[t] \centering
% \mbox{}\hspace{10pt}\rotatebox[origin=c]{90}{All \clahe \diverse \quad}\hspace{-3ex}
% \begin{tabular}{|p{2ex}l||r||r|r|r|} \hline % f5d5
\begin{tabular}{|l||r||r|r|r|} \hline % f5d5
	Night Data & \!Avg\! & \!Tokyo\! &\!\roxf\!&\!\rpar\!\\\hline\hline
    DRIT \clahe \diverse & 73.5 & {\bf 90.2} & 59.8 & 70.5 \\\hlinex
	CUT \clahe \diverse & 73.0 & 87.7 & 60.3 & {\bf 71.1} \\\hlinex
%    CUT (tuned) \clahe \diverse & 73.4 & 88.6 & 60.8 & 70.8 \\\hlinex
    CyEDA~\cite{beh2022cyeda} BDD \clahe\diverse & 72.9 & 87.9 & 60.4 & 70.3 \\\hlinex
	CyEDA~\cite{beh2022cyeda} GTA \clahe\diverse & 72.9 & 87.9 & 60.3 & 70.4 \\\hlinex
	CyEDA \clahe\diverse & 70.9 & 82.0 & 60.1 & 70.5 \\\hlinex
%    \rcfn GAN \clahe\diverse & 73.5 & 89.5 & 60.2 & 70.8 \\\hlinex
    \rcfn GAN \clahe\diverse & 73.2 & 88.3 & 60.4 & 70.8 \\\hlinex
%	CyEDA SfM (tuned) \clahe\diverse & 73.3 & 88.2 & 60.5 & 71.2 \\\hlinex
    HEDGAN \clahe \diverse & 73.2 & 88.1 & 61.0 & 70.5 \\\hlinex
	\hedn GAN \clahe \diverse & 73.4 & 88.9 & {\bf 61.1} & 70.3 \\\hlinex
%	\hedn GAN 50\% \clahe \diverse & 73.4 & 90.3 & 60.0 & 70.0 \\\hlinex
	C-GAN \clahe \diverse & {\bf 74.0} & {\bf 90.2} & 60.7 & 71.0 \\\hline\hline
%	CycleGAN 50\% \clahe \diverse & 73.9 & {\bf 91.4} & 60.0 & 70.4 \\\hline\hline
% 	CycleGAN Aachen \clahe \diverse & 73.8 & 89.9 & 60.7 & 70.8 \\\hlinex
%
	C-GAN+N/D \clahe \diverse \nddata & 73.5 & 88.6 & 60.8 & {\bf 71.1} \\\hlinex
	C-GAN+N/D 50\% \clahe \diverse \nddata & {\bf 73.9} & {\bf 90.1} & {\bf 61.1} & 70.6 \\\hlinex
%	GAN+GAN Aachen \clahe \diverse & {\bf 74.0} & 90.4 & 60.5 & {\bf 71.1} \\\hlinex
%	GAN+GAN Aachen 50\% \clahe \diverse & {\bf 74.0} & {\bf 91.4} & 60.3 & 70.4 \\\hline
\end{tabular}\\[3pt]
    \caption{The impact of retrieval training data. In the top block, generator architectures DRIT~\cite{lee2020drit}, CUT~\cite{park2020contrastive}, CyEDA~\cite{beh2022cyeda} (pretrained models from~\cite{beh2022cyeda} and trained by us on \textit{SfM}), \rcfn GAN (trained RCF), HEDGAN (frozen HED), \hedn GAN (trained HED), and CycleGAN~\cite{cycleGAN} are tested. In the bottom block, the best performing CycleGAN generator architecture is further combined with the \textit{SfM-N/D} dataset with ratio 1:1 (CycleGAN+N/D); scores for 25\% (default in experiments) and 50\% of night images in the training data are reported. Methods marked by \nddata use paired day-night training data. The best score for each dataset in each block is in bold.}
%  and a version of our method with both night image generators (GAN+GAN Aachen), one trained on SfM and one on Aachen,
\label{tab:traindata}
\end{table}

\setlength{\tabcolsep}{5.4pt}
\begin{table}[t] \centering
\begin{tabular}{|l||r|r||r|r|r|} \hline
    Training & \!Train (h)\! & \!Epoch (h)\! & \!Epochs\! & \!Params\! \\\hline\hline
    DRIT        & 196 & 0:39 & 300 & 75.5 \\\hline
    CycleGAN    & 102 & 1:01 & 100 & 51.0 \\\hline
    CUT         & 35 & 0:42 & 50 & 26.0 \\\hline
    CyEDA       & 28 & 0:24 & 69 & 39.6 \\\hline
    \rcfn GAN   & 27 & 0:32 & 50 & 40.3 \\\hline
    \hedn GAN   & 21 & 0:25 & 50 & 40.2 \\\hline
%    RCFGAN      & 16 & 0:20 & 50 & 25.5 \\\hline
    HEDGAN      & 16 & 0:19 & 50 & 25.5 \\\hline
\end{tabular}\\[3pt]
\caption{Generators training comparison. In the first two columns, training times in hours are reported (measured on NVIDIA Tesla P100 16GB). In the pre-last column, the total number of epochs necessary to converge to the top performance is reported. In the last column, a number of trainable parameters in millions is illustrated.}
\label{tab:resources}
\end{table}

%-------------------------------------------------------------------------
\section{Conclusions}
\label{sec:conclusion}
The training of a deep neural network that embeds images into a descriptor space suitable for image retrieval insensitive to severe day-night illumination changes was proposed. Synthetically generated night images are used so that the training does not require corresponding pairs of night and day images. The proposed method outperforms prior work, including the work of~\cite{jenicek} which uses ground-truth day-night annotated image pairs.

We have shown that the proposed method is capable of generating diverse training examples, and that 
a larger diversity of synthesized training data outperforms smaller diversity of real training data.

Besides evaluating a number of existing generators, a light-weight generator exploiting edge consistency was proposed. In our \hedn GAN method, day/night edge detector \hednx is trained. Its superior performance to HED detector was shown both qualitatively and quantitatively.

Finally, we have introduced a method of mining diverse anchor images, that further improves the diversity in the training data which is reflected in increased retrieval performance. Such an approach is applicable to other metric-learning and similar tasks where strong modes of the training-data distribution do not correspond to the distribution of test data.

\section{Acknowledgements}

This research was supported by the Research Center for Informatics (project CZ.02.1.01/0.0/0.0/16 019/0000765 funded by OP VVV) and by the Grant Agency of the Czech Technical University in Prague, grant No. SGS23/173/OHK3/3T/13.

{\small
\bibliographystyle{ieee_fullname}
\bibliography{egbib}
}

\end{document}

\maketitle

\section{Appendix A}

Full table versions. Method variants not shown in the main paper are emphasized by bold.

In Table~\ref{tab:results}, and Table~\ref{tab:traindata}, we also show results using image generators tuned by us. Specifically in CUT~\cite{park2020contrastive}, the downsampling and upsampling convolutional layers are trainable in both the generator and discriminator, generator weights are initialized following~\cite{he2015delving} as in our \hedn GAN, the contrastive loss is applied on the output of a different set of layers (4,7,10,14), and weight $\lambda_Y$ of identity loss is set to 10. We observe that learning the convolutional layers and changing the layers in the contrastive loss aids the performance of the final retrieval models, while the other changes has shown to stabilize training in our setup.
In CyEDA~\cite{beh2022cyeda}, while training on the \textit{SfM}120k dataset~\cite{radenovic2018fine}, weight $\lambda$ of cycle consistency is set to $0.3$. We observed that the original $\lambda = 10$ causes the night generator to synthesise images nearly identical to the input day images. 

In the Table~\ref{tab:traindata}, we also tested augmentation with CycleGAN trained on the \textit{Aachen} dataset~\cite{Sattler2012BMVC, Sattler2018CVPR, zhang2021reference}, with both Aachen v1.0 and Aachen v1.1 images together.
\textit{Aachen} dataset contains much less night training examples than \textit{SfM}120k dataset (206 compared to 10039, respectively).
Despite that, the performance loss of CycleGAN Aachen augmentation is negligible, see Table~\ref{tab:traindata}.

% aachen 1.0 has 226 night images, aachen 1.1 has 93 night images, but 206 used for CycleGAN training
% /mnt/datagrid/personal/mohwaalb/cirtorch/data/train/aachen_v1.1/dataset/train_night_all.txt

\renewcommand{\thetable}{A\arabic{table}}

\begin{table}[t] \centering
\newcommand{\firstsup}[1]{{\bf\color{red} #1}}
\newcommand{\secondsup}[1]{{\bf #1}}
\begin{tabular}{|l||r||r|r|r|}
    \multicolumn{5}{l}{VGG-16 backbone}\\\hline % 89e8
    Method & Avg & Tokyo & \roxf & \rpar \\\hline
    GeM~\cite{radenovic2018fine} & 69.9 & 79.4 & \secondsup{60.9} & 69.3 \\\hline
    GeM N/D~\cite{jenicek}~\nddata & 71.1 & 83.5 & 60.0 & 69.8 \\\hline
    CIConv~\cite{Lengyel-ICCV21} & - & 83.3 & - & - \\\hline
%    EdgeMAC~\cite{Radenovic-ECCV18} & 45.6 & 75.9 & 17.3 & 43.5 \\\hline
    CLAHE \clahe~\cite{jenicek} & 71.6 & 84.1 & 60.8 & 69.8 \\\hline
%	CLAHE \clahe \diverse (our \diverse) & 72.2 & 85.9 & 60.3 & 70.5 \\\hline
    CLAHE N/D \clahe~\cite{jenicek}~\nddata & 72.4 & 87.0 & 60.2 & 70.0 \\\hline
%	CLAHE N/D \clahe \diverse (our \diverse) \nddata & 73.0 & 87.7 & 60.8 & \secondsup{70.7} \\\hline
	\hedn GAN \clahe \diverse (ours) & \secondsup{73.4} & \secondsup{88.9} & \firstsup{61.1} & \secondsup{70.3} \\\hline
	CycleGAN \clahe \diverse (ours) & \firstsup{74.0} & \firstsup{90.2} & 60.7 & \firstsup{71.0} \\\hline
%
% ----------
%
    \multicolumn{5}{l}{}\\[-0.2em]
    \multicolumn{5}{l}{\bf ResNet-18 backbone}\\\hline
    Method & Avg & Tokyo & \roxf & \rpar \\\hline
    \bf GeM & 65.4 & 76.1 & 50.5 & 69.5 \\\hline
    \bf GeM N/D \nddata & 67.0 & 79.6 & 51.3 & \firstsup{70.0} \\\hline
    \bf CLAHE \clahe & 67.3 & 80.6 & 52.6 & 68.6 \\\hline
    \bf CLAHE N/D \clahe~\nddata & 68.0 & 82.5 & 52.5 & 69.0 \\\hline
    \bf \rcfn GAN \clahe \diverse (ours) & 69.4 & 83.5 & \secondsup{54.4} & \firstsup{70.0} \\\hline
	\bf \hedn GAN \clahe \diverse (ours) & \secondsup{69.6} & \firstsup{85.1} & 53.6 & \firstsup{70.0} \\\hline
	\bf CycleGAN \clahe \diverse (ours) & \firstsup{69.7} & \secondsup{84.4} & \firstsup{55.1} & 69.6 \\\hline
%
% ----------
%
    \multicolumn{5}{l}{}\\[-0.2em]
    \multicolumn{5}{l}{\bf ResNet-50 backbone}\\\hline
    Method & Avg & Tokyo & \roxf & \rpar \\\hline
    \bf GeM & 74.6 & 85.4 & 63.4 & \secondsup{75.1} \\\hline
    \bf GeM N/D \nddata & 75.7 & 88.3 & 63.1 & \firstsup{75.6} \\\hline
    \bf CLAHE \clahe & 74.7 & 87.4 & 62.5 & 74.2 \\\hline
    \bf CLAHE N/D \clahe~\nddata & 75.3 & 89.0 & 62.3 & 74.5 \\\hline
    \bf \rcfn GAN \clahe \diverse (ours) & 76.8 & 91.4 & \firstsup{64.5} & 74.4 \\\hline
	\bf \hedn GAN \clahe \diverse (ours) & \firstsup{77.0} & \secondsup{91.7} & \secondsup{64.4} & 74.9 \\\hline
	\bf CycleGAN \clahe \diverse (ours) & \firstsup{77.0} & \firstsup{92.3} & 64.0 & 74.7 \\\hline
%
% ----------
%
    \multicolumn{5}{l}{}\\[-0.2em]
    \multicolumn{5}{l}{ResNet-101 backbone}\\\hline
        Method & Avg & Tokyo & \roxf & \rpar \\\hline
        GeM~\cite{radenovic2018fine} & 75.7 & 85.0 & 65.3 & 76.7 \\\hline
    \bf GeM N/D \nddata & 77.0 & 88.6 & 65.7 & 76.8 \\\hline
        CIConv~\cite{Lengyel-ICCV21} & 75.0 & 88.3 & 62.0 & 74.7 \\\hline
    \bf CLAHE \clahe & 76.9 & 88.1 & 66.1 & 76.6 \\\hline
    \bf CLAHE N/D \clahe~\nddata & 77.4 & 89.5 & 66.1 & 76.5 \\\hline
%    HEDGAN \clahe \diverse (ours) & 78.0 & 91.2 & 66.5 & 76.4 \\\hline
    \bf CUT \clahe \diverse & 77.9 & 90.2 & 65.7 & \firstsup{77.7} \\\hline
    \bf CUT (tuned) \clahe \diverse & 78.0 & 90.9 & 65.7 & 77.3 \\\hline
    \bf CyEDA BDD \clahe \diverse & 77.8 & 90.3 & 65.7 & 77.3 \\\hline
    \bf CyEDA GTA \clahe \diverse & 78.2 & 91.2 & 65.8 & \secondsup{77.6} \\\hline
    \bf CyEDA (tuned) \clahe \diverse & 78.0 & 90.9 & 65.5 & 77.5 \\\hline
    \bf \rcfn GAN \clahe \diverse (ours) & 78.2 & 91.5 & \firstsup{66.8} & 76.3 \\\hline
        \hedn GAN \clahe \diverse (ours) & \firstsup{78.4} & \firstsup{92.2} & 66.3 & 76.6 \\\hline
	CycleGAN \clahe \diverse (ours) & \firstsup{78.4} & \secondsup{92.0} & \firstsup{66.8} & 76.4 \\\hline
%
% ----------
%
    % \multicolumn{5}{l}{}\\[-0.2em]
    % \multicolumn{5}{l}{HOW ResNet-18 backbone}\\\hline
    % Method & Avg & Tokyo & \roxf & \rpar \\\hline
    % HOW~\cite{tolias2020learning} & 80.8 & 87.8 & \secondsup{75.1} & 79.4 \\\hline
    % HOW N/D~\nddata & \secondsup{82.0} & 89.2 & \firstsup{75.5} & \firstsup{81.4} \\\hline
    % \hedn GAN \clahe \diverse (ours) & \secondsup{82.0} & \secondsup{91.6} & 74.6 & 79.7 \\\hline
    % CycleGAN \clahe \diverse (ours) & \firstsup{82.4} & \firstsup{92.9} & 74.6 & \secondsup{79.8} \\\hline
	
\end{tabular}\\[15pt]

(continues)

\end{table}

\begin{table}[t] \centering
\newcommand{\firstsup}[1]{{\bf\color{red} #1}}
\newcommand{\secondsup}[1]{{\bf #1}}
\begin{tabular}{|l||r||r|r|r|}

    \multicolumn{5}{l}{}\\[-0.2em]
    \multicolumn{5}{l}{HOW ResNet-18 backbone}\\\hline
    Method & Avg & Tokyo & \roxf & \rpar \\\hline
    HOW~\cite{tolias2020learning} & 80.8 & 87.8 & \secondsup{75.1} & 79.4 \\\hline
    HOW N/D~\nddata & \secondsup{82.0} & 89.2 & \firstsup{75.5} & \firstsup{81.4} \\\hline
    \bf \rcfn GAN \clahe \diverse (ours) & 81.8 & 91.5 & 74.6 & 79.4 \\\hline
    \hedn GAN \clahe \diverse (ours) & \secondsup{82.0} & \secondsup{91.6} & 74.6 & 79.7 \\\hline
    CycleGAN \clahe \diverse (ours) & \firstsup{82.4} & \firstsup{92.9} & 74.6 & \secondsup{79.8} \\\hline
	
\end{tabular}\\[3pt]

    \caption{Comparison in terms of mAP on \Tokyo, \roxf Medium and \rpar Medium datasets and their average on retrieval. Methods marked by \nddata use paired day-night training data. Methods starting with GeM and CLAHE not marked with a reference were trained by us. The best score for each backbone architecture (in separate tables) is emphasized by red bold, second best by bold.}
\label{tab:results}
\end{table}

\setcounter{table}{3}
\begin{table}[t] \centering
\newcommand{\doubledim}{{\color{red}\bf \textdaggerdbl}\xspace}
\newcommand{\firstsup}[1]{{\bf #1}}
\newcommand{\secondsup}[1]{#1}
\begin{tabular}{|l||r||r|r|r|} \hline
	Method & Avg & Tokyo & \roxf & \rpar \\\hline\hline
    EdgeMAC~\cite{Radenovic-ECCV18} & 45.6 & 75.9 & 17.3 & 43.5 \\\hline
    HEDMAC & 56.8 & 79.5 & 38.3 & 52.5 \\\hline
    \hedn MAC & \firstsup{59.2} & 81.9 & \firstsup{38.4} & \firstsup{57.2} \\\hline
    \rcfn MAC & 58.5 & \firstsup{88.9} & 35.1 & 51.4 \\\hline\hline
    HEDMAC+GeM \doubledim & 72.0 & 84.8 & 60.9 & 70.3 \\\hline
    \hedn MAC+GeM \doubledim & 72.6 & 85.7 & \firstsup{61.1} & \secondsup{70.9} \\\hline
    \bf HEDMAC+\textsuperscript{N}GAN \doubledim & \secondsup{73.8} & \secondsup{90.9} & 60.1 & \secondsup{70.5} \\\hline
    \hedn MAC+\textsuperscript{N}GAN \doubledim & \secondsup{74.4} & \secondsup{91.4} & \secondsup{60.6} & \secondsup{71.3} \\\hline
    \bf HEDMAC+GAN \doubledim & \secondsup{74.2} & \secondsup{91.5} & 60.0 & \secondsup{71.2} \\\hline
    \hedn MAC+GAN \doubledim & \firstsup{74.7} & \firstsup{91.8} & \secondsup{60.4} & \firstsup{71.9} \\\hline
    
\end{tabular}\\[3pt]
    \caption{The effect of our trained \hedn~detector (from \hedn GAN) on the EdgeMAC~\cite{Radenovic-ECCV18} method. HEDMAC and \hedn MAC is a variant of EdgeMAC method with the HED~\cite{xie2015holistically} edge detector with either original or our weights, respectively. \rcfn MAC is a variant of EdgeMAC with the RCF~\cite{liu2019richer} edge detector with our weights. In the bottom block, ensembles of EdgeMAC variants with chosen methods from Table~\ref{tab:results} are reported. GeM is from~\cite{radenovic2018fine}, \textsuperscript{N}GAN corresponds to \hedn GAN \clahe \diverse, and GAN to CycleGAN \clahe \diverse, all from Table~\ref{tab:results}. Ensembles have double the dimensionality (1024) and are marked with \doubledim. The best scores for each dimensionality are in bold.}
\label{tab:edge}
\end{table}

\begin{table}[t] \centering
\newcommand{\firstsup}[1]{{\bf\color{red} #1}}
\newcommand{\secondsup}[1]{{\bf #1}}
\begin{tabular}{|l||r||r|r|r|} \hline % 0f01
	Method & Avg & Tokyo & \roxf & \rpar \\\hline\hline
	\bf GeM & 70.0 & 80.4 & 59.9 & 69.8 \\\hline
	\bf GeM \diverse & 70.3 & 81.2 & 59.8 & 69.9 \\\hline
	CLAHE \clahe & 71.9 & 85.4 & 60.0 & 70.1 \\\hline
	CLAHE \clahe \diverse & {\bf 72.2} & {\bf 85.9} & {\bf 60.3} & {\bf 70.5} \\\hline\hline
	\bf GeM N/D \nddata & 71.5 & 84.0 & 60.4 & 70.0 \\\hline
	\bf GeM N/D \diverse \nddata & 71.5 & 84.1 & 60.3 & 70.1 \\\hline
	CLAHE N/D \clahe~\nddata & 72.5 & 87.5 & 59.9 & 70.1 \\\hline
	CLAHE N/D \clahe \diverse \nddata & {\bf 73.0} & {\bf 87.7} & {\bf 60.8} & {\bf 70.7} \\\hline\hline
        \bf \rcfn GAN & 72.3 & 86.8 & 59.8 & 70.2 \\\hline
	\bf \rcfn GAN \diverse & 72.7 & 87.6 & 60.1 & 70.4 \\\hline
	\bf \rcfn GAN \clahe & 72.9 & 87.8 & {\bf 60.4} & 70.6 \\\hline
	\bf \rcfn GAN \clahe \diverse & {\bf 73.2} & {\bf 88.3} & {\bf 60.4} & {\bf 70.8} \\\hline\hline
	\hedn GAN & 72.7 & 88.0 & 60.2 & 70.0 \\\hline
	\bf \hedn GAN \diverse & 73.0 & 88.7 & 60.2 & 70.1 \\\hline
	\hedn GAN \clahe & 73.2 & 88.7 & 60.5 & 70.4 \\\hline
	\hedn GAN \clahe \diverse & {\bf 73.4} & {\bf 88.8} & {\bf 60.7} & {\bf 70.6} \\\hline\hline
	\bf CycleGAN & 73.0 & 88.8 & 59.6 & 70.5 \\\hline
	\bf CycleGAN \diverse & 73.3 & 89.1 & 59.9 & 70.7 \\\hline
	\bf CycleGAN \clahe & 73.6 & 89.6 & 60.5 & 70.9 \\\hline
	\bf CycleGAN \clahe \diverse & {\bf 74.0} & {\bf 90.2} & {\bf 60.7} & {\bf 71.0} \\\hline
\end{tabular}\\[3pt]
    \caption{The effect of diverse anchors (\diverse). Methods GeM~\cite{radenovic2018fine}, GeM N/D, CLAHE \clahe~\cite{jenicek}, and CLAHE N/D \clahe~\cite{jenicek} from Table~\ref{tab:results} are reported in the top two blocks. Please note that we re-train the models for this ablation, so we obtain a slightly higher performance. In the last three blocks, the effect of CLAHE (\clahe) and diverse anchors (\diverse) is reported on the \rcfn GAN, \hedn GAN and CycleGAN methods. The best score for each dataset in each block is in bold.}
\label{tab:diverse}
\end{table}

\setlength{\tabcolsep}{3.1pt}

\begin{table}[h] \centering
\begin{tabular}{|l||r|r|r|r|r|r|} \hline % 71f8
    \multirow{2}{*}{Method} & \multicolumn{2}{c|}{\tiny \{day, sunset\}} & \multicolumn{2}{c|}{\tiny \{sunset, night\}} & \multicolumn{2}{c|}{\tiny \{day, night\}}\\
	& D$\veryshortarrow$S & S$\veryshortarrow$D & \hspace{0.1em} S$\veryshortarrow$N & N$\veryshortarrow$S & D$\veryshortarrow$N & N$\veryshortarrow$D \\\hline\hline
    CLAHE \clahe & 97.7 & 98.2 & 80.1 & 81.3 & 70.9 & 76.1 \\\hline
    \textbf{CLAHE \clahe \diverse} & 97.5 & 98.4 & 80.2 & 81.3 & 72.4 & 77.0 \\\hline\hline
    CLAHE N/D \clahe~\nddata & 97.5 & 98.2 & 80.3 & 86.2 & 73.0 & 81.3 \\\hline\hline
    \textbf{CLAHE N/D \clahe \diverse \nddata} & 97.4 & 97.8 & 80.8 & 86.5 & 74.0 & 82.2 \\\hline\hline
    \hedn GAN \clahe & 97.1 & 98.2 & 84.5 & 86.9 & 77.1 & 80.3 \\\hline
    \hedn GAN \clahe \diverse & 97.5 & 98.0 & 84.3 & 88.3 & 77.9 & 81.1 \\\hline
    \textbf{CycleGAN \clahe} & 97.6 & 98.2 & 85.7 & 88.5 & 78.8 & 81.4 \\\hline
    \textbf{CycleGAN \clahe \diverse} & 97.8 & 98.2 & 86.8 & 89.3 & 80.0 & 82.6 \\\hline
\end{tabular}\\[3pt]
    \caption{Retrieval performance (mAP) on \textit{Tokyo} for a combination of three different subsets of the dataset -- day (D), sunset (S), and night (N). Images from the first class are always queries and from the second class are positives (query$\rightarrow$positive); the last image of the scene from the unused class is excluded from the evaluation. Scores for selected methods from Table~\ref{tab:diverse} are reported.}
\label{tab:titech_classes}
\end{table}
\newcommand{\firstsup}[1]{{\bf\color{red} #1}}
\newcommand{\secondsup}[1]{{\bf #1}}
\newcommand{\hlinex}{\hline}
\begin{table}[t] \centering
% \mbox{}\hspace{10pt}\rotatebox[origin=c]{90}{All \clahe \diverse \quad}\hspace{-3ex}
% \begin{tabular}{|p{2ex}l||r||r|r|r|} \hline % f5d5
\begin{tabular}{|l||r||r|r|r|} \hline % f5d5
	Night Data & Avg & Tokyo & \roxf & \rpar \\\hline\hline
    DRIT \clahe \diverse & 73.5 & 90.2 & 59.8 & 70.5 \\\hlinex
    CUT \clahe \diverse & 73.0 & 87.7 & 60.3 & {\bf 71.1} \\\hlinex
    \bf CUT (tuned) \clahe \diverse & 73.4 & 88.6 & 60.8 & 70.8 \\\hlinex
    CyEDA~\cite{beh2022cyeda} BDD \clahe\diverse & 72.9 & 87.9 & 60.4 & 70.3 \\\hlinex
    CyEDA~\cite{beh2022cyeda} GTA \clahe\diverse & 72.9 & 87.9 & 60.3 & 70.4 \\\hlinex
    CyEDA \clahe\diverse & 70.9 & 82.0 & 60.1 & 70.5 \\\hlinex
    \bf CyEDA (tuned) \clahe\diverse & 73.3 & 88.2 & 60.5 & 71.2 \\\hlinex
    \rcfn GAN \clahe \diverse & 73.2 & 88.3 & 60.4 & 70.8 \\\hlinex
    \bf \rcfn GAN 50\% \clahe \diverse & 73.4 & 90.0 & 59.7 & 70.4 \\\hlinex
    HEDGAN \clahe \diverse & 73.2 & 88.1 & 61.0 & 70.5 \\\hlinex
    \hedn GAN \clahe \diverse & 73.4 & 88.9 & {\bf 61.1} & 70.3 \\\hlinex
    \bf \hedn GAN 50\% \clahe \diverse & 73.4 & 90.3 & 60.0 & 70.0 \\\hlinex
    CycleGAN \clahe \diverse & {\bf 74.0} & 90.2 & 60.7 & 71.0 \\\hline
    \bf CycleGAN 50\% \clahe \diverse & 73.9 & {\bf 91.4} & 60.0 & 70.4 \\\hline
    \bf CycleGAN Aachen \clahe \diverse & 73.8 & 89.9 & 60.7 & 70.8 \\\hline\hline
    CycleGAN + N/D \clahe \diverse \nddata & 73.5 & 88.6 & 60.8 & {\bf 71.1} \\\hline
    CycleGAN + N/D 50\% \clahe \diverse \nddata & 73.9 & 90.1 & {\bf 61.1} & 70.6 \\\hline
    \bf \hedn GAN + N/D \clahe \diverse \nddata & 73.3 & 87.9 & {\bf 61.1} & 70.8 \\\hline
    \bf \hedn GAN + N/D 50\% \clahe \diverse \nddata & 73.6 & 89.1 & {\bf 61.1} & 70.6 \\\hline
    \bf \hedn GAN + HEDGAN \clahe \diverse & 73.4 & 88.0 & 60.7 & 70.6 \\\hline
    \bf \hedn GAN + HEDGAN 50\% \clahe \diverse & 73.4 & 90.0 & 60.5 & 69.8 \\\hline
    \bf CycleGAN + CycleGAN Aachen \clahe \diverse & 74.0 & 90.4 & 60.5 & {\bf 71.1} \\\hlinex
    \bf CycleGAN + CycleGAN Aachen 50\% \clahe \diverse & 74.0 & {\bf 91.4} & 60.3 & 70.4 \\\hline
    \bf \hedn GAN + CycleGAN \clahe \diverse & 74.0 & 90.0 & 61.0 & 70.9 \\\hline
    \bf \hedn GAN + CycleGAN 50\% \clahe \diverse & {\bf 74.1} & {\bf 91.4} & 60.5 & 70.5 \\\hline
\end{tabular}\\[3pt]
    \caption{The impact of retrieval training data. In the top block, generator architectures DRIT~\cite{lee2020drit}, CUT~\cite{park2020contrastive} (original and our tuned variation), CyEDA~\cite{beh2022cyeda} (pretrained models from~\cite{beh2022cyeda}, trained by us on \textit{SfM} dataset, and our tuned variation trained on \textit{SfM}120k), \rcfn GAN (trained RCF), HEDGAN, \hedn GAN (trained HED), and CycleGAN~\cite{cycleGAN} are tested. In the bottom block, the \hedn GAN or CycleGAN generator architecture is further combined with the \textit{SfM-N/D} dataset (\eg \hedn GAN + N/D) or a different generator (\eg \hedn GAN + CycleGAN) with ratio 1:1; scores for 25\% (default in experiments) and 50\% of night images in the training data are reported. Methods marked by \nddata use paired day-night training data. The best score for each dataset in each block is in bold.}
%  and a version of our method with both night image generators (GAN+GAN Aachen), one trained on SfM and one on Aachen,
\label{tab:traindata}
\end{table}

\clearpage

% \section{Appendix B}

% Diverse anchors ablation further expanding the Table~\ref{tab:diverse}.
% In Table~\ref{tab:titech_classes}, the results are broken down into combinations of query and result domains (for example, column D$\veryshortarrow$S means querying with a Day image where the Sunset image of the same scene is the only positive).

% \vspace{30pt}
% \renewcommand{\thetable}{B\arabic{table}}
% \setcounter{table}{0}
% \input{tab/titech_classes.tex}

% \clearpage

\section{Appendix B}

We provide example outputs for all generative models referred in the paper. Generators for translation from day to night as well as from night to day are examined in Figure~\ref{fig:day_examples}, and Figure~\ref{fig:night_examples}, respectively. Please note that \hedn GAN, \rcfn GAN, and CUT architectures do not contain a night $\rightarrow$ day generator.

We observe that for night to day translation (Figure~\ref{fig:night_examples}), the ToDayGAN~\cite{toDayGAN} method achieves satisfactory results only on the \textit{RobotCar} dataset~\cite{RobotCarDatasetIJRR}.
Also, these results are visually more  appealing than other results on this dataset, even though a car is hallucinated close to the middle of the image. This can be explained by the fact that the ToDayGAN method is specifically trained for this dataset.

\renewcommand{\thefigure}{B\arabic{figure}}
\setcounter{figure}{0}
\begin{figure*}
    \newcommand{\examplesize}{.187}
    \centering
    \begin{tabularx}{\textwidth}{YYYYYY}
    Day original \, & ToDayGAN~\cite{toDayGAN} & \hedn GAN & \rcfn GAN & CycleGAN
    \end{tabularx}
    \\[0.1em]
    \subfloat{\includegraphics[width=\examplesize\linewidth]{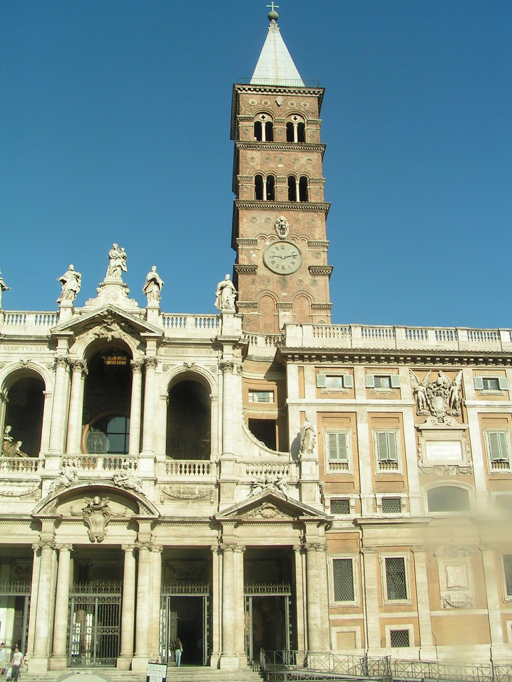}} \,
    \subfloat{\includegraphics[width=\examplesize\linewidth]{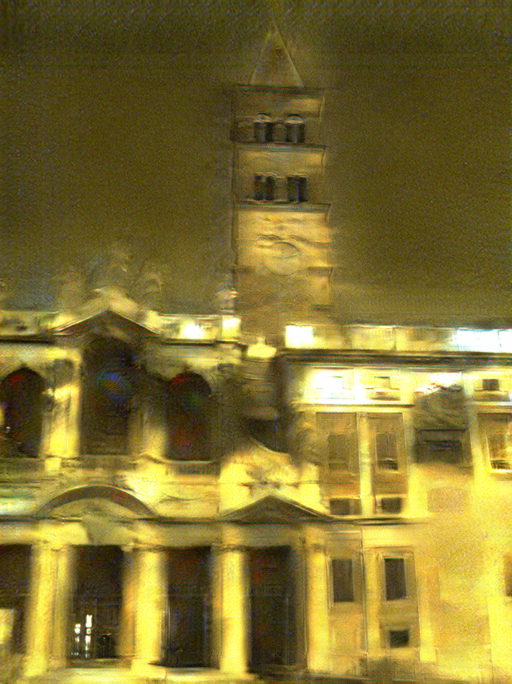}} \,
    \subfloat{\includegraphics[width=\examplesize\linewidth]{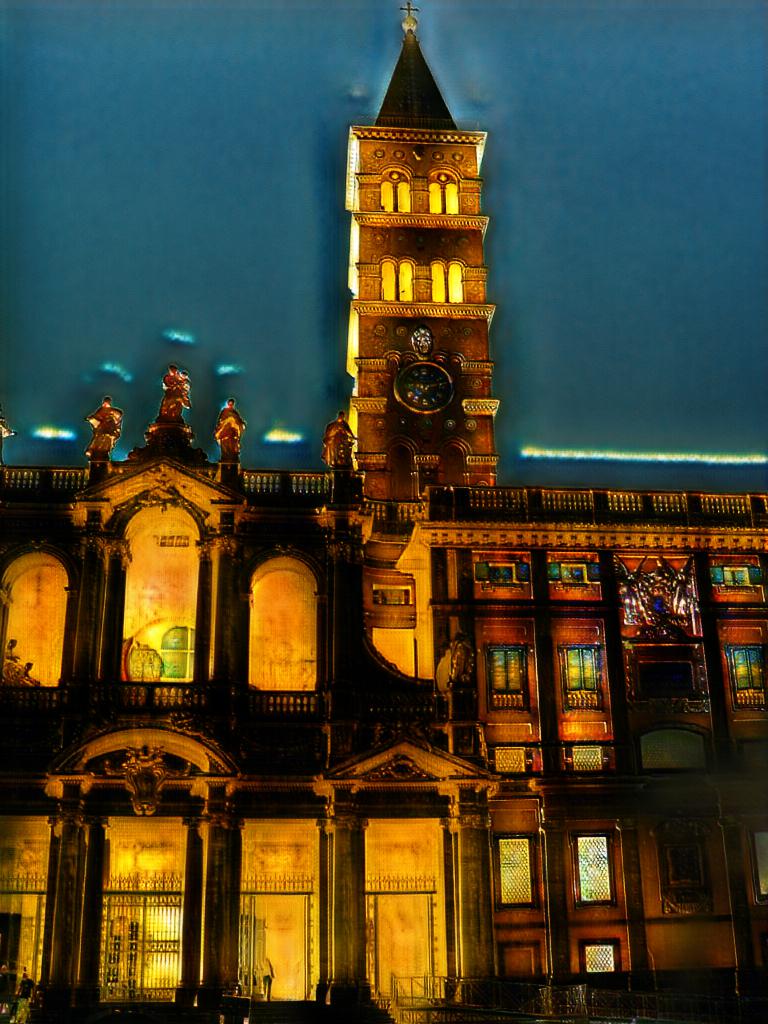}} \,
    \subfloat{\includegraphics[width=\examplesize\linewidth]{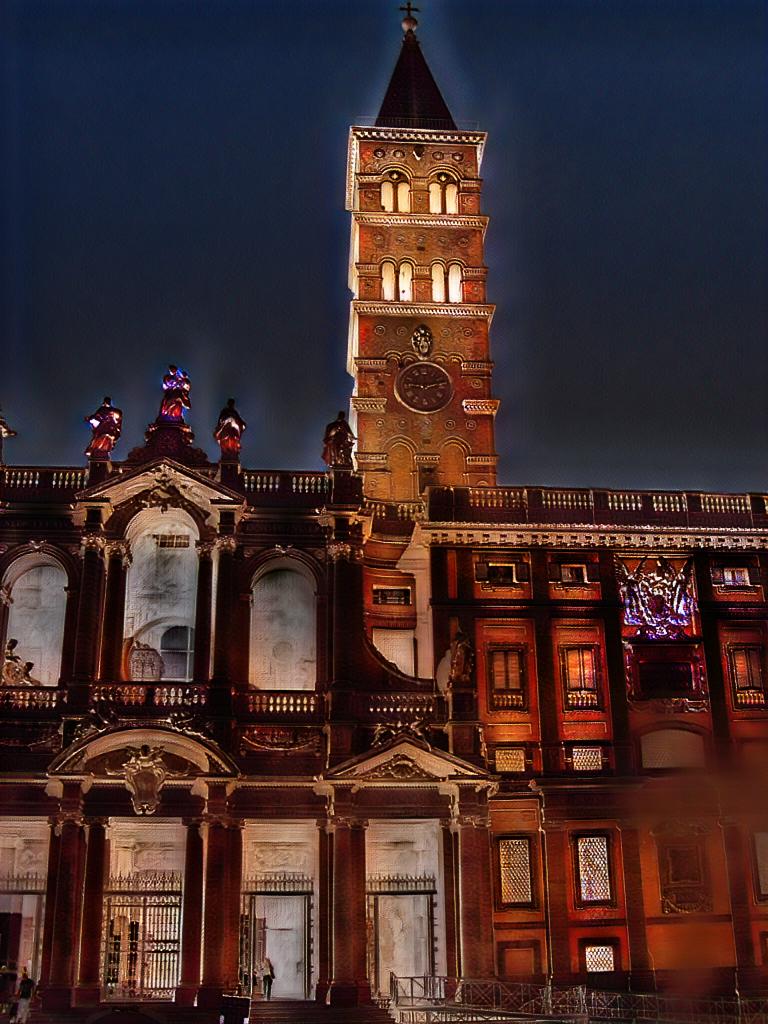}} \,
    \subfloat{\includegraphics[width=\examplesize\linewidth]{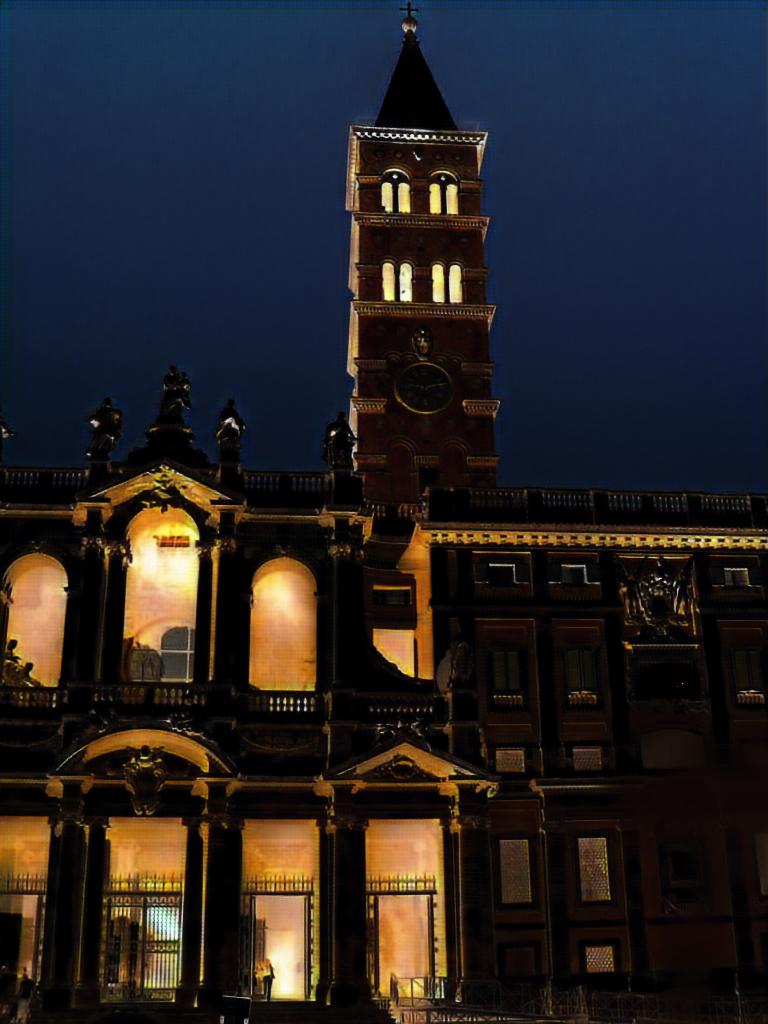}} \,
    \\[0.1em]
    \subfloat{\includegraphics[width=\examplesize\linewidth]{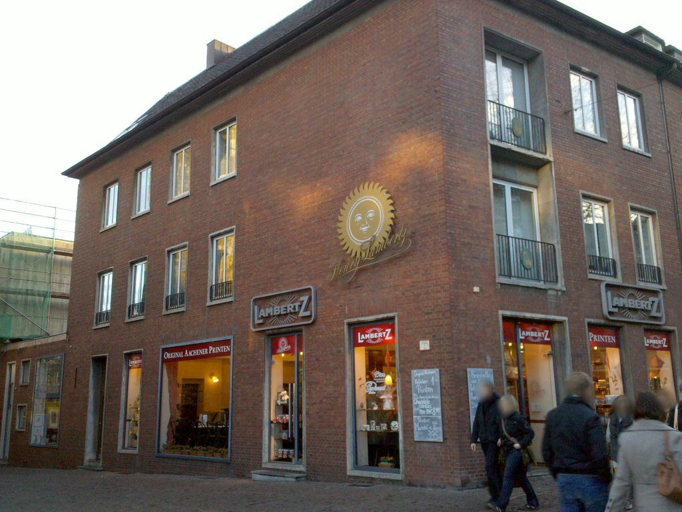}} \,
    \subfloat{\includegraphics[width=\examplesize\linewidth]{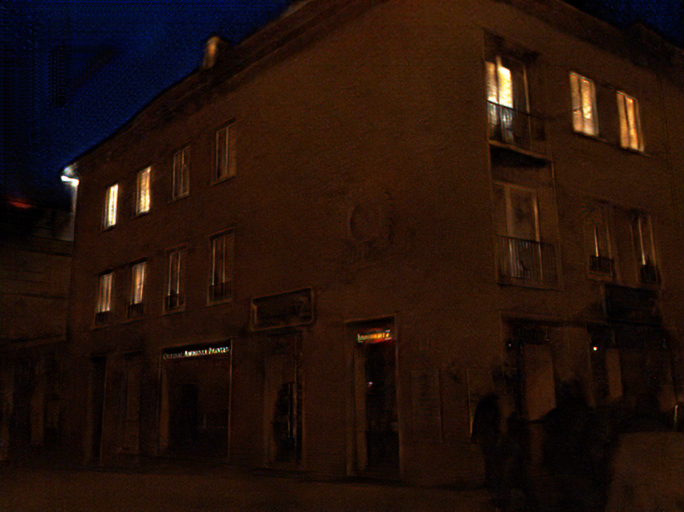}} \,
    \subfloat{\includegraphics[width=\examplesize\linewidth]{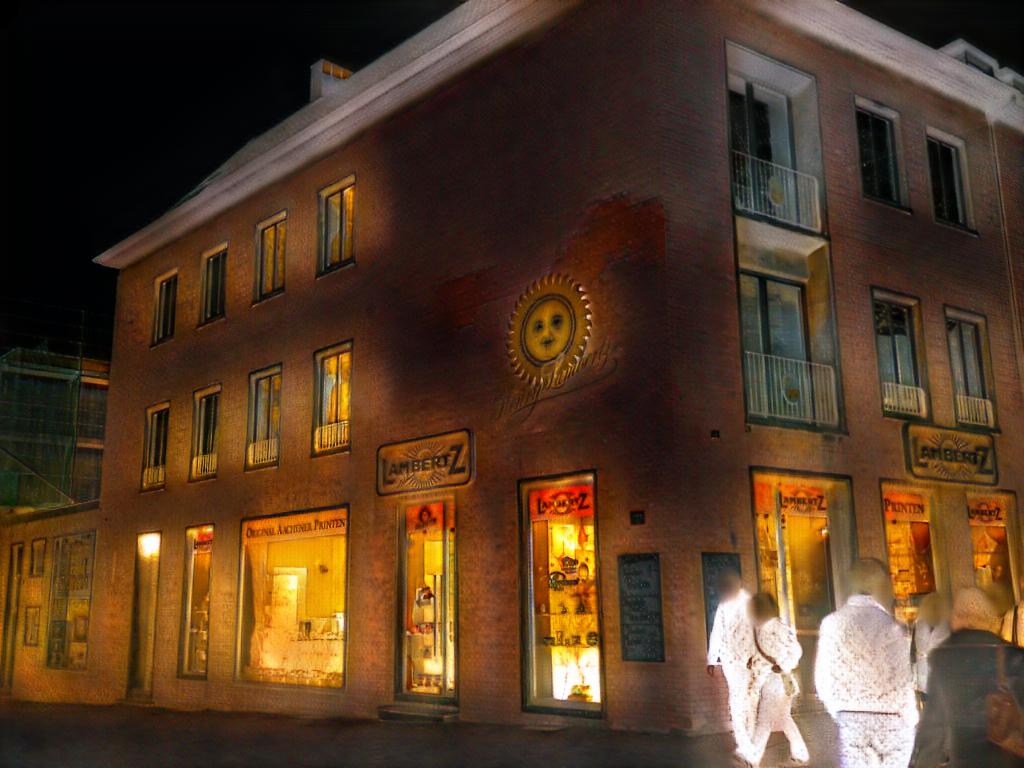}} \,
    \subfloat{\includegraphics[width=\examplesize\linewidth]{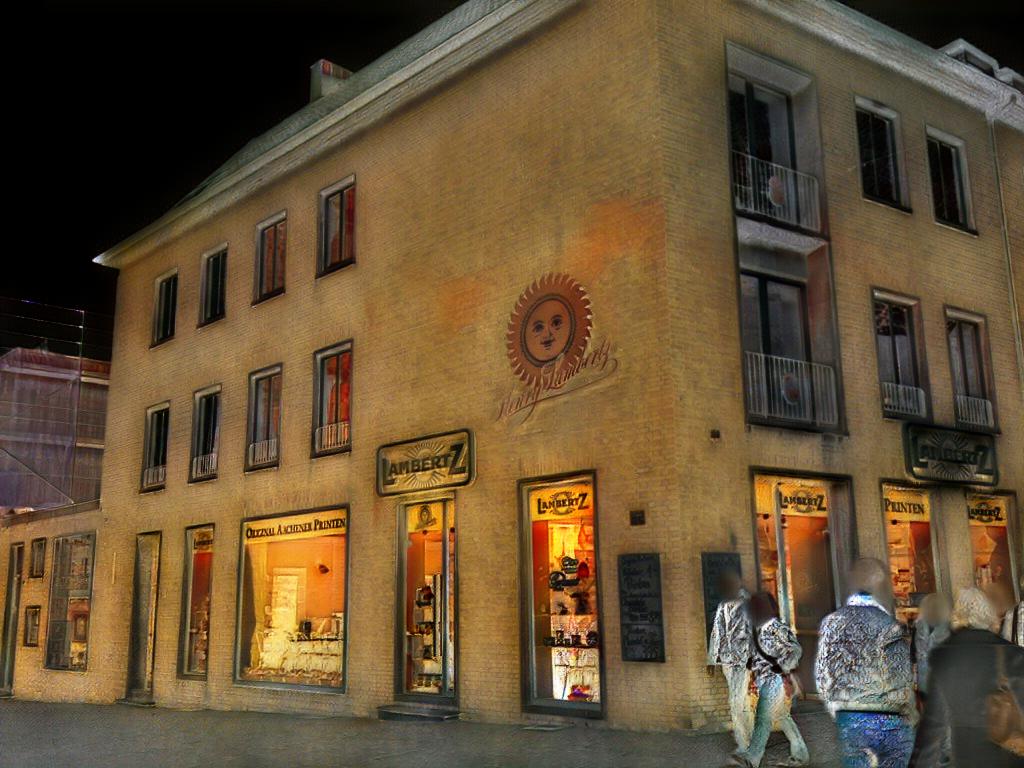}} \,
    \subfloat{\includegraphics[width=\examplesize\linewidth]{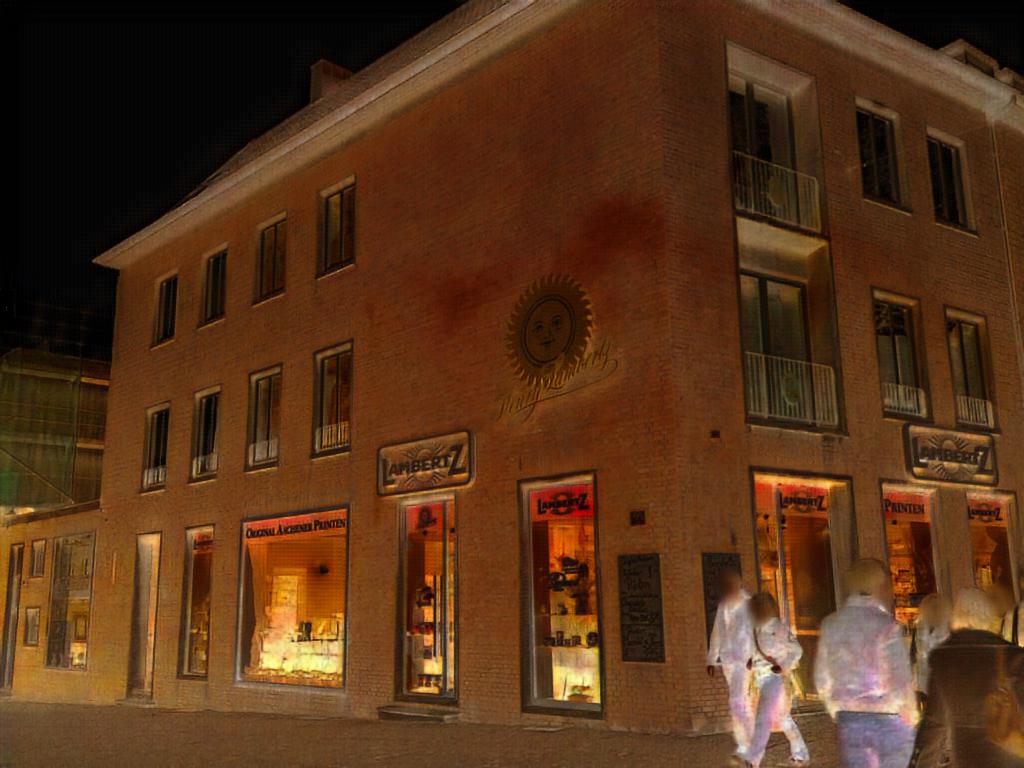}} \,
    \\[0.1em]
    \subfloat{\includegraphics[width=\examplesize\linewidth]{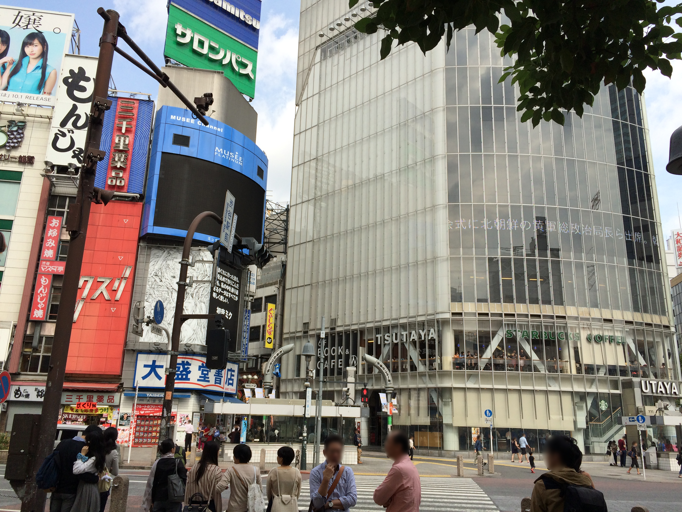}} \,
    \subfloat{\includegraphics[width=\examplesize\linewidth]{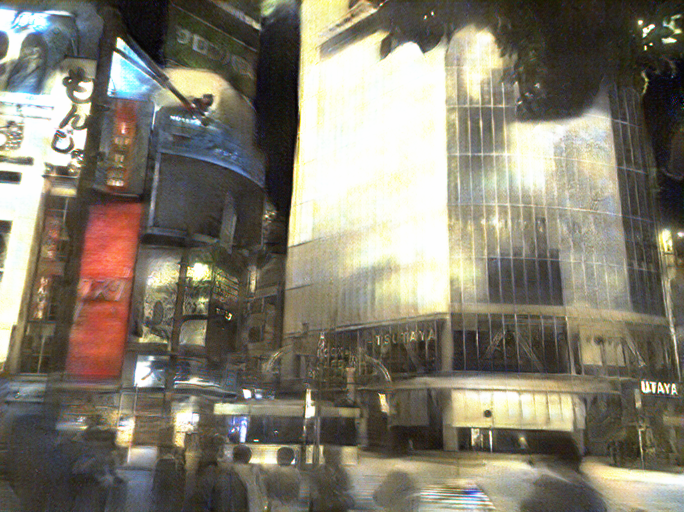}} \,
    \subfloat{\includegraphics[width=\examplesize\linewidth]{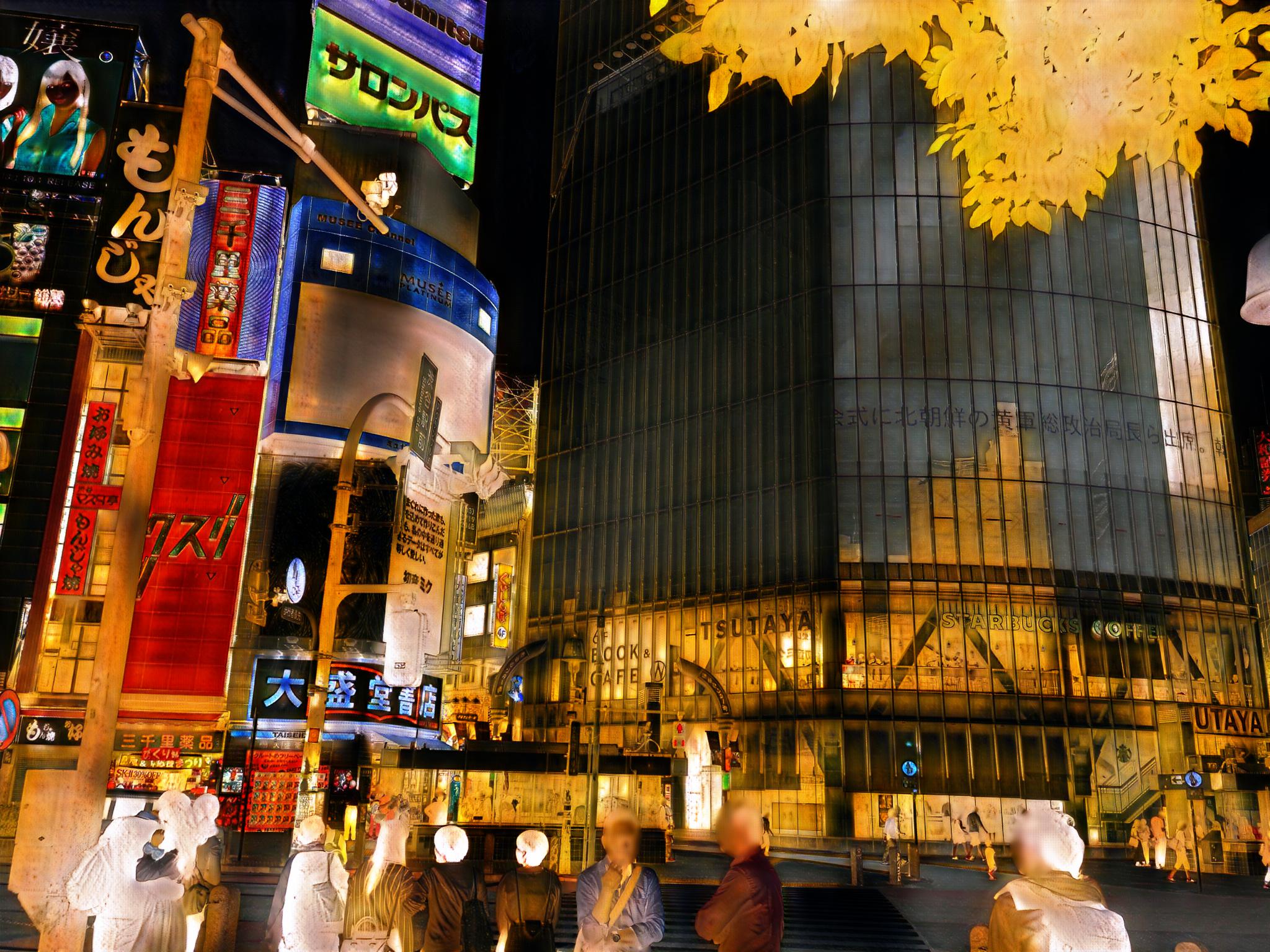}} \,
    \subfloat{\includegraphics[width=\examplesize\linewidth]{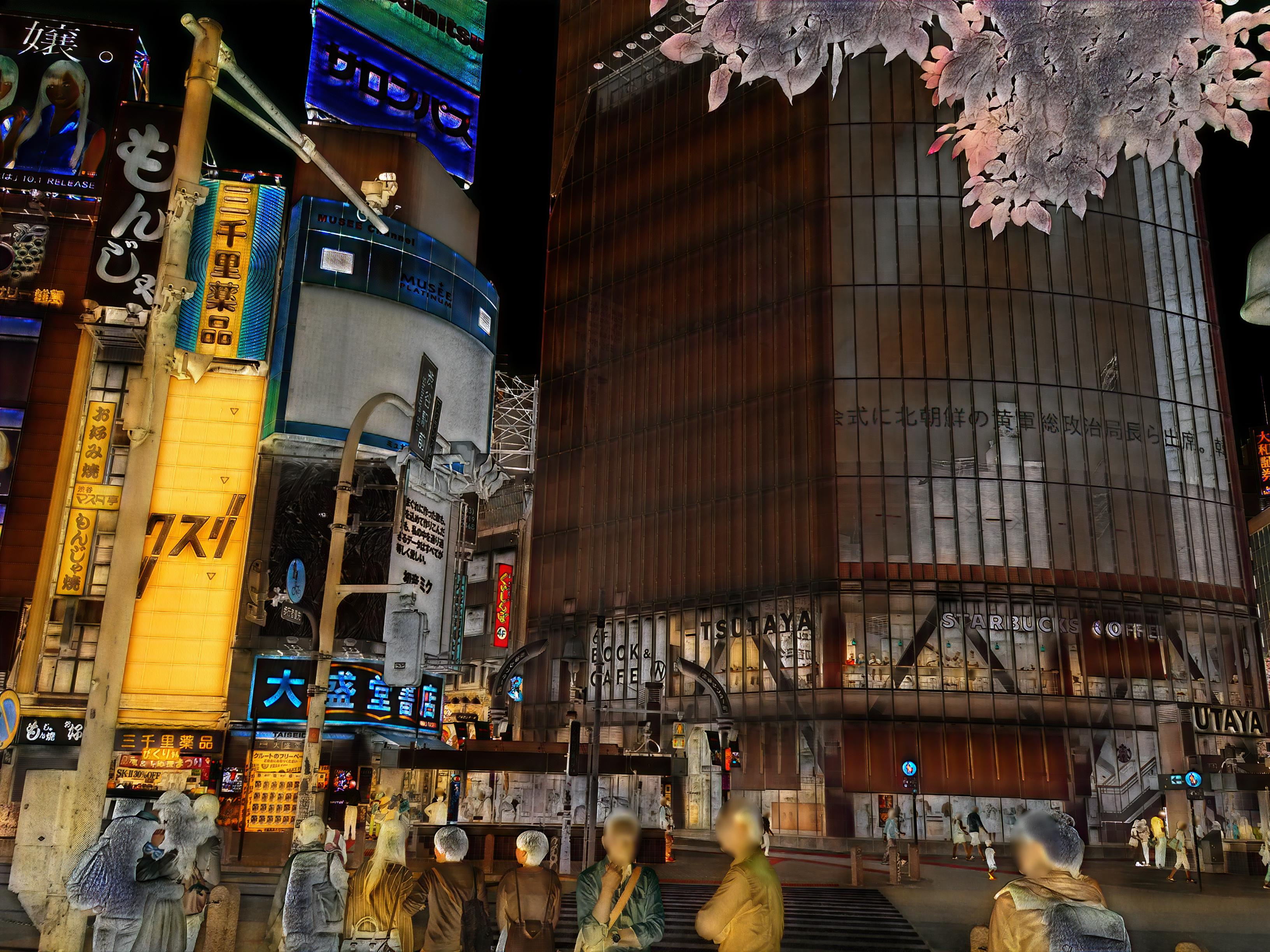}} \,
    \subfloat{\includegraphics[width=\examplesize\linewidth]{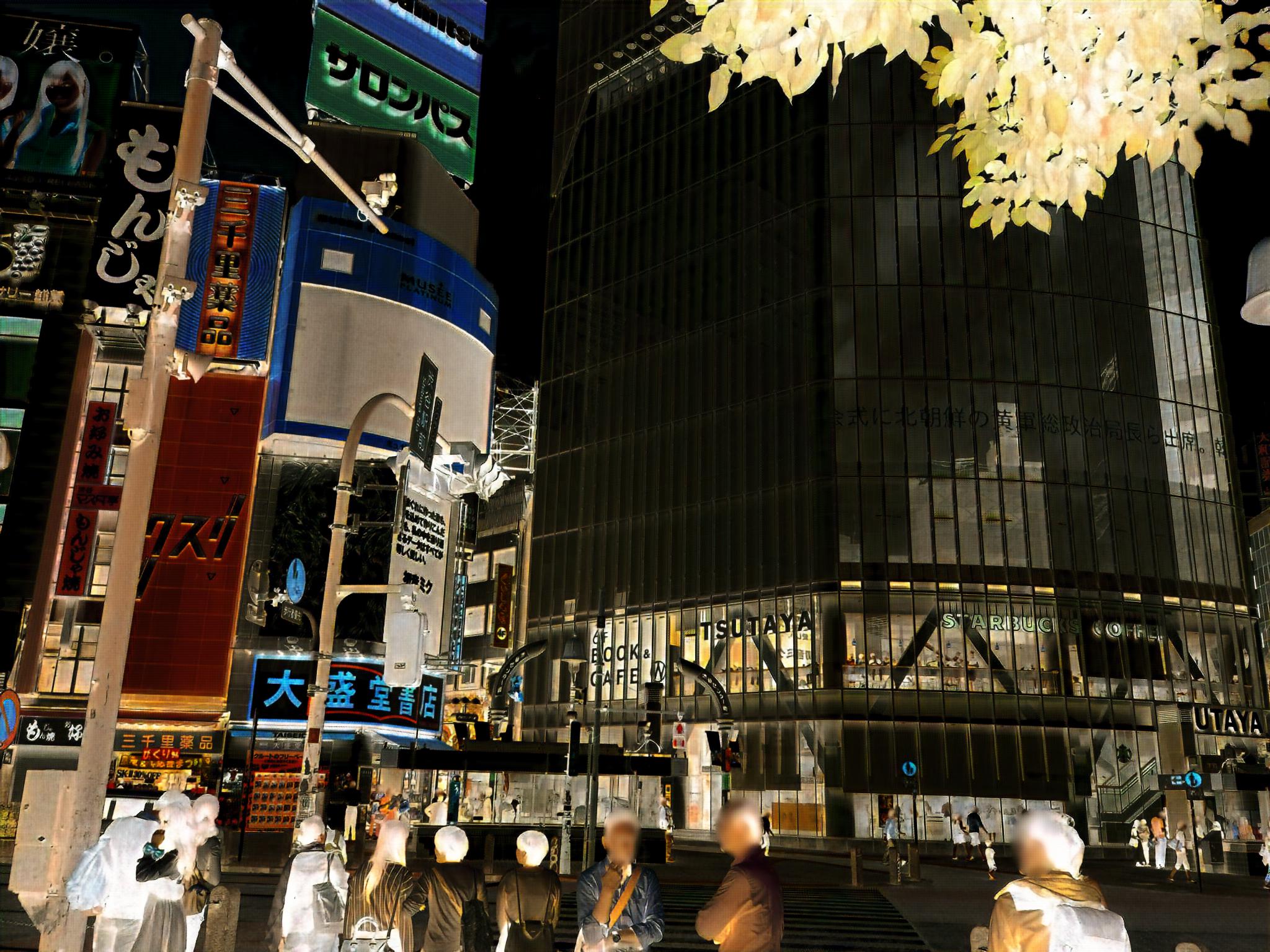}} \,
    \\[0.1em]
    \subfloat{\includegraphics[width=\examplesize\linewidth]{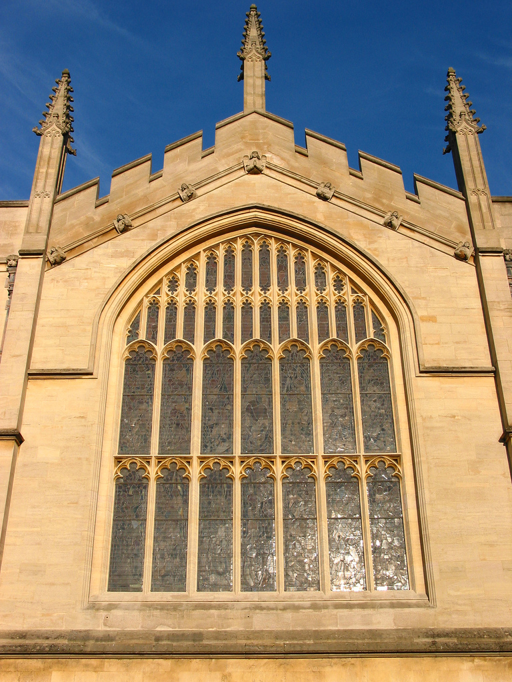}} \,
    \subfloat{\includegraphics[width=\examplesize\linewidth]{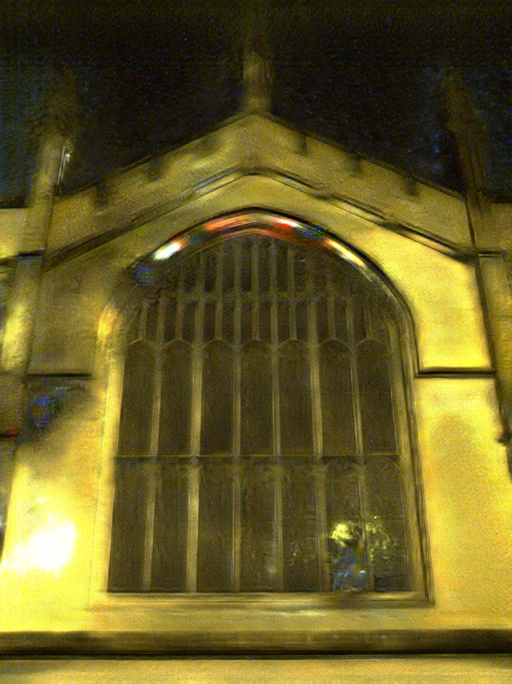}} \,
    \subfloat{\includegraphics[width=\examplesize\linewidth]{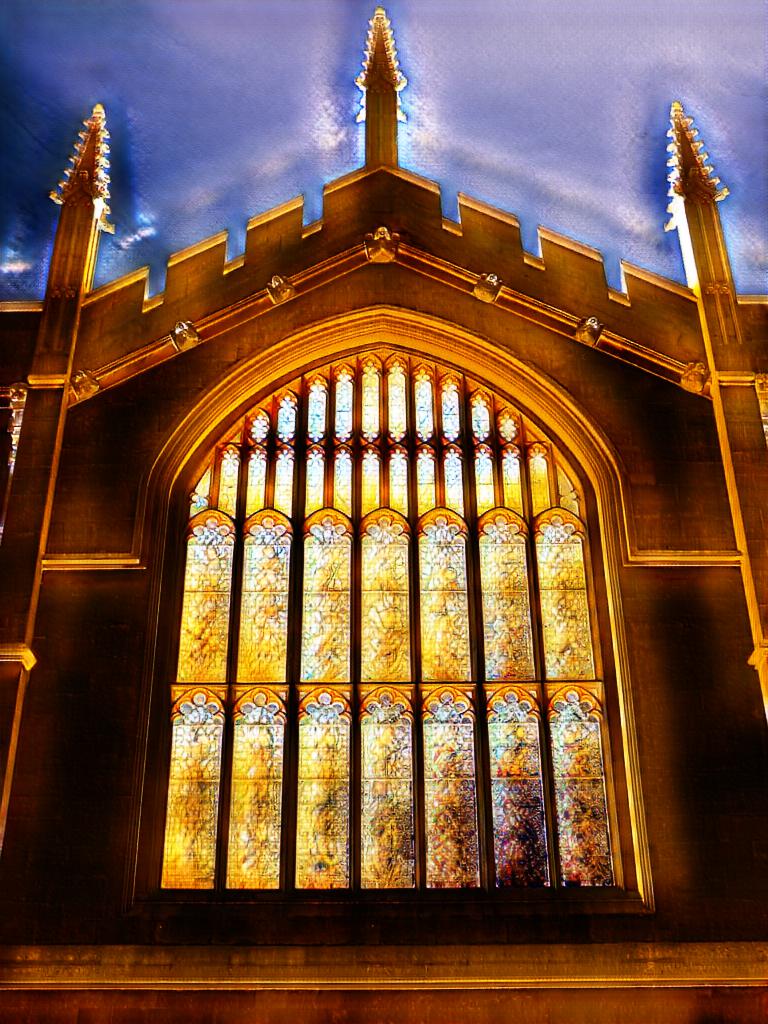}} \,
    \subfloat{\includegraphics[width=\examplesize\linewidth]{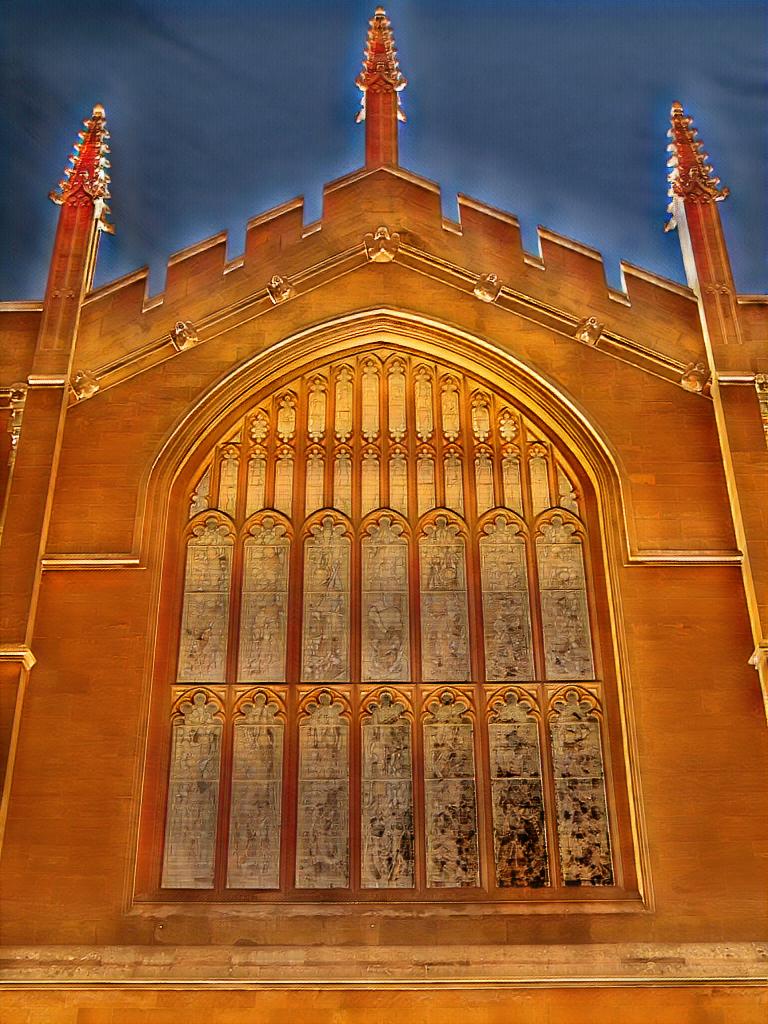}} \,
    \subfloat{\includegraphics[width=\examplesize\linewidth]{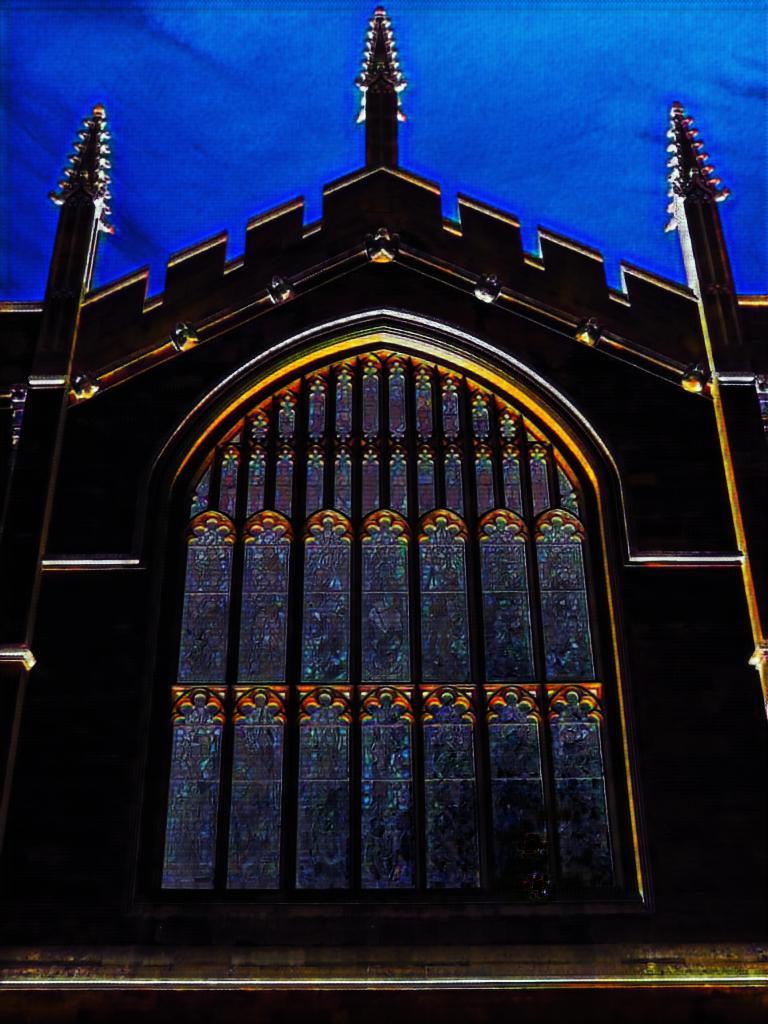}} \,
    \\[0.1em]
    \subfloat{\includegraphics[width=\examplesize\linewidth]{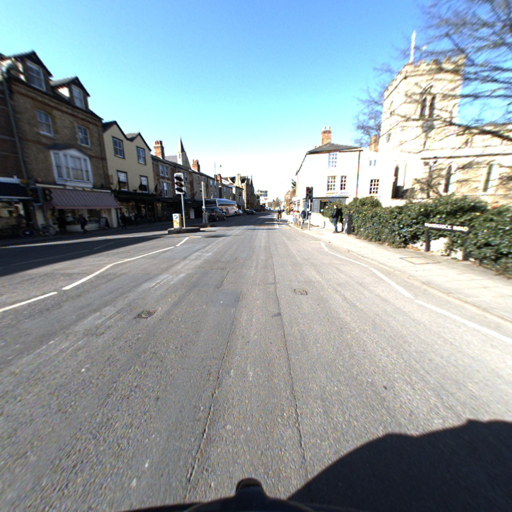}} \,
    \subfloat{\includegraphics[width=\examplesize\linewidth]{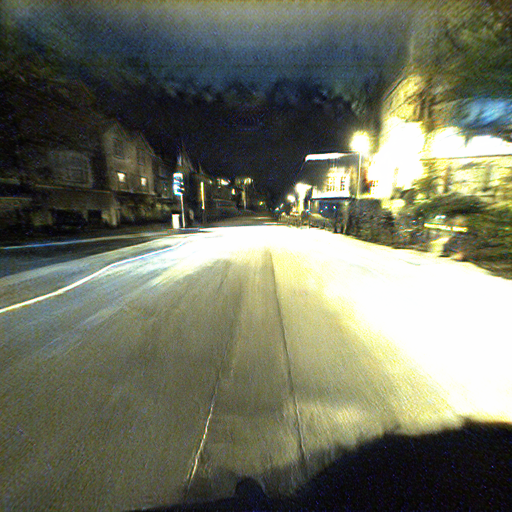}} \,
    \subfloat{\includegraphics[width=\examplesize\linewidth]{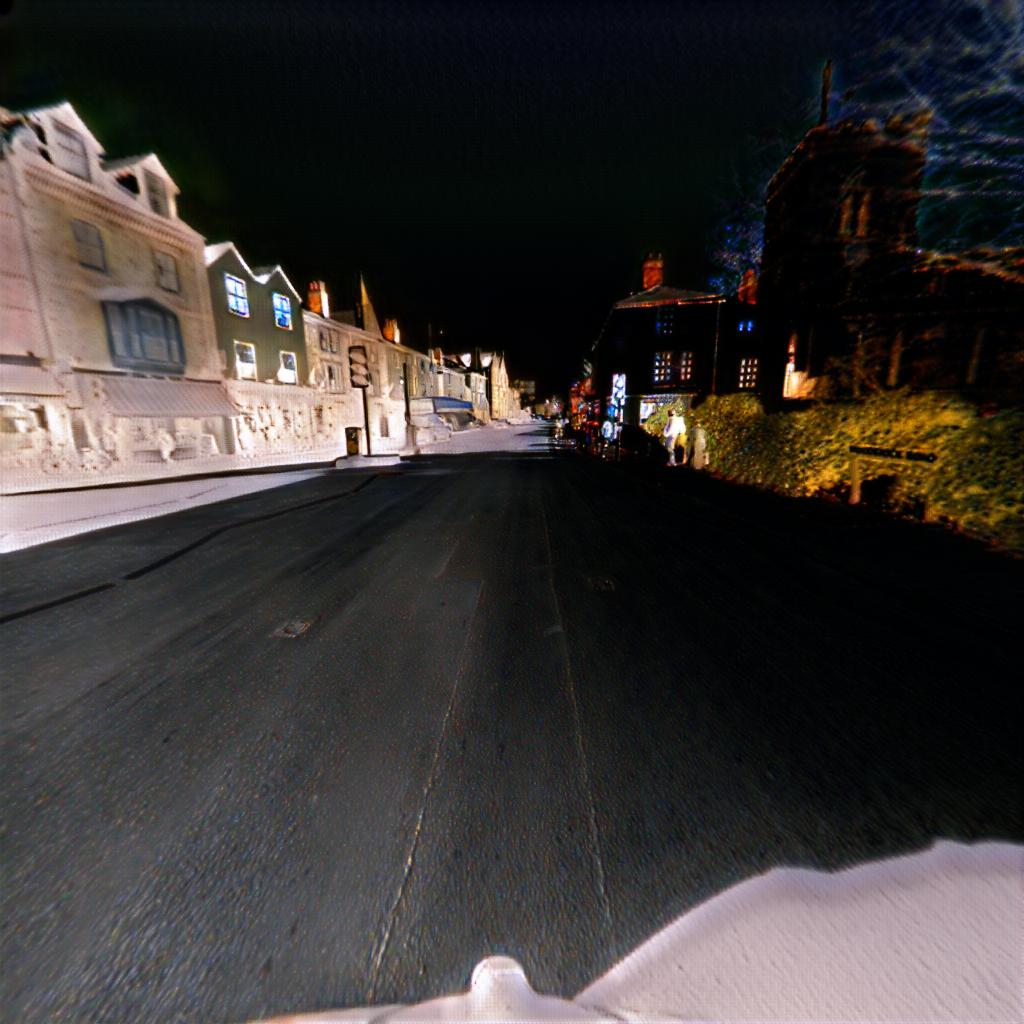}} \,
    \subfloat{\includegraphics[width=\examplesize\linewidth]{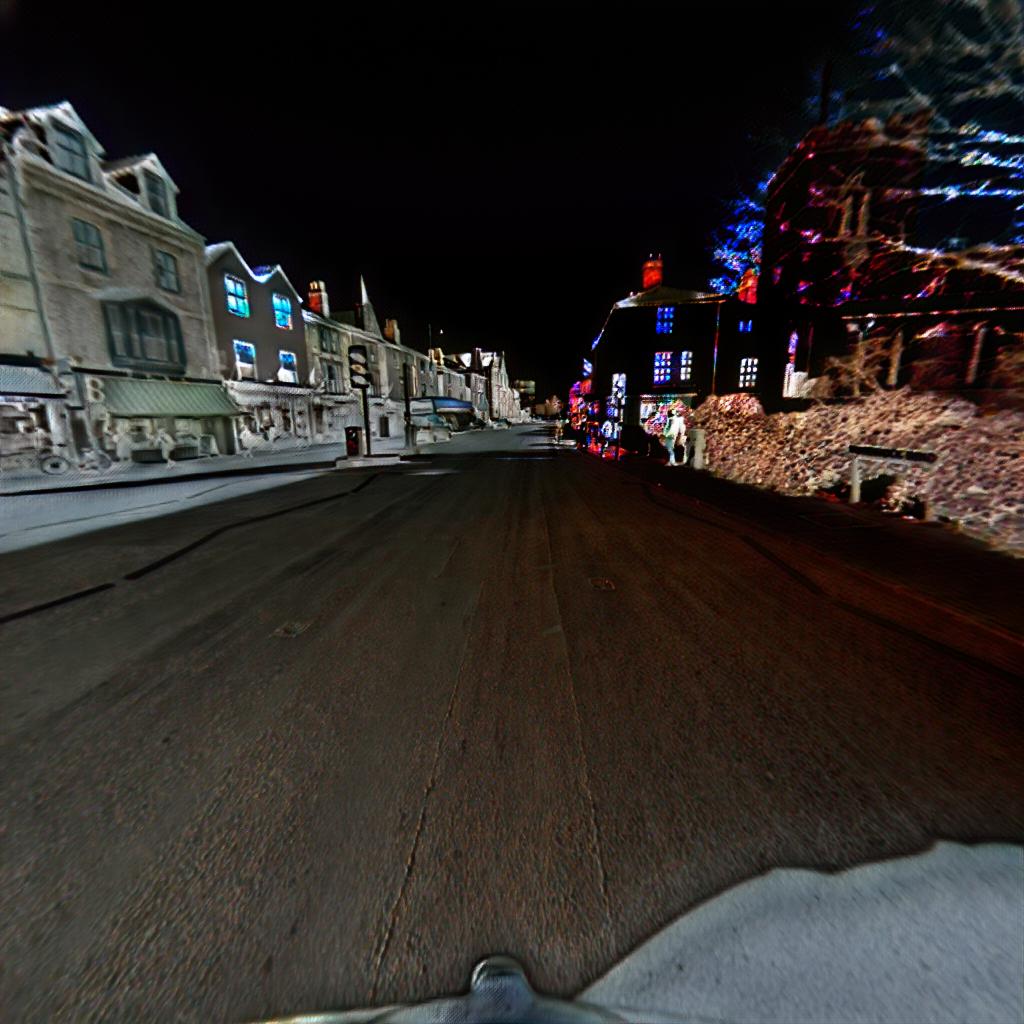}} \,
    \subfloat{\includegraphics[width=\examplesize\linewidth]{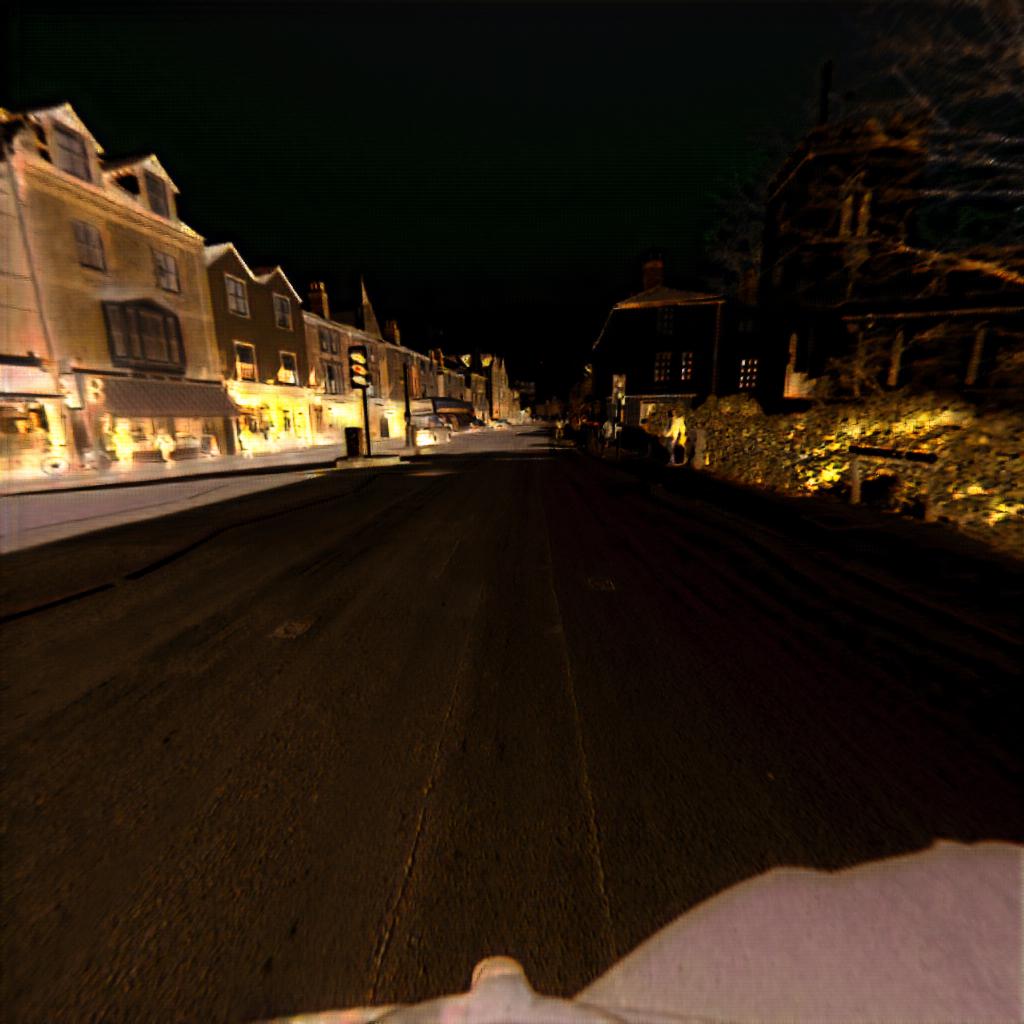}} \,
    \\[1.5em]
    (continues)
\end{figure*}

\begin{figure*}
% \ContinuedFloat 
    \newcommand{\examplesize}{.187}
    \centering
    \begin{tabularx}{\textwidth}{YYYYYY}
    Day original \, & CyEDA BDD & CyEDA (tuned) & CUT & DRIT
    \end{tabularx}
    \\[0.1em]
    \subfloat{\includegraphics[width=\examplesize\linewidth]{img/day2night/sfm120k_6bfa80c630c2cb408c1323009dedd3c1_orig.png}} \,
    \subfloat{\includegraphics[width=\examplesize\linewidth]{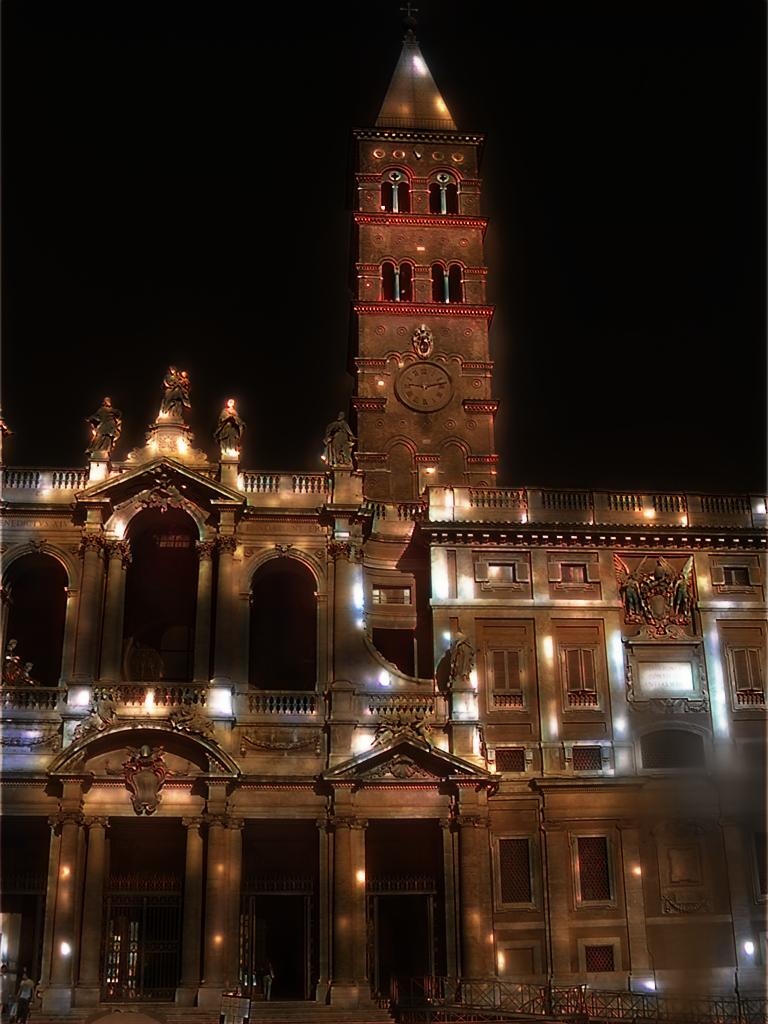}} \,
    \subfloat{\includegraphics[width=\examplesize\linewidth]{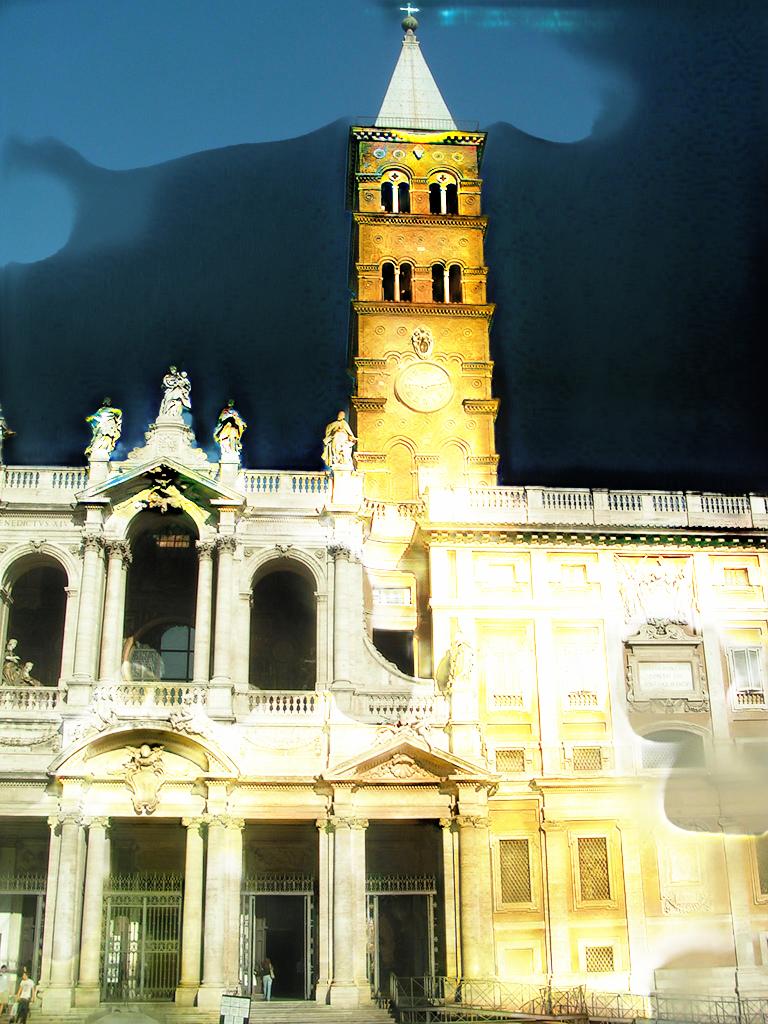}} \,
    \subfloat{\includegraphics[width=\examplesize\linewidth]{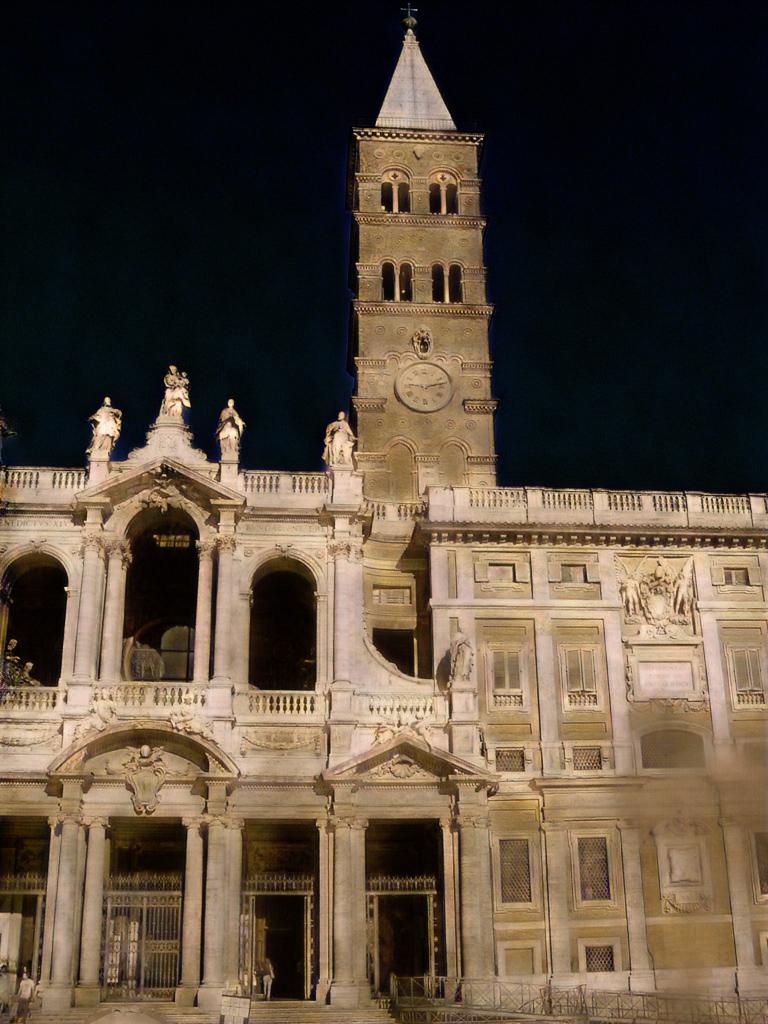}} \,
    \subfloat{\includegraphics[width=\examplesize\linewidth]{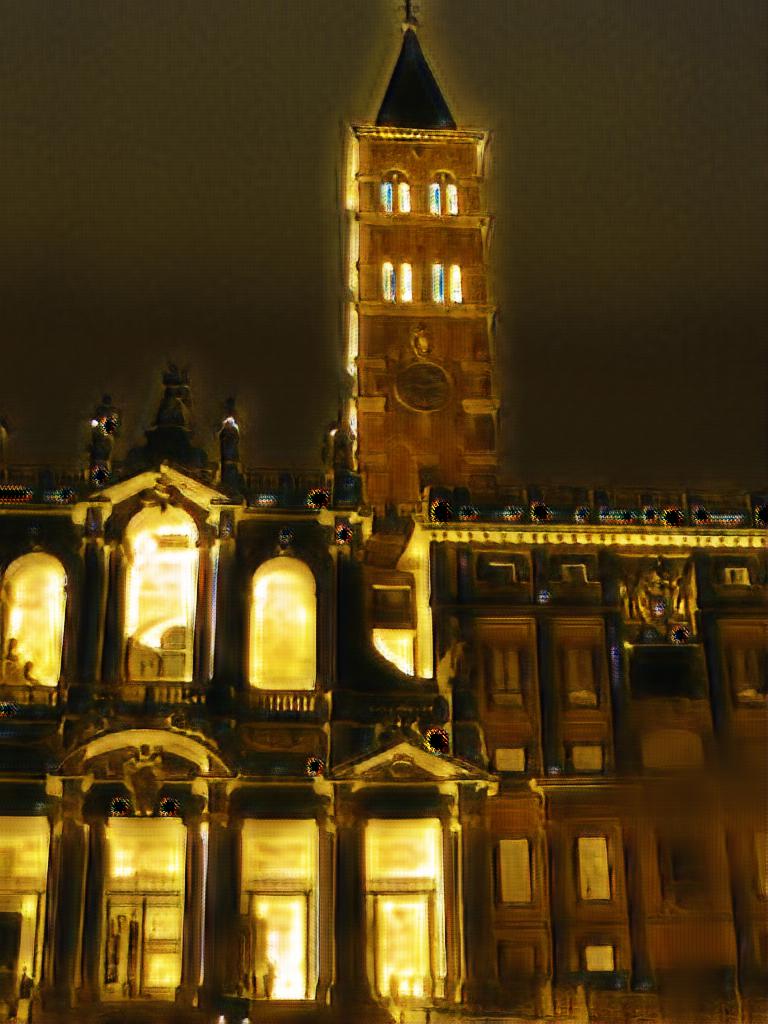}} \,
    \\[0.1em]
    \subfloat{\includegraphics[width=\examplesize\linewidth]{img/day2night/aachen_2010-10-30_17-47-25_73_orig.png}} \,
    \subfloat{\includegraphics[width=\examplesize\linewidth]{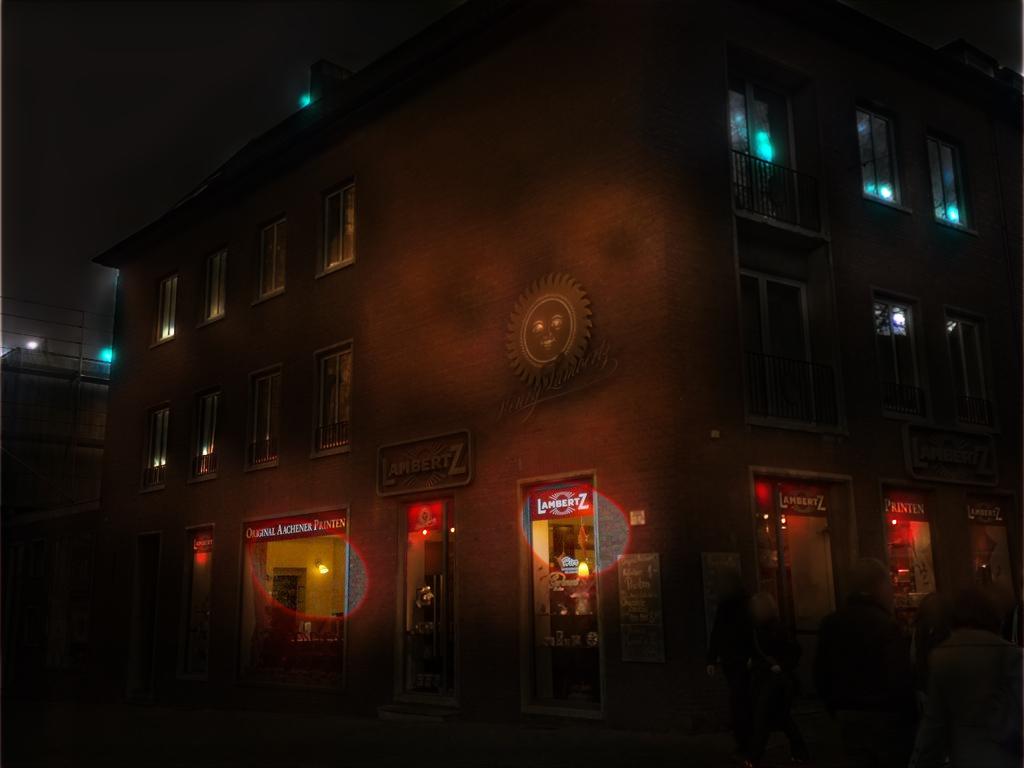}} \,
    \subfloat{\includegraphics[width=\examplesize\linewidth]{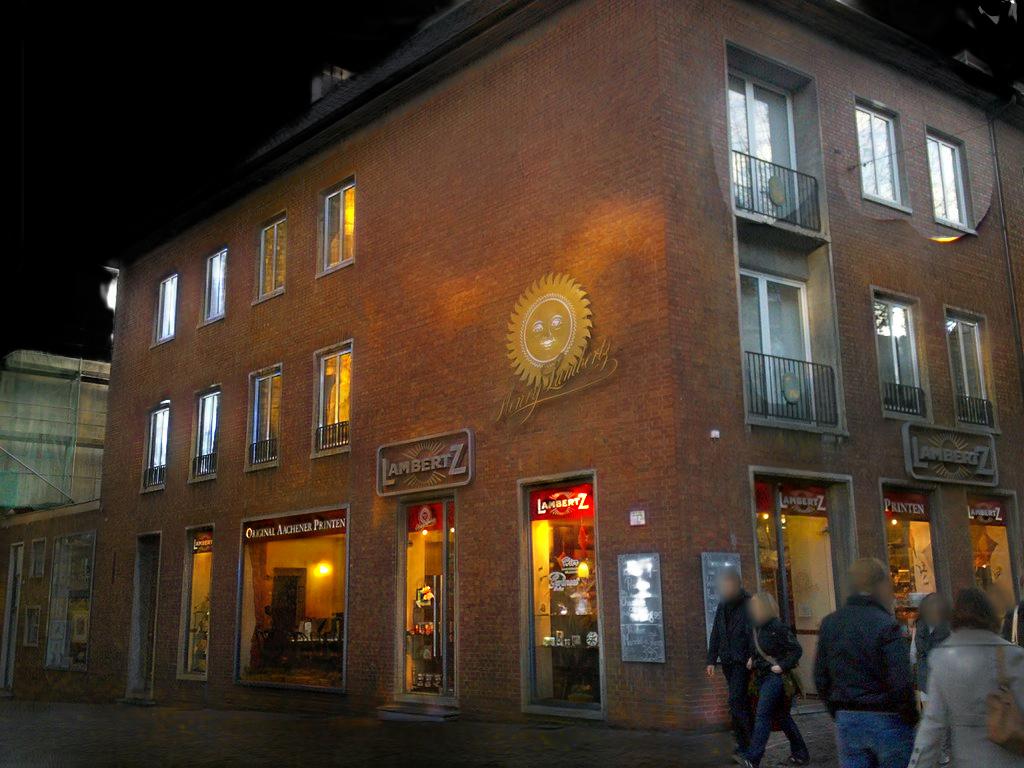}} \,
    \subfloat{\includegraphics[width=\examplesize\linewidth]{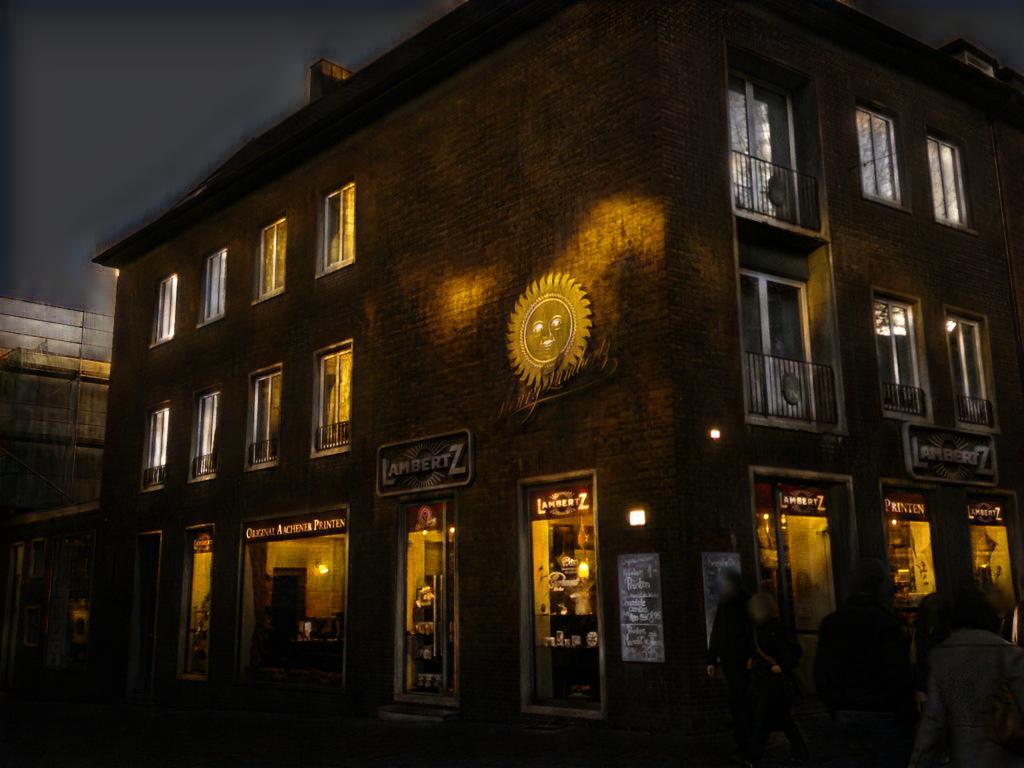}} \,
    \subfloat{\includegraphics[width=\examplesize\linewidth]{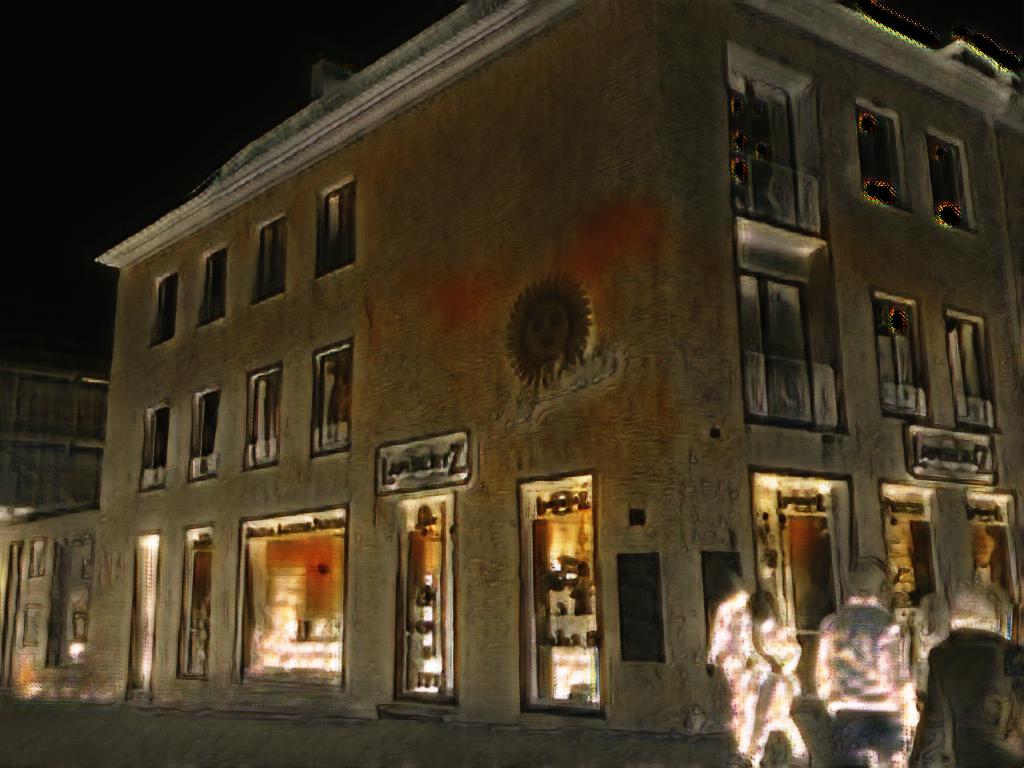}} \,
    \\[0.1em]
    \subfloat{\includegraphics[width=\examplesize\linewidth]{img/day2night/tokio_00001_orig.png}} \,
    \subfloat{\includegraphics[width=\examplesize\linewidth]{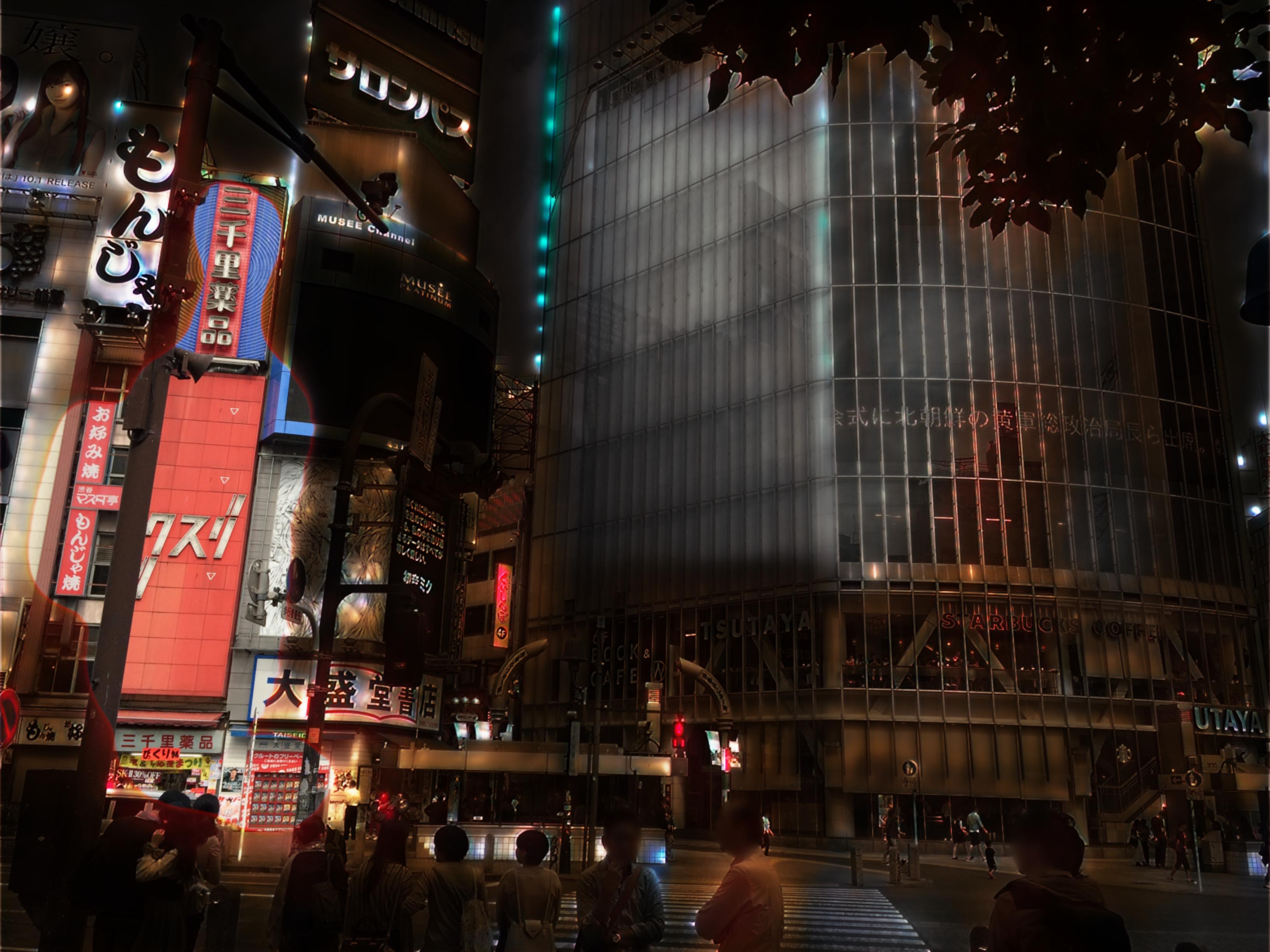}} \,
    \subfloat{\includegraphics[width=\examplesize\linewidth]{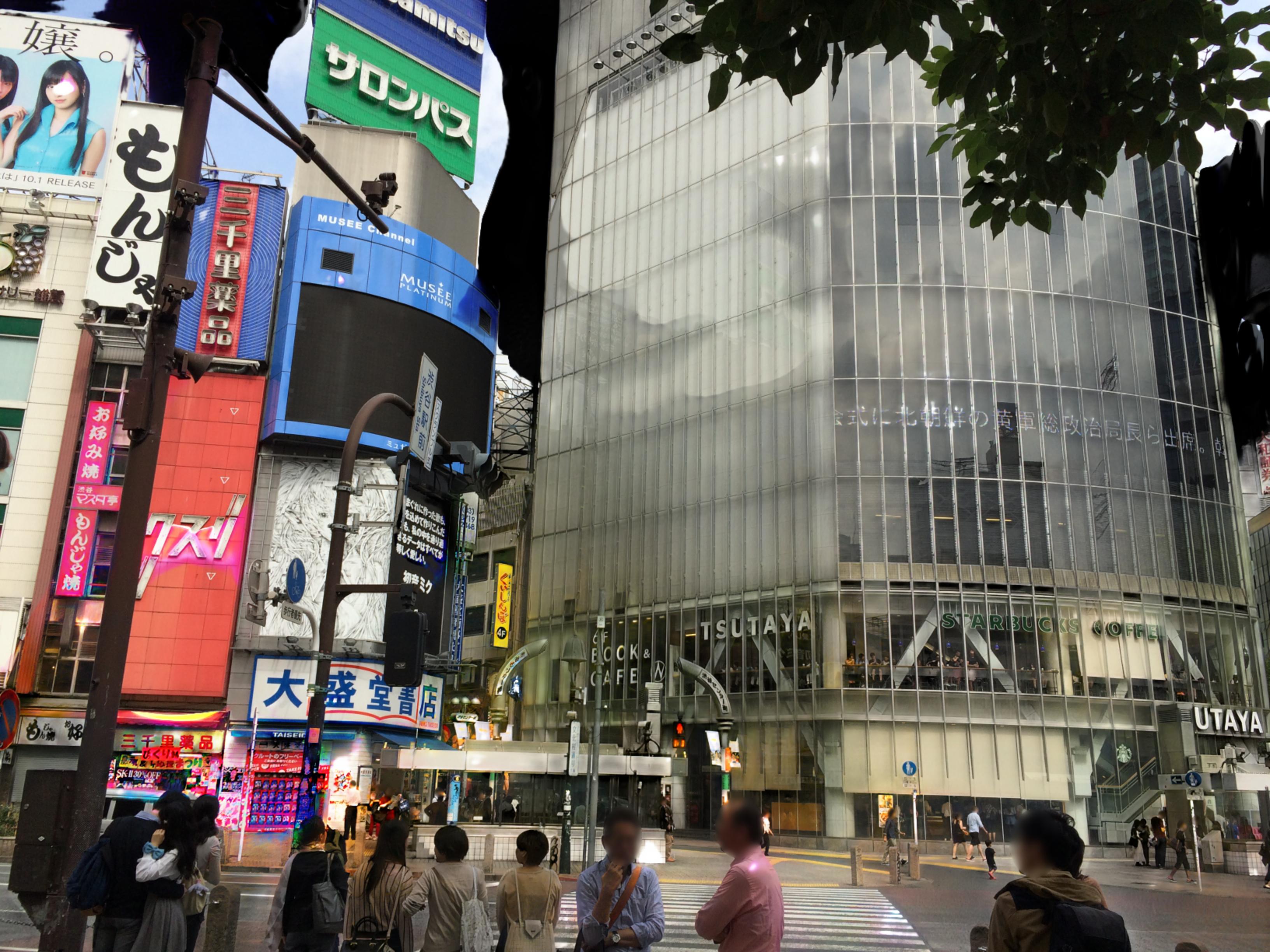}} \,
    \subfloat{\includegraphics[width=\examplesize\linewidth]{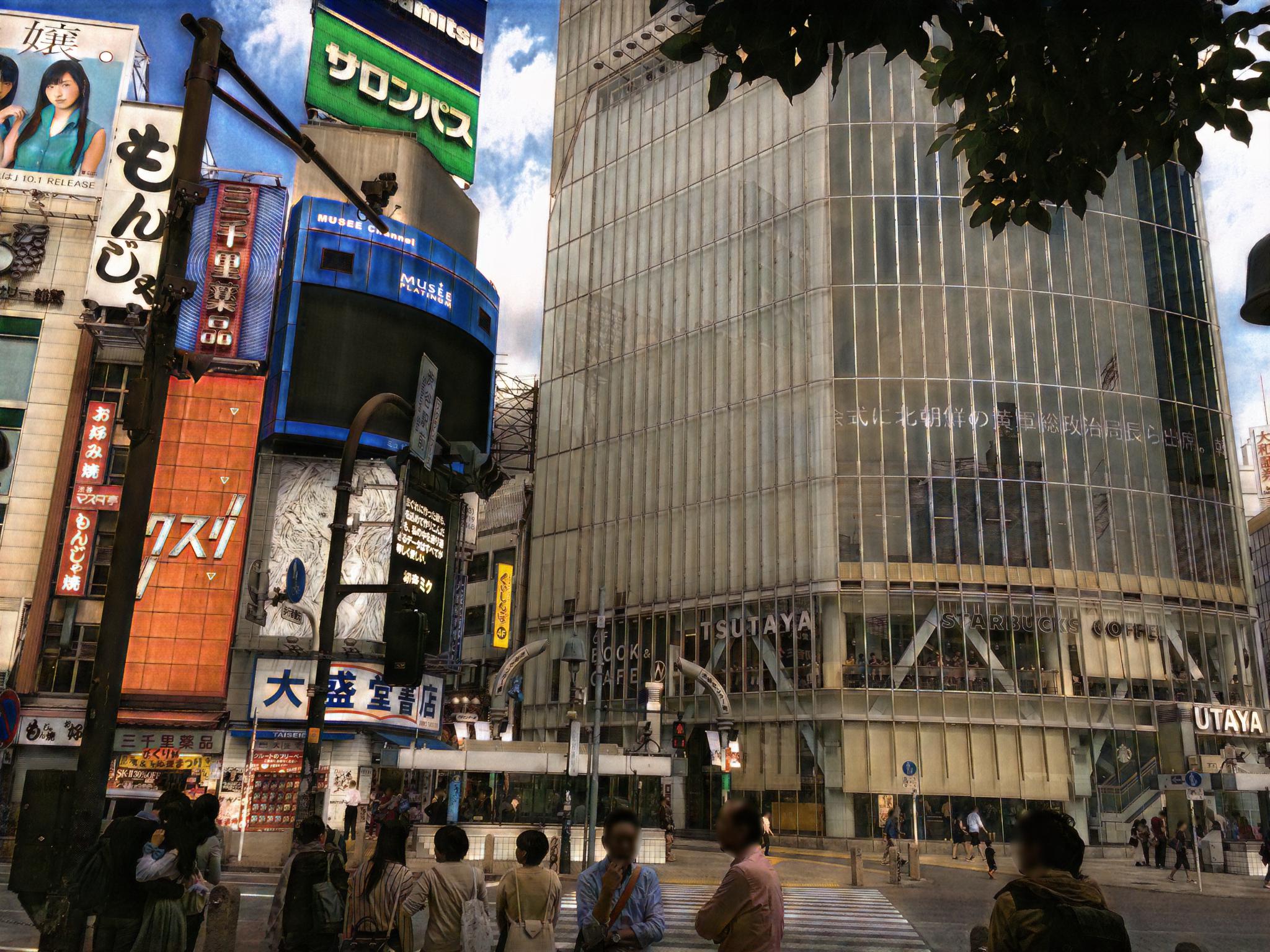}} \,
    \subfloat{\includegraphics[width=\examplesize\linewidth]{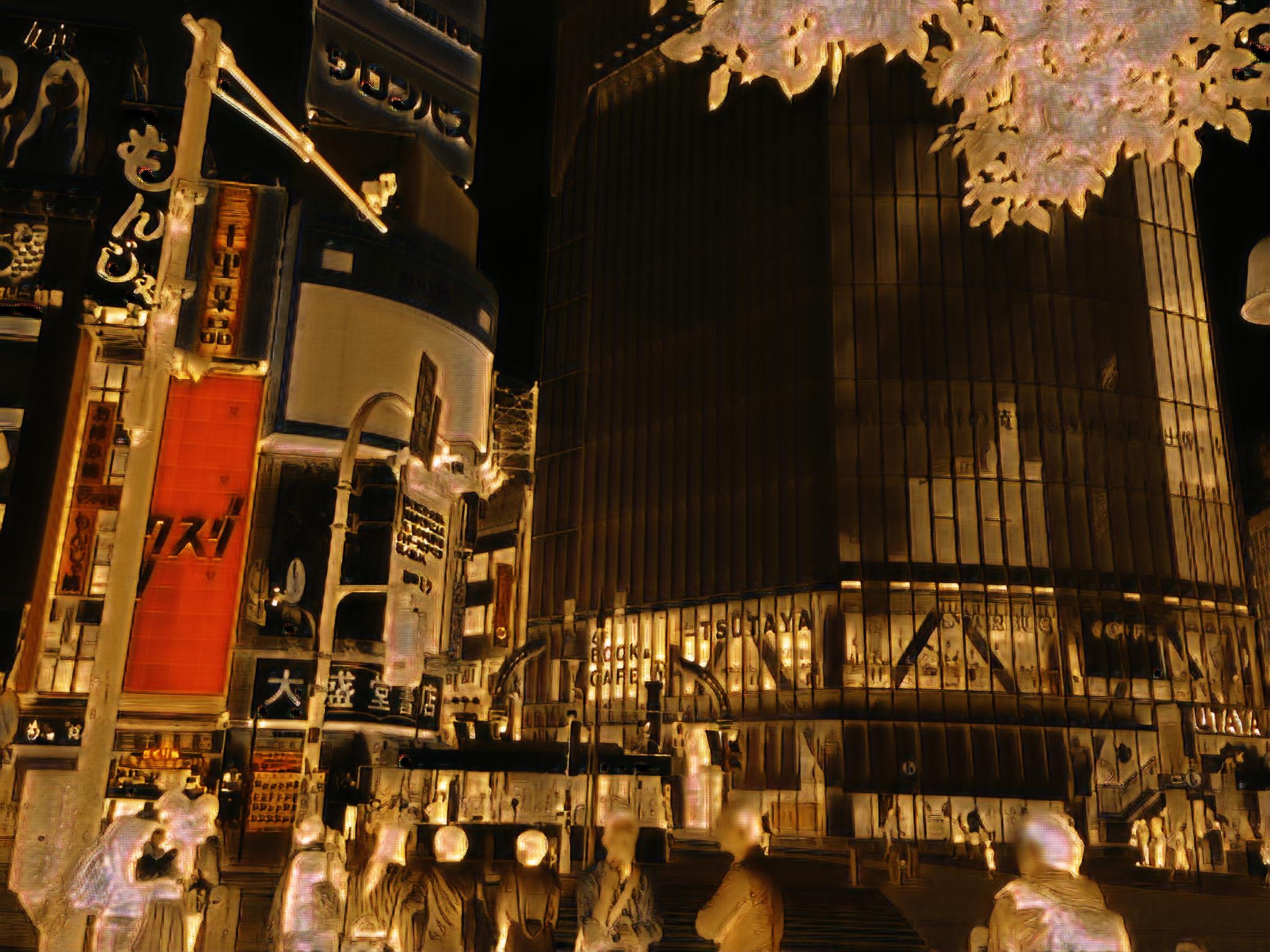}} \,
    \\[0.1em]
    \subfloat{\includegraphics[width=\examplesize\linewidth]{img/day2night/oxford_all_souls_000003_orig.png}} \,
    \subfloat{\includegraphics[width=\examplesize\linewidth]{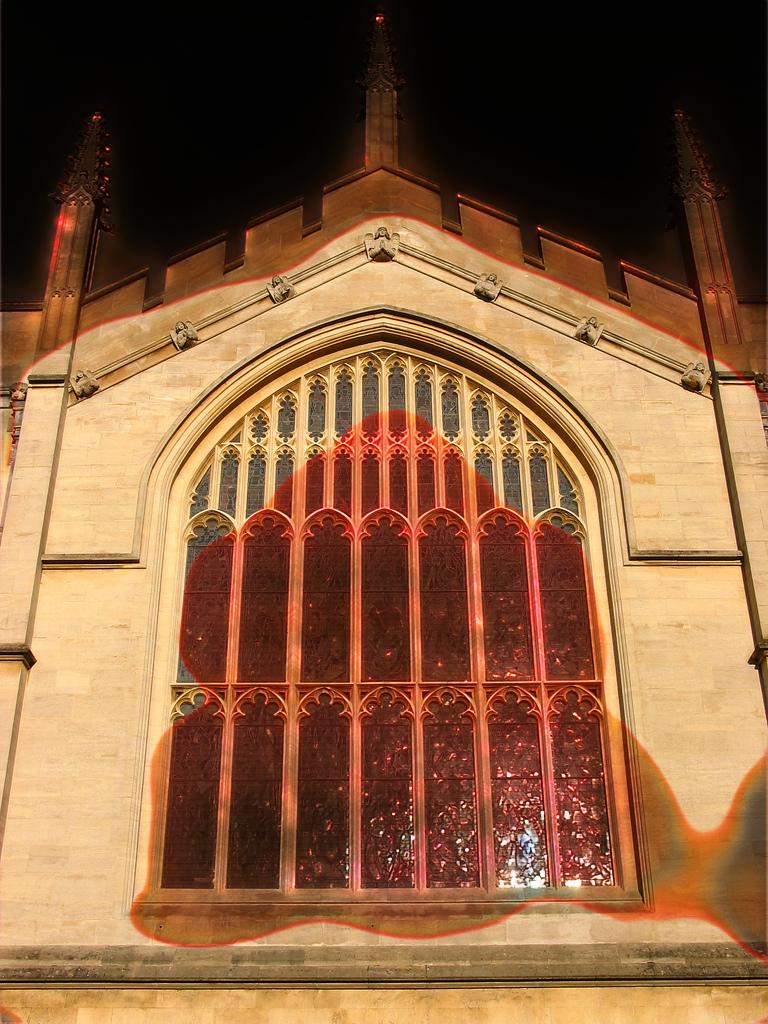}} \,
    \subfloat{\includegraphics[width=\examplesize\linewidth]{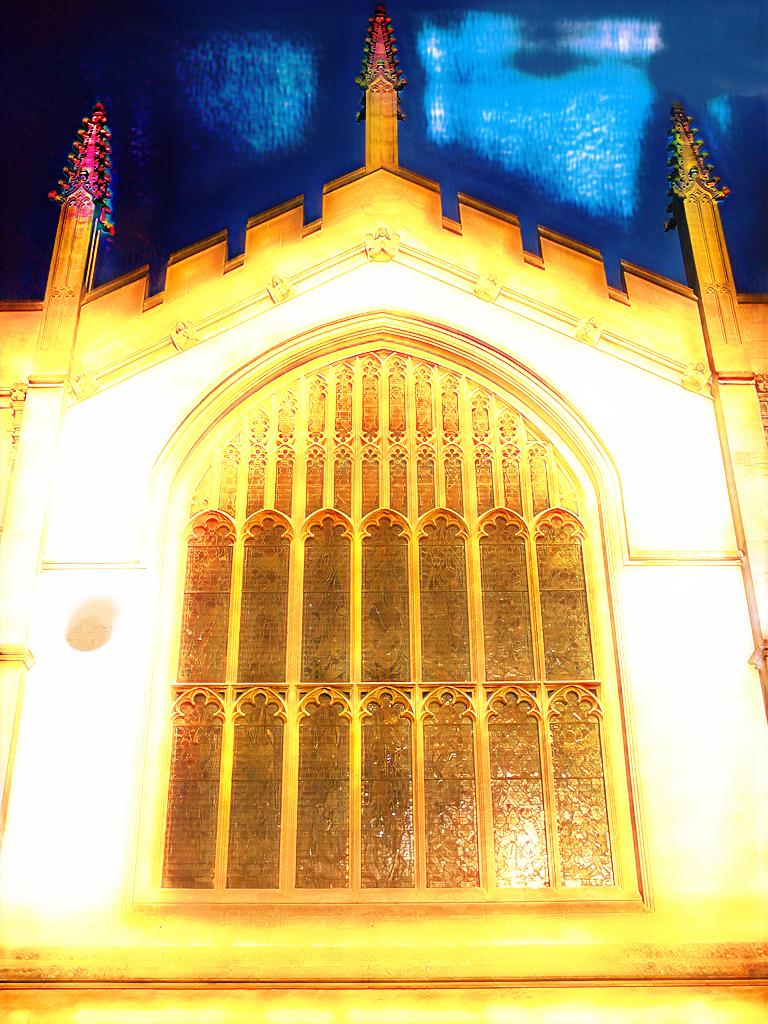}} \,
    \subfloat{\includegraphics[width=\examplesize\linewidth]{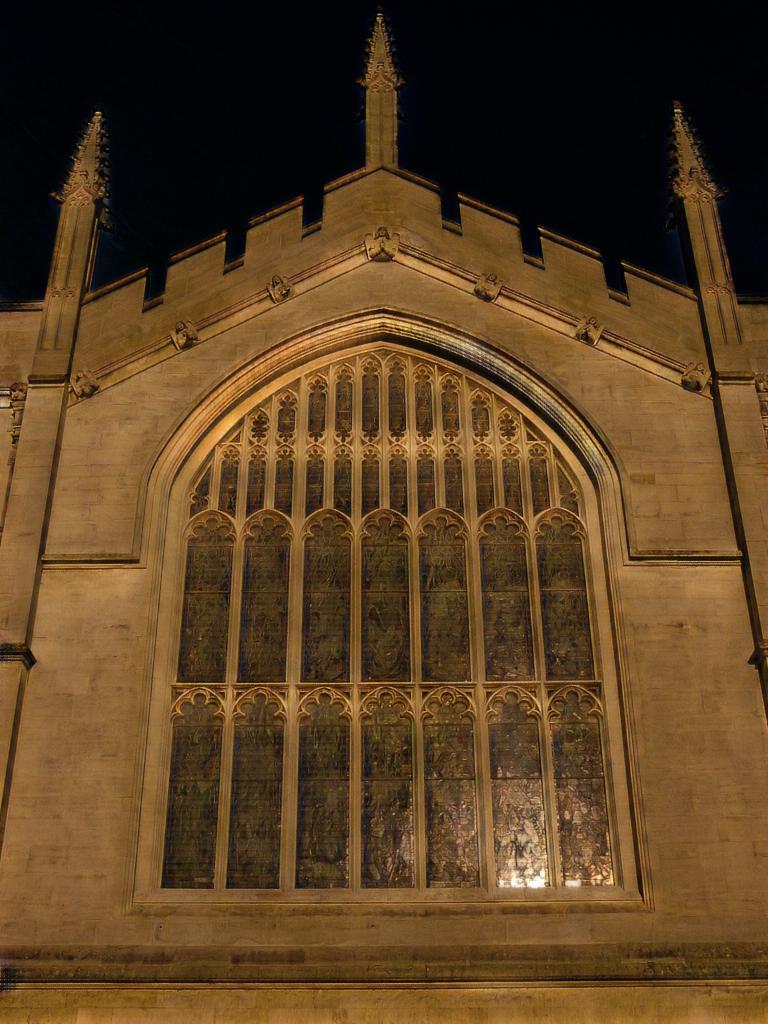}} \,
    \subfloat{\includegraphics[width=\examplesize\linewidth]{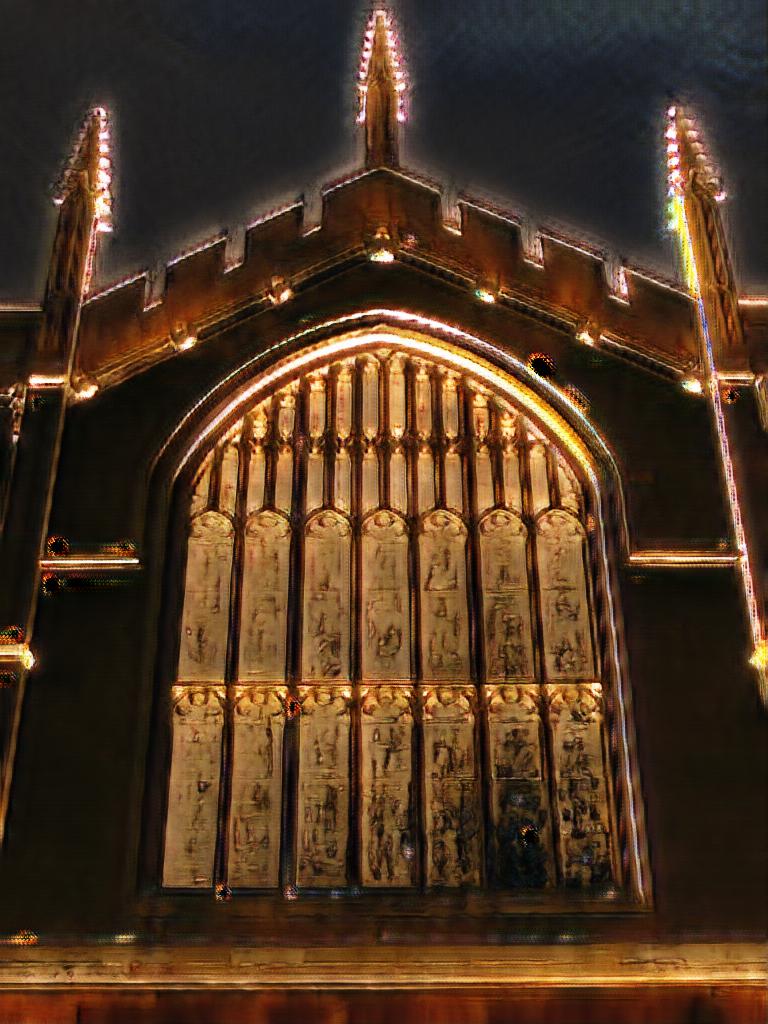}} \,
    \\[0.1em]
    \subfloat{\includegraphics[width=\examplesize\linewidth]{img/day2night/robotcar_1425997185269644.rear_orig.png}} \,
    \subfloat{\includegraphics[width=\examplesize\linewidth]{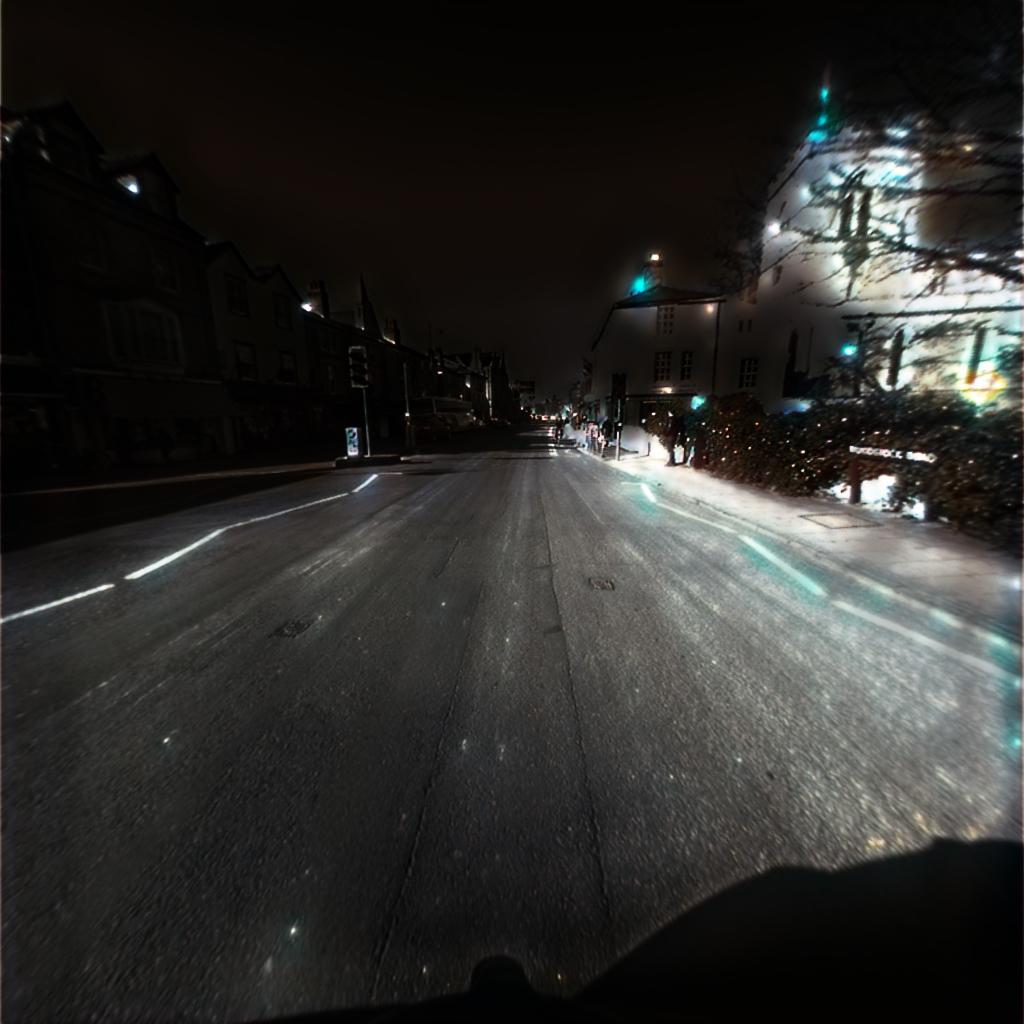}} \,
    \subfloat{\includegraphics[width=\examplesize\linewidth]{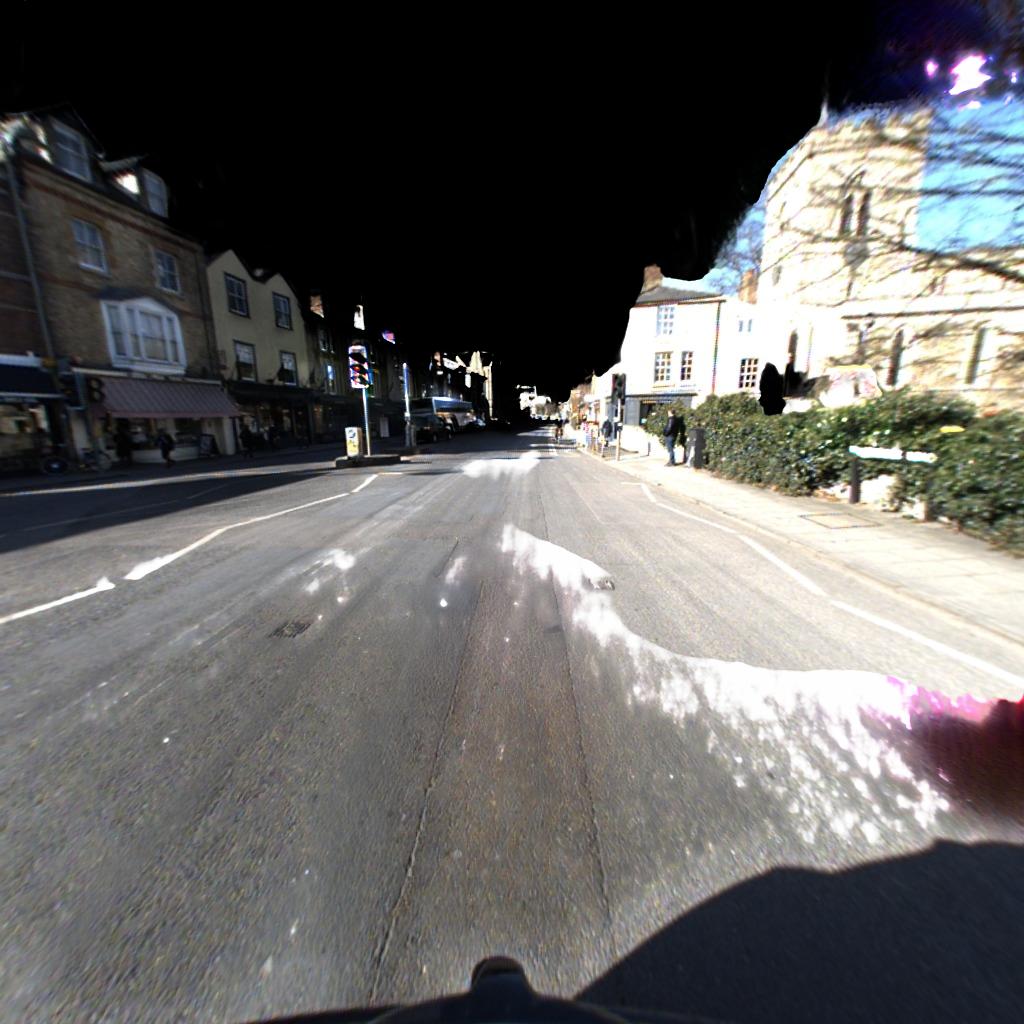}} \,
    \subfloat{\includegraphics[width=\examplesize\linewidth]{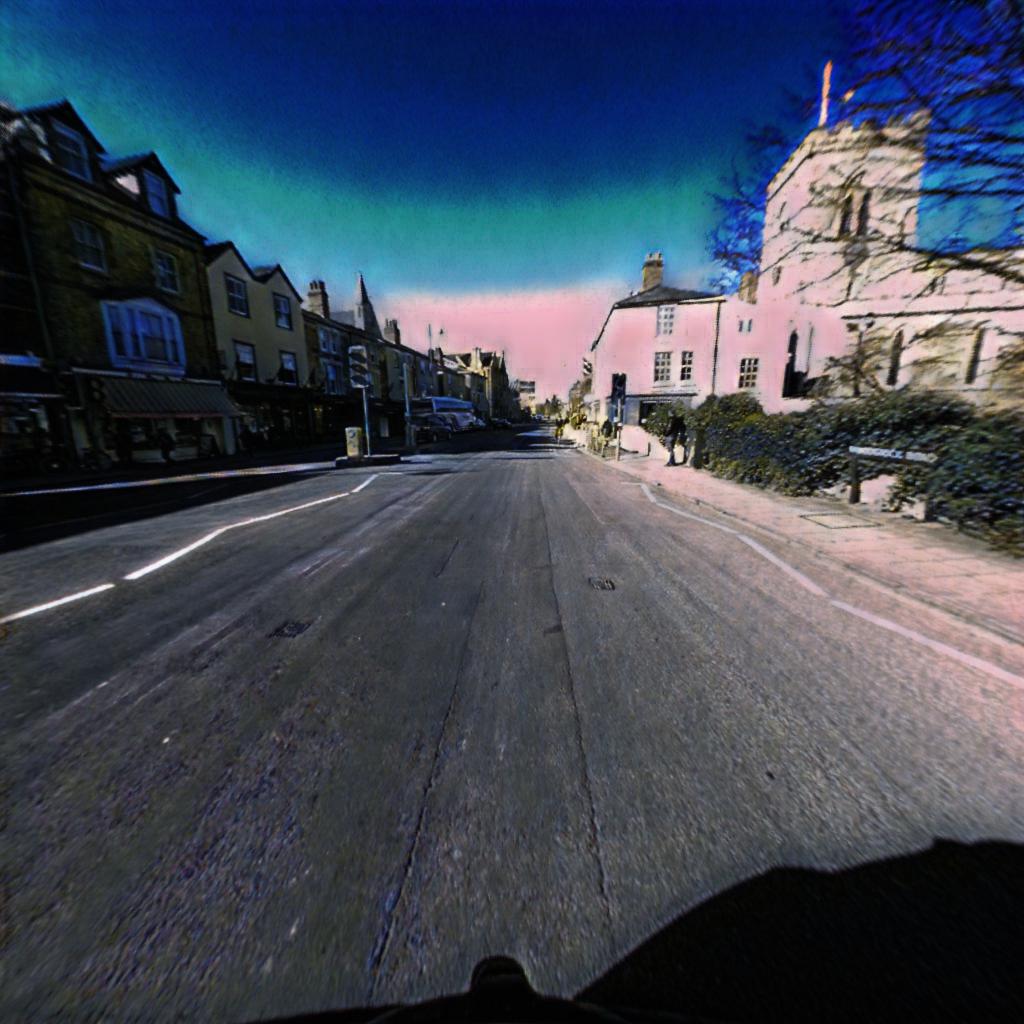}} \,
    \subfloat{\includegraphics[width=\examplesize\linewidth]{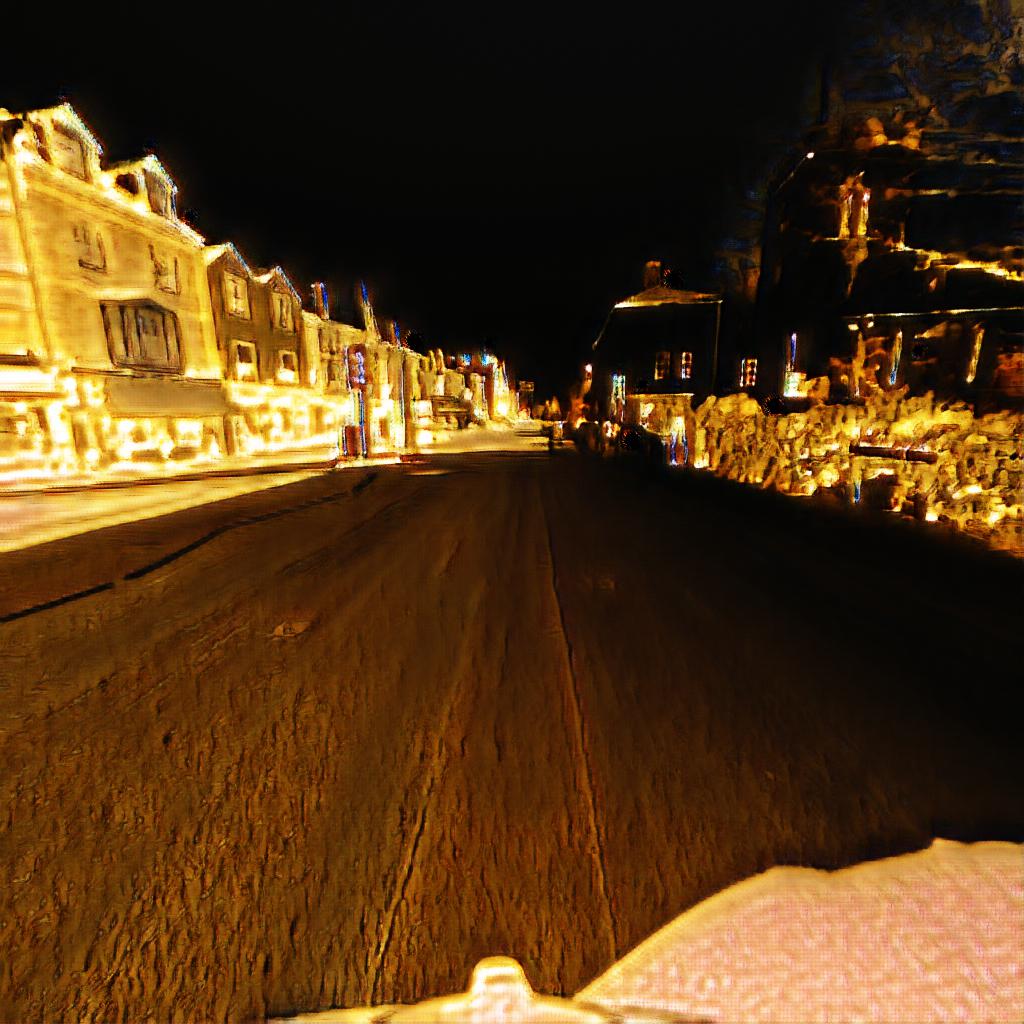}} \,
    \caption{Examples of day images translated into the night domain by different day $\rightarrow$ night generators. The columns correspond to (left-to-right): the original image, ToDayGAN generator~\cite{toDayGAN}, our generators \hedn GAN, \rcfn GAN, and CycleGAN, the original image, CyEDA~\cite{beh2022cyeda} generator trained on BDD dataset~\cite{yu2020bdd100k}, CyEDA generator trained on \textit{SfM}120k~\cite{radenovic2018fine} and tuned by us, and our CUT, and DRIT generators. The rows show example images from different datasets (top-to-bottom): \textit{SfM}120k~\cite{radenovic2018fine}, \textit{Aachen}~\cite{Sattler2018CVPR,Sattler2012BMVC}, \textit{Tokyo}~\cite{Torii-CVPR2015}, Oxford~\cite{Radenovic-CVPR18}, and \textit{RobotCar}~\cite{RobotCarDatasetIJRR}, respectively.}
    \label{fig:day_examples}
\end{figure*}

\setcounter{figure}{2}
\begin{figure*}
    \newcommand{\examplesize}{.238}
    \centering
    \begin{tabularx}{0.75\textwidth}{YYY}
    Night original \, & ToDayGAN~\cite{toDayGAN} & CycleGAN
    \end{tabularx}
    \\[0.5em]
    \subfloat{\includegraphics[width=\examplesize\linewidth]{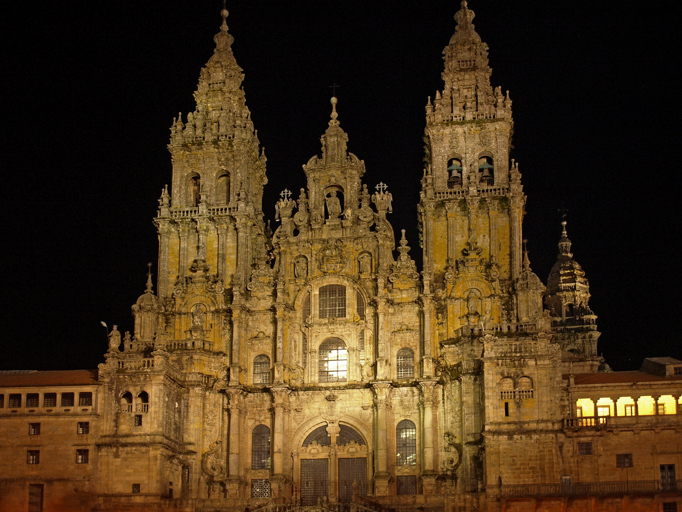}} \,
    \subfloat{\includegraphics[width=\examplesize\linewidth]{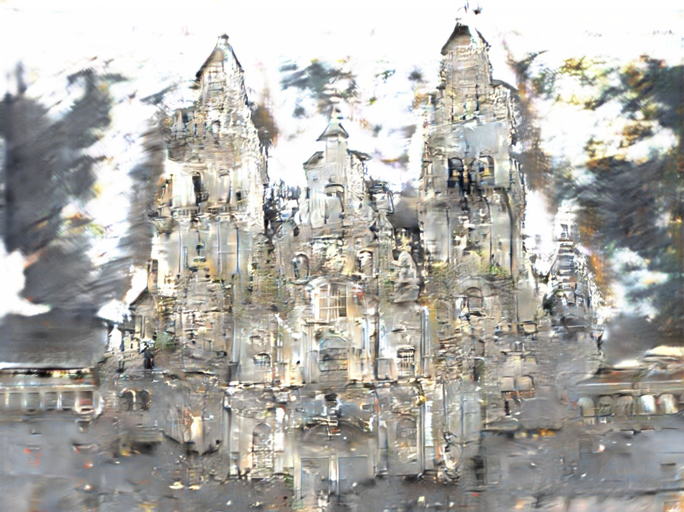}} \,
    \subfloat{\includegraphics[width=\examplesize\linewidth]{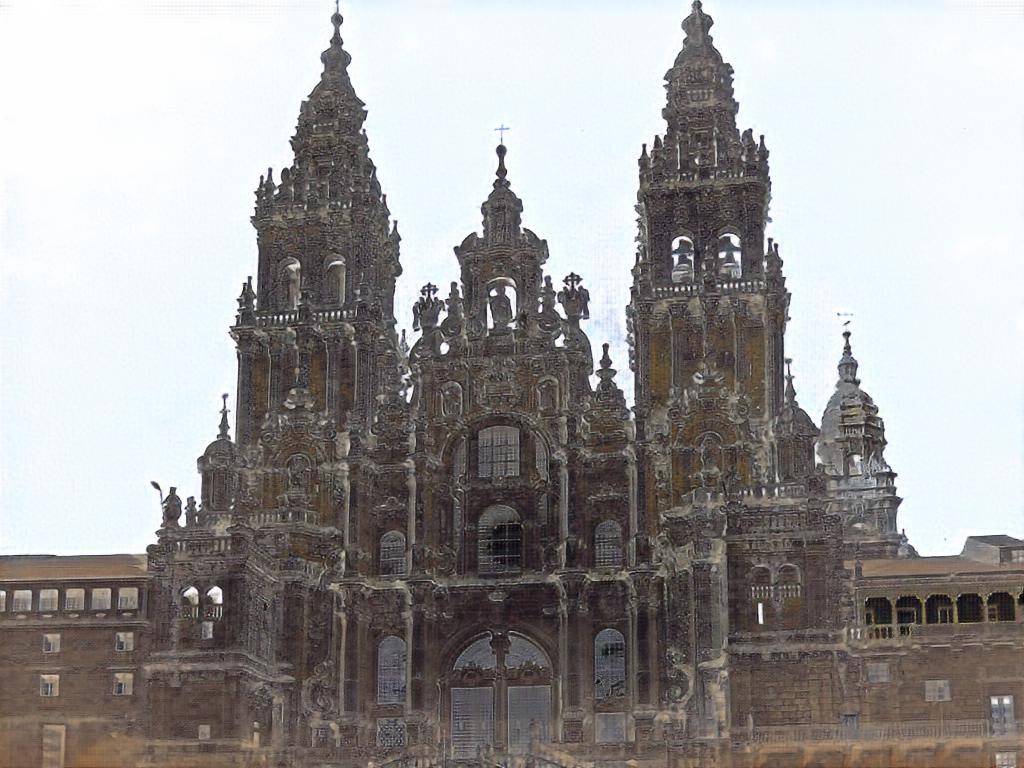}} \,
    \\[0.5em]
    \subfloat{\includegraphics[width=\examplesize\linewidth]{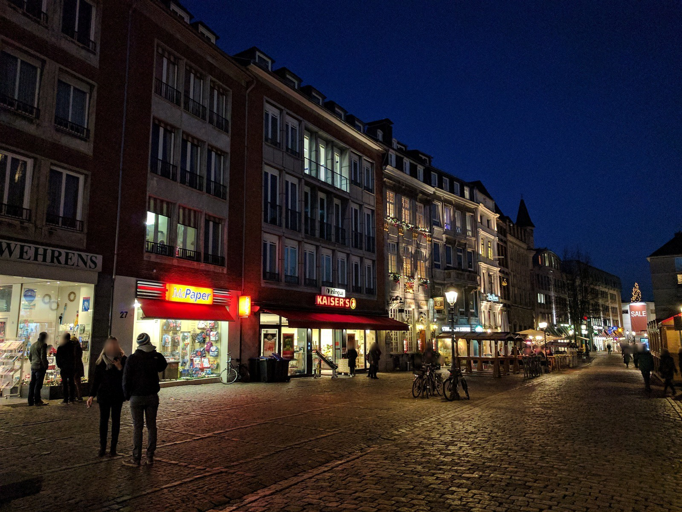}} \,
    \subfloat{\includegraphics[width=\examplesize\linewidth]{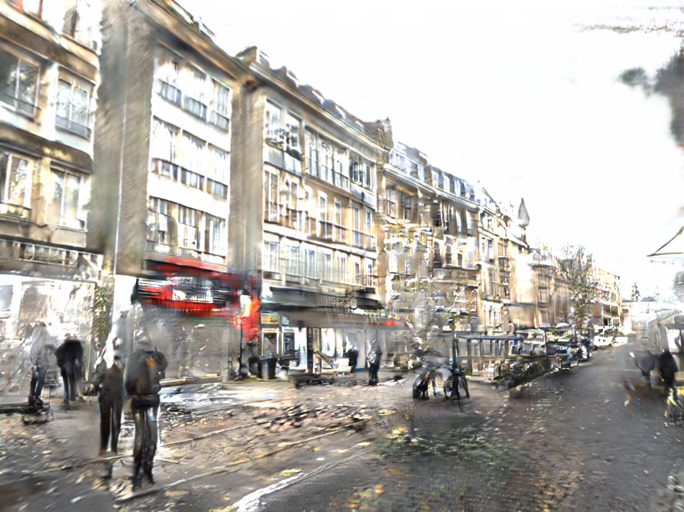}} \,
    \subfloat{\includegraphics[width=\examplesize\linewidth]{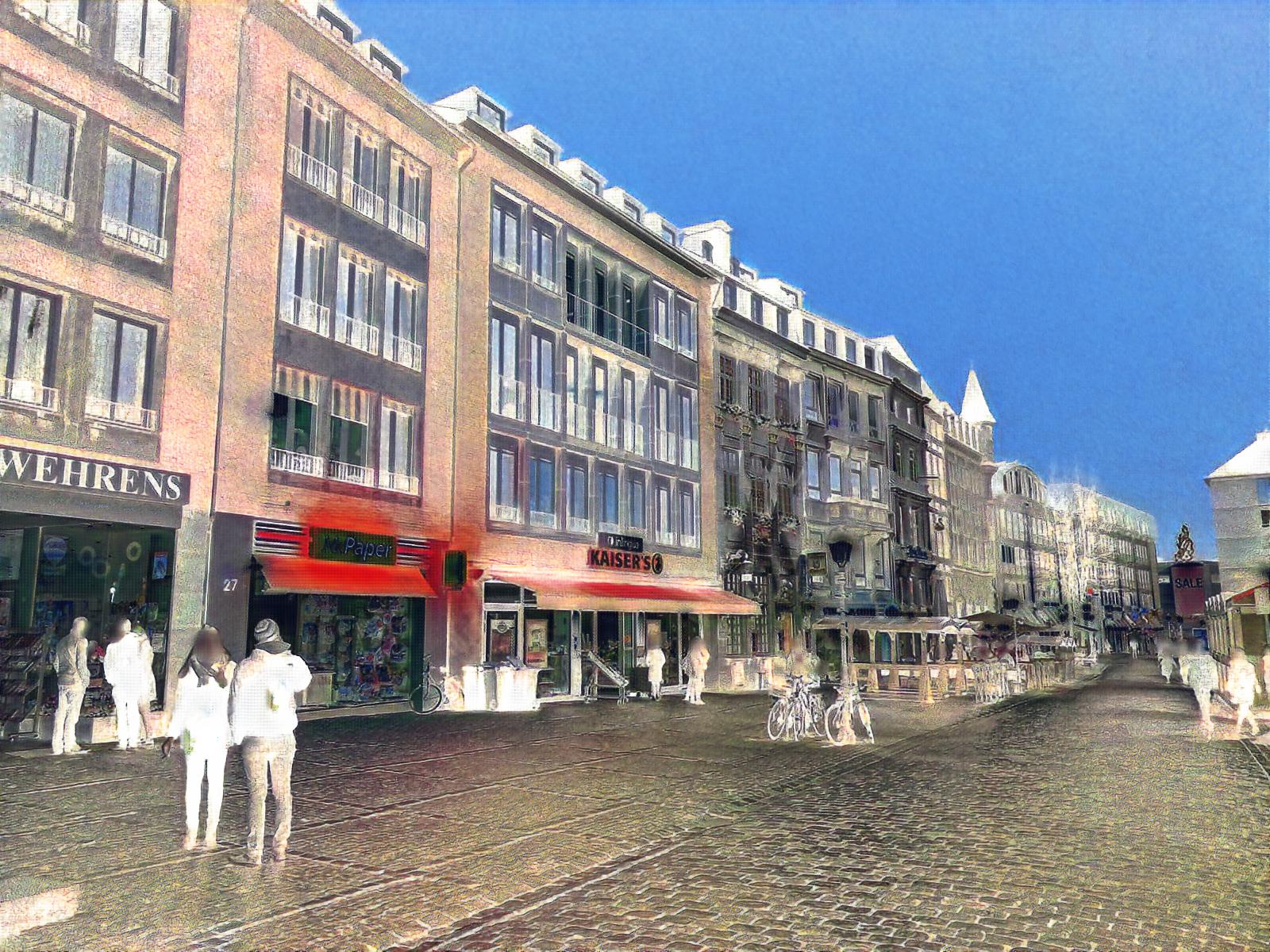}} \,
    \\[0.5em]
    \subfloat{\includegraphics[width=\examplesize\linewidth]{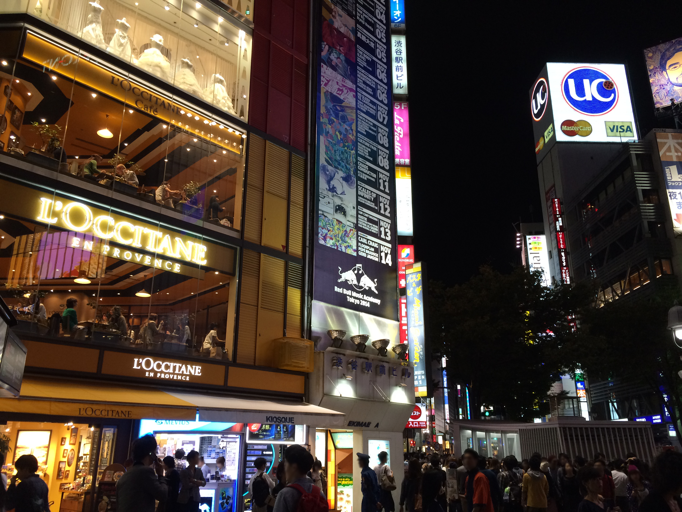}} \,
    \subfloat{\includegraphics[width=\examplesize\linewidth]{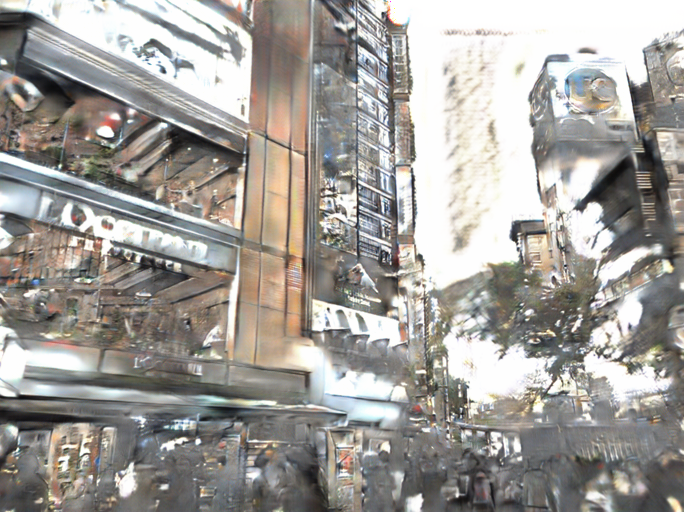}} \,
    \subfloat{\includegraphics[width=\examplesize\linewidth]{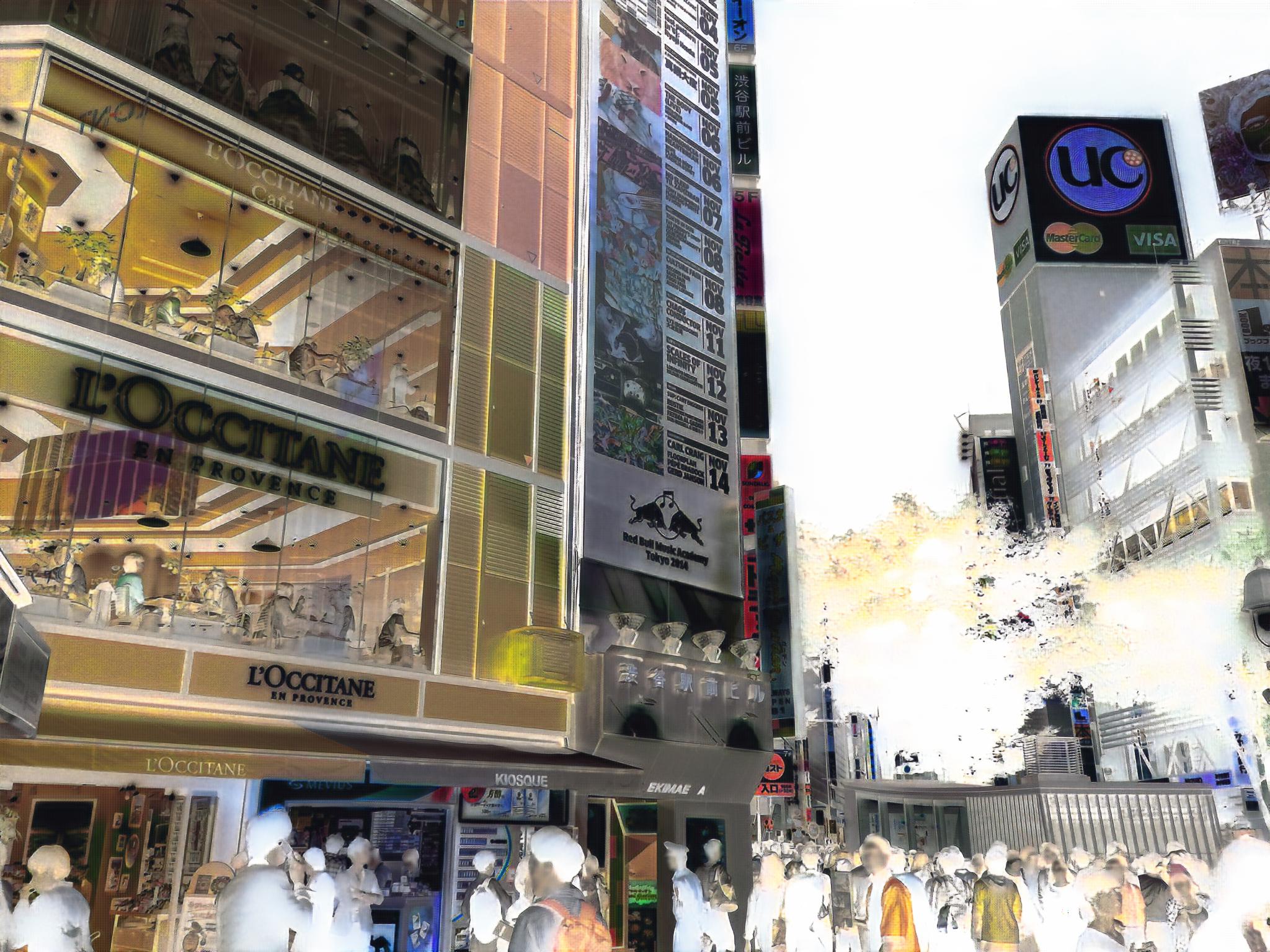}} \,
    \\[0.5em]
    \subfloat{\includegraphics[width=\examplesize\linewidth]{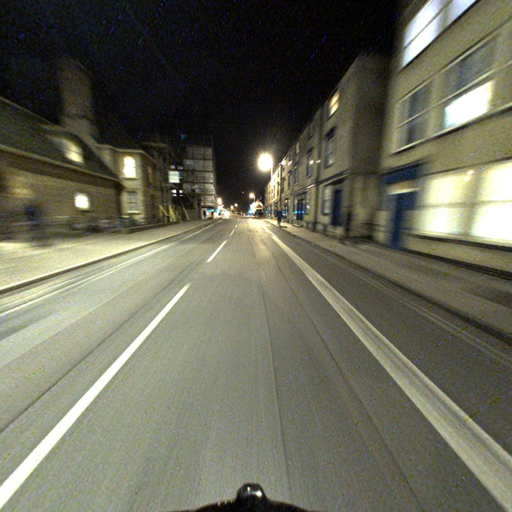}} \,
    \subfloat{\includegraphics[width=\examplesize\linewidth]{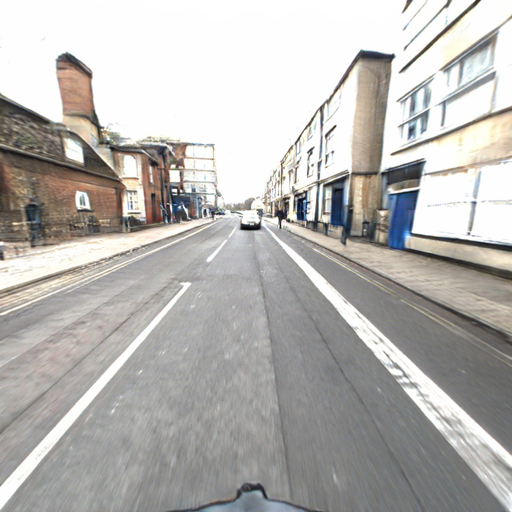}} \,
    \subfloat{\includegraphics[width=\examplesize\linewidth]{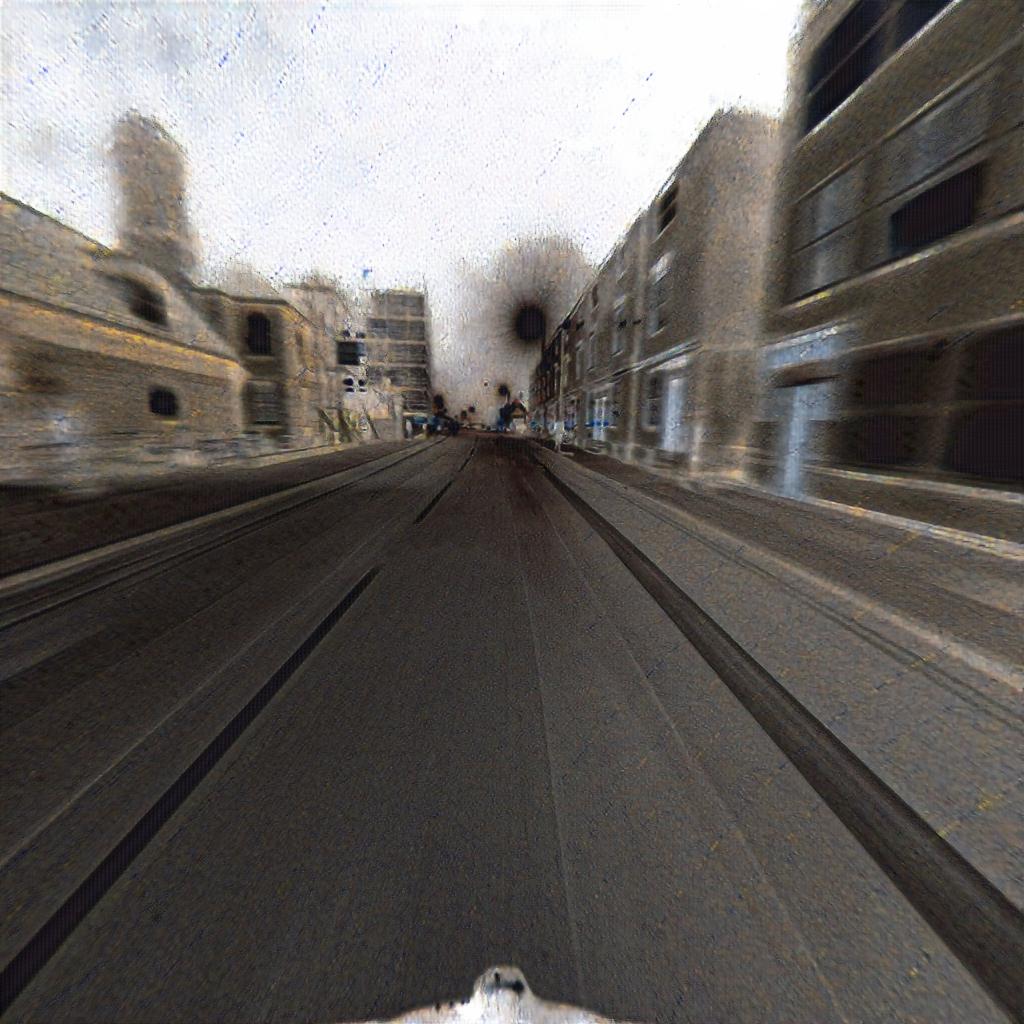}} \,
    \\[1.5em]
    (continues)
\end{figure*}

\begin{figure*}
\ContinuedFloat 
    \newcommand{\examplesize}{.238}
    \centering
    \begin{tabularx}{\textwidth}{YYYY}
    Night original \, & CyEDA BDD & CyEDA (tuned) & DRIT
    \end{tabularx}
    \\[0.5em]
    \subfloat{\includegraphics[width=\examplesize\linewidth]{img/night2day/sfmnd_8a4faac169e25011fac16bd352980d90_orig.png}} \,
    \subfloat{\includegraphics[width=\examplesize\linewidth]{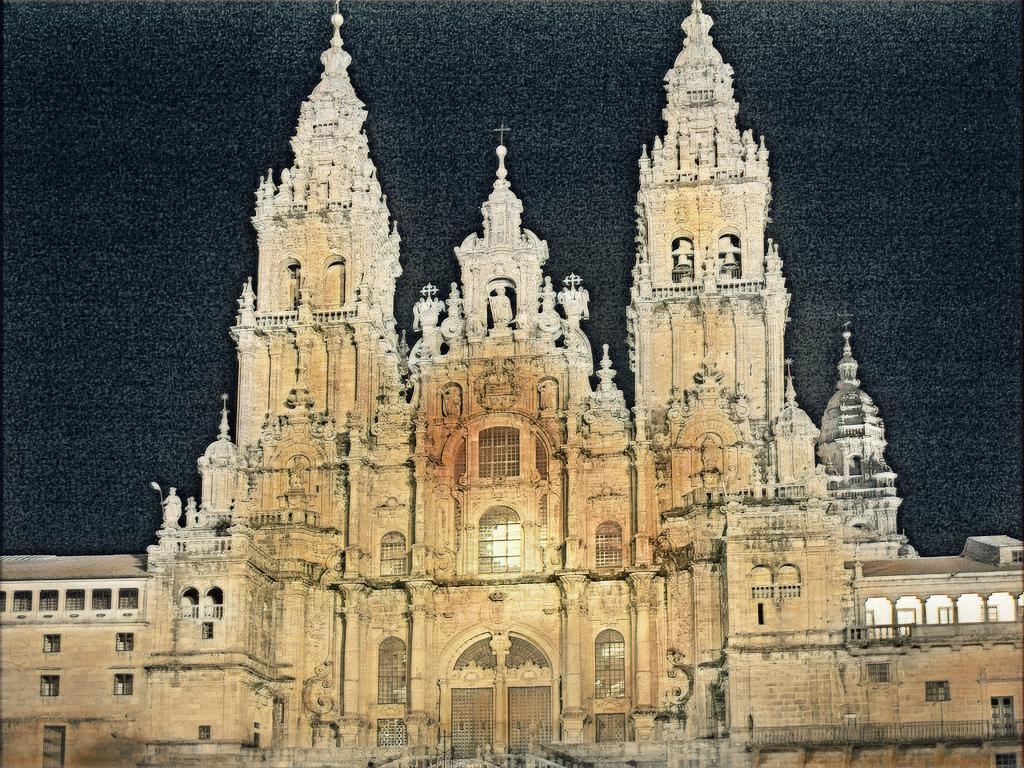}} \,
    \subfloat{\includegraphics[width=\examplesize\linewidth]{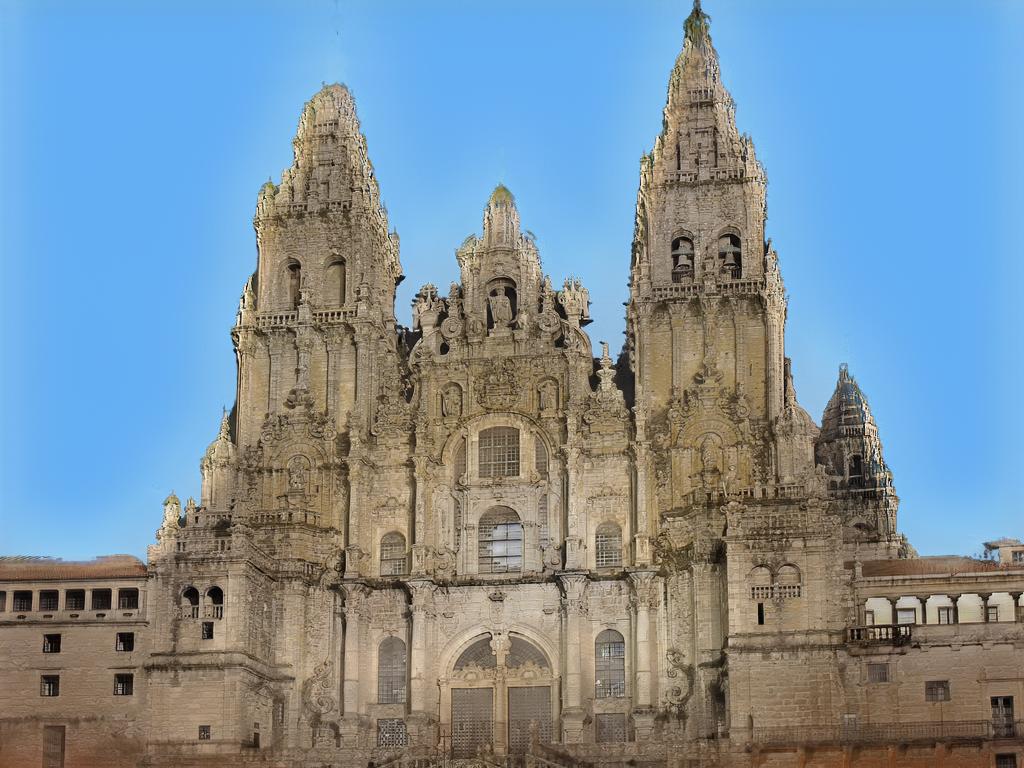}} \,
    \subfloat{\includegraphics[width=\examplesize\linewidth]{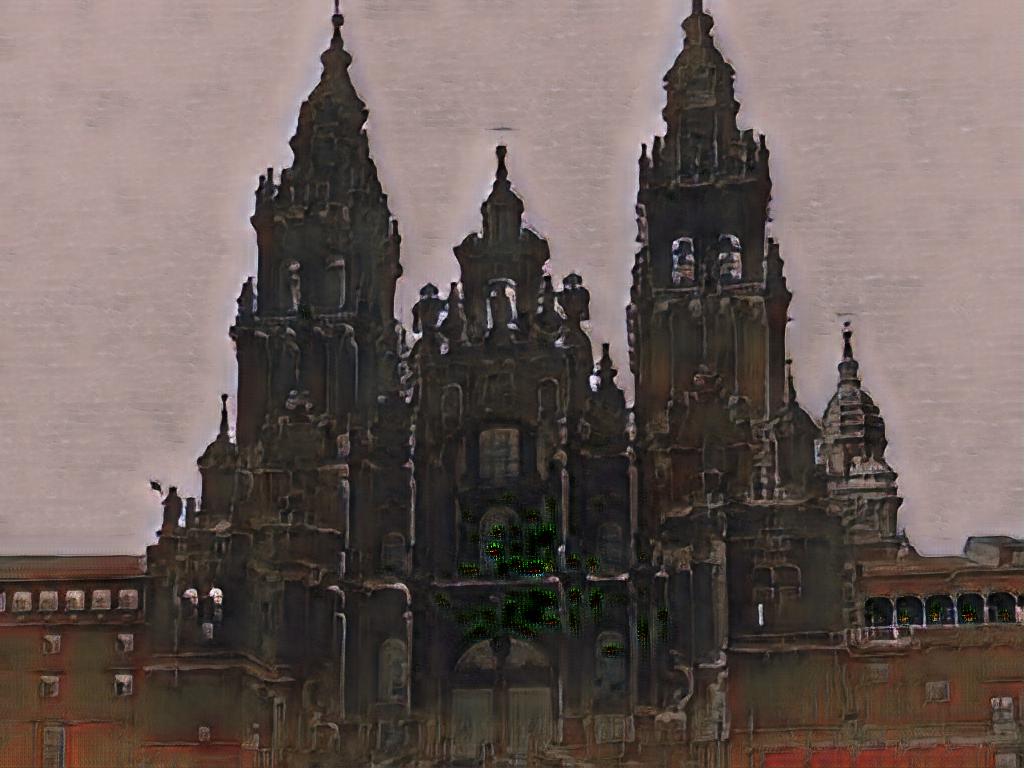}} \,
    \\[0.5em]
    \subfloat{\includegraphics[width=\examplesize\linewidth]{img/night2day/aachen_IMG_20161227_172439_orig.png}} \,
    \subfloat{\includegraphics[width=\examplesize\linewidth]{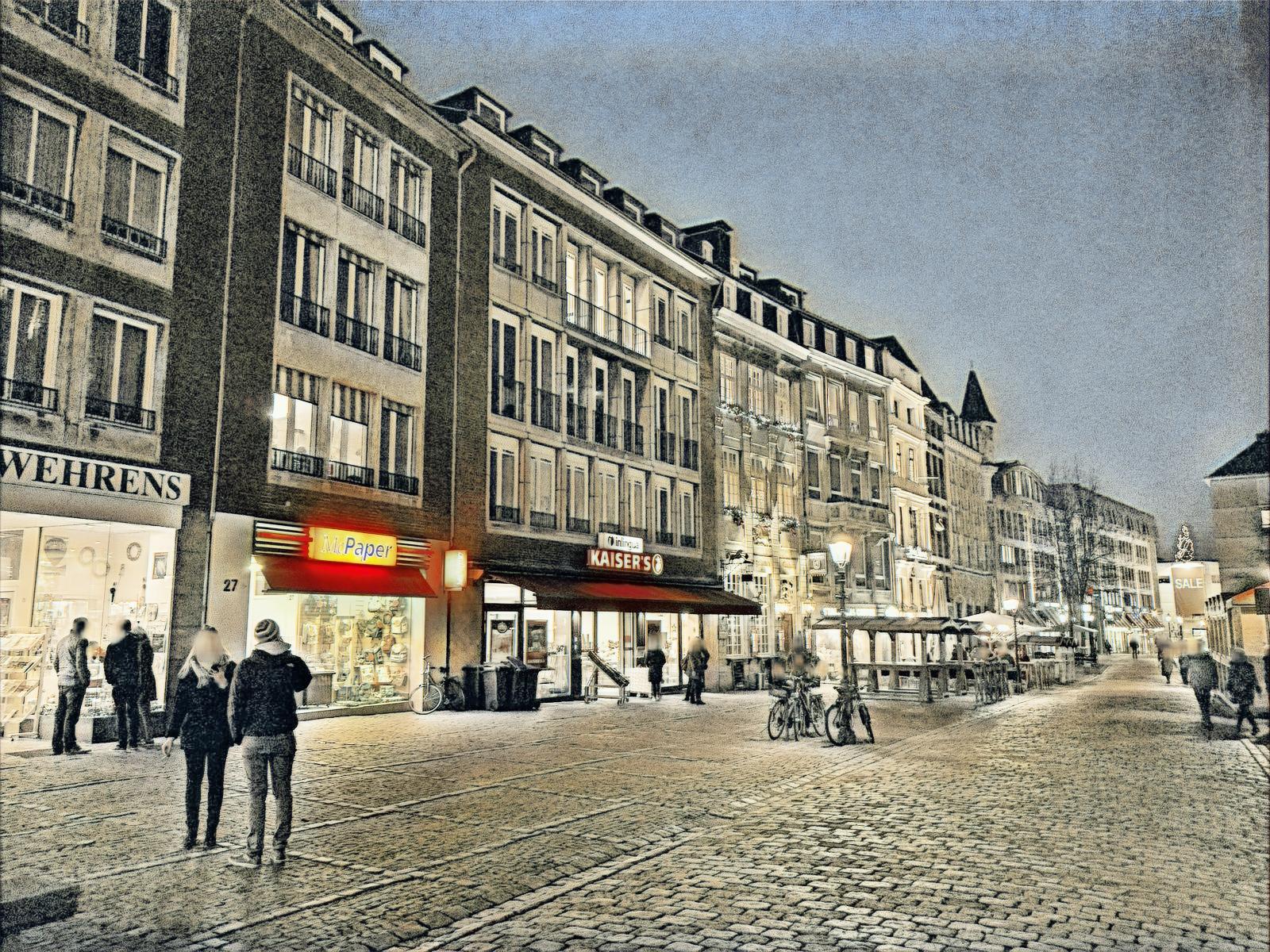}} \,
    \subfloat{\includegraphics[width=\examplesize\linewidth]{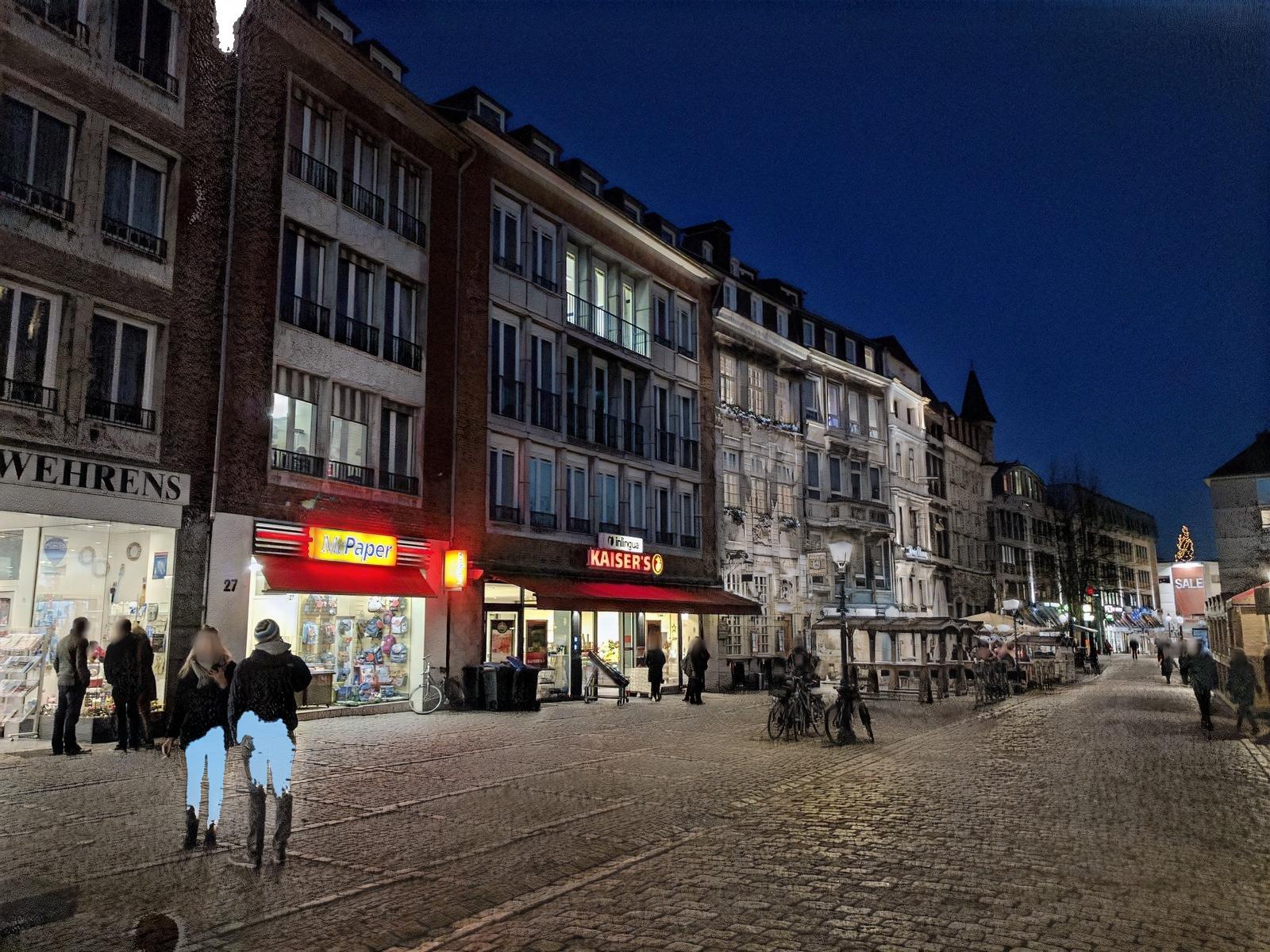}} \,
    \subfloat{\includegraphics[width=\examplesize\linewidth]{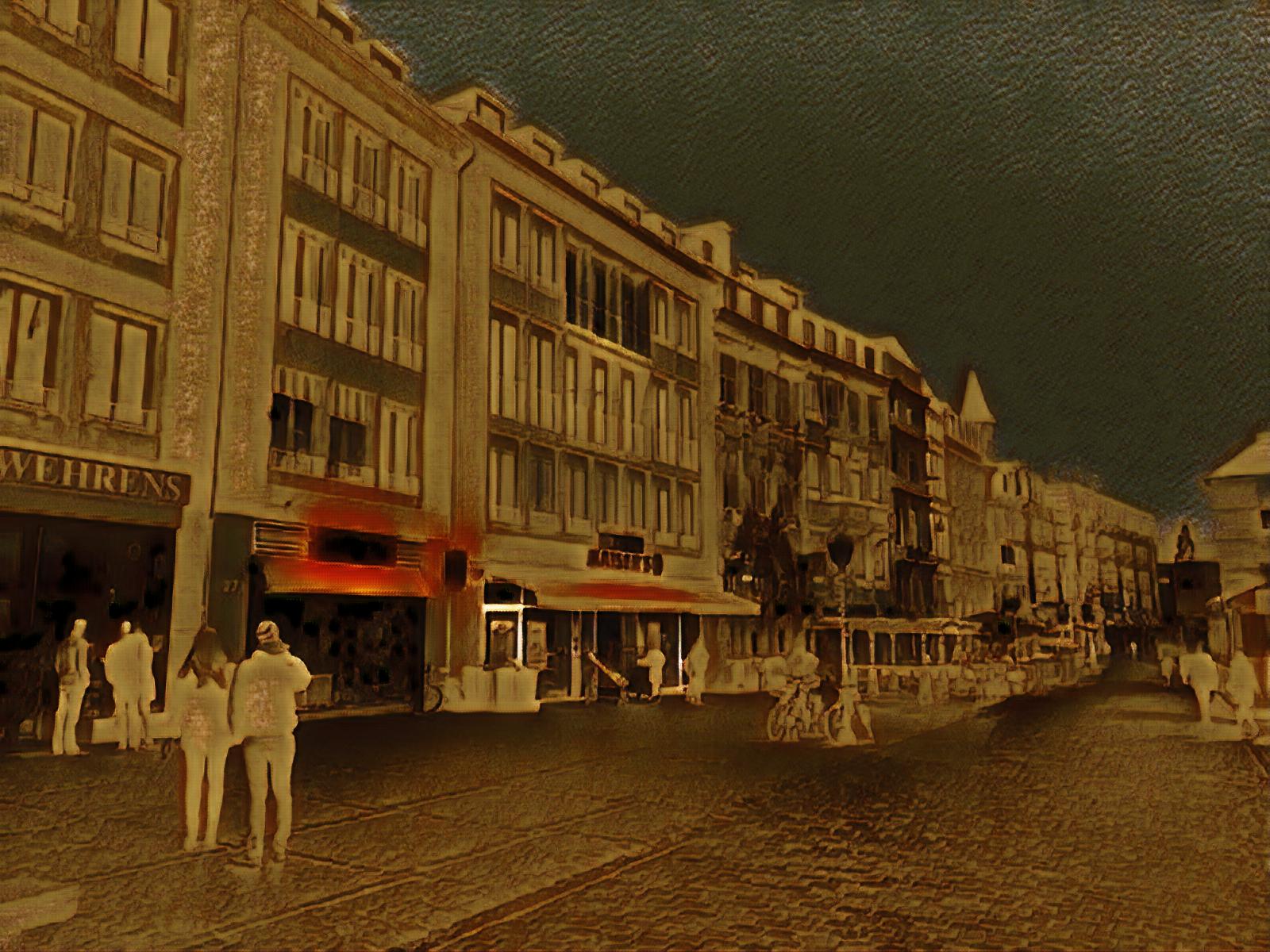}} \,
    \\[0.5em]
    \subfloat{\includegraphics[width=\examplesize\linewidth]{img/night2day/tokio_00009_orig.png}} \,
    \subfloat{\includegraphics[width=\examplesize\linewidth]{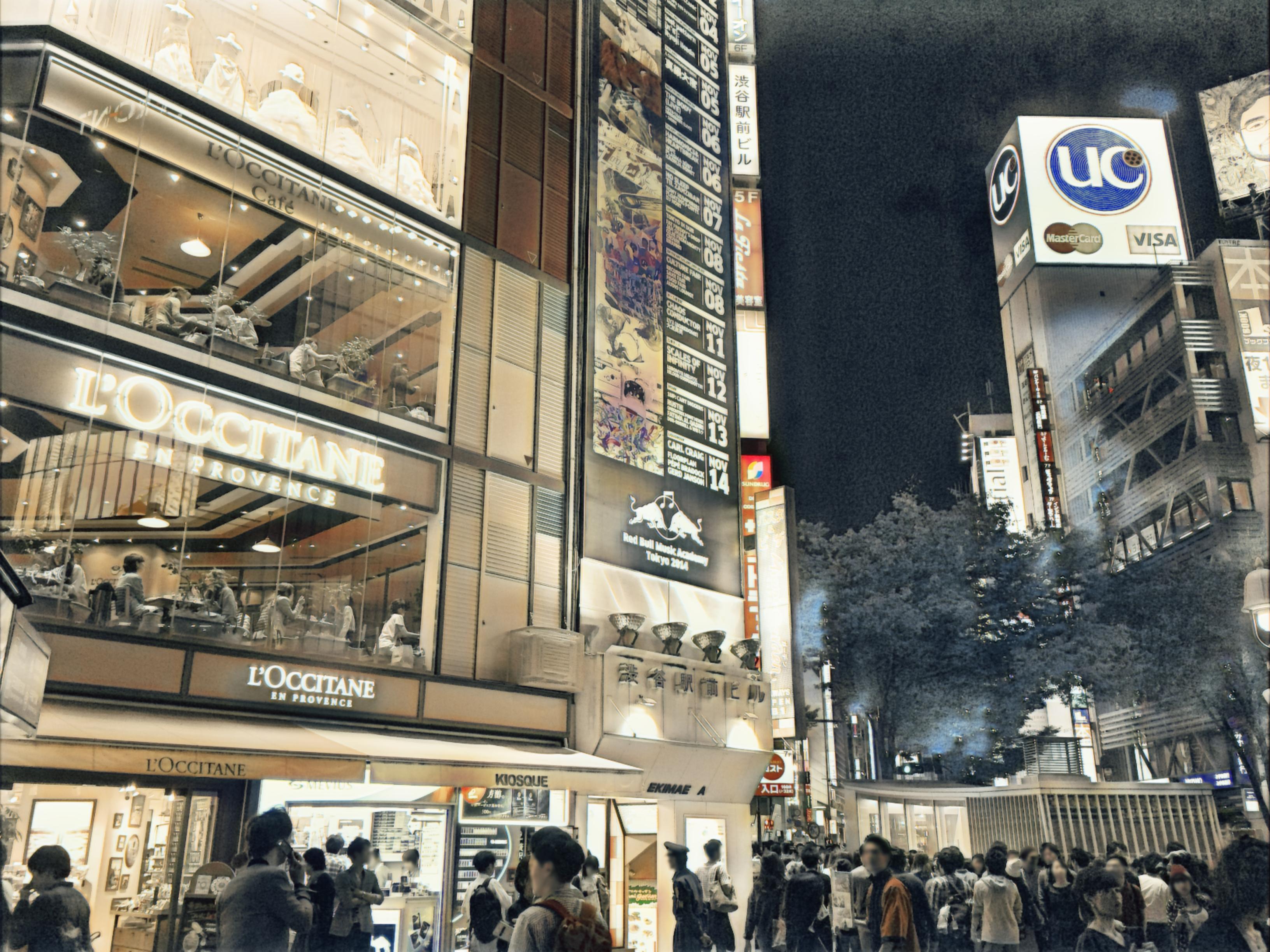}} \,
    \subfloat{\includegraphics[width=\examplesize\linewidth]{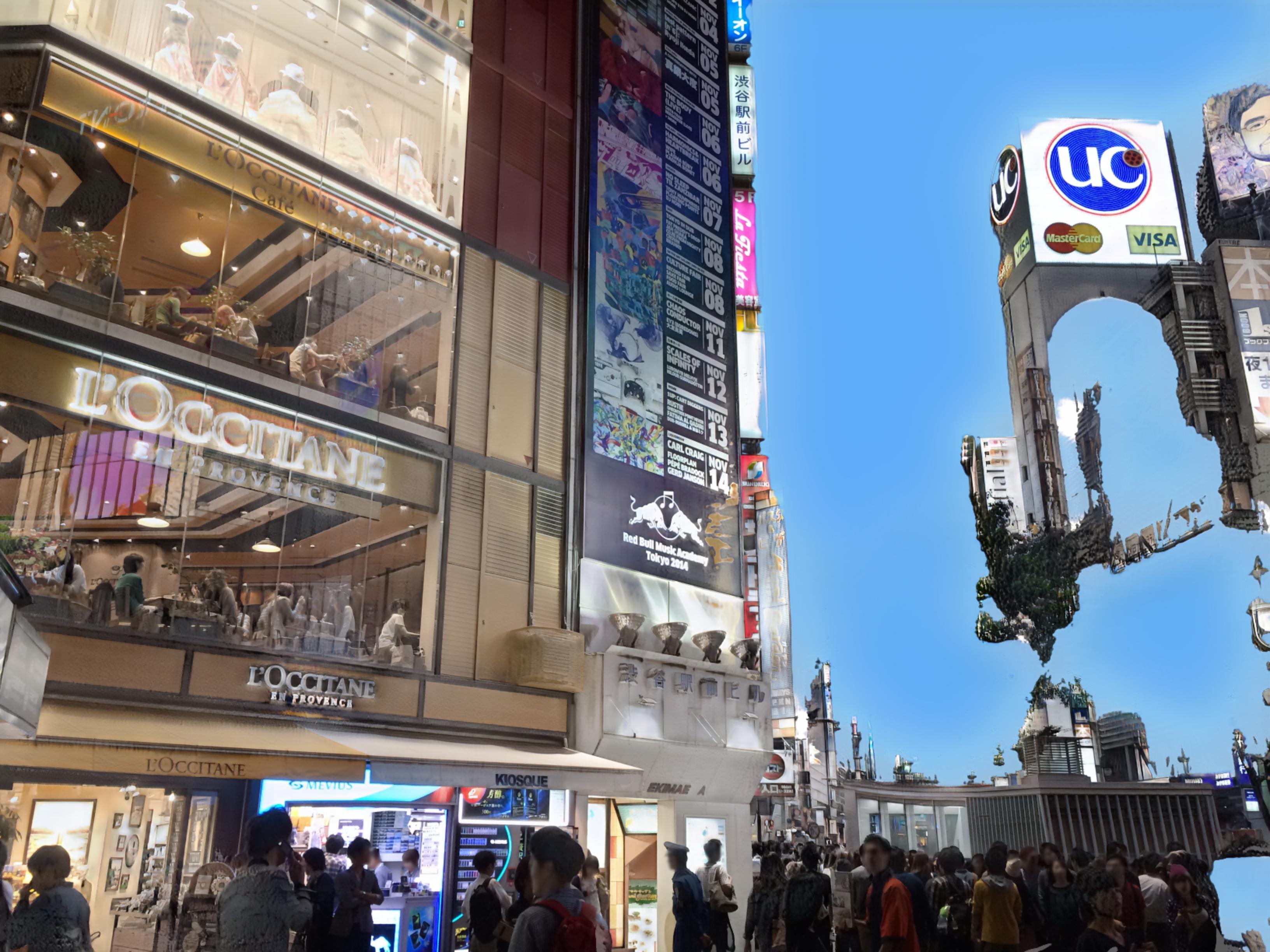}} \,
    \subfloat{\includegraphics[width=\examplesize\linewidth]{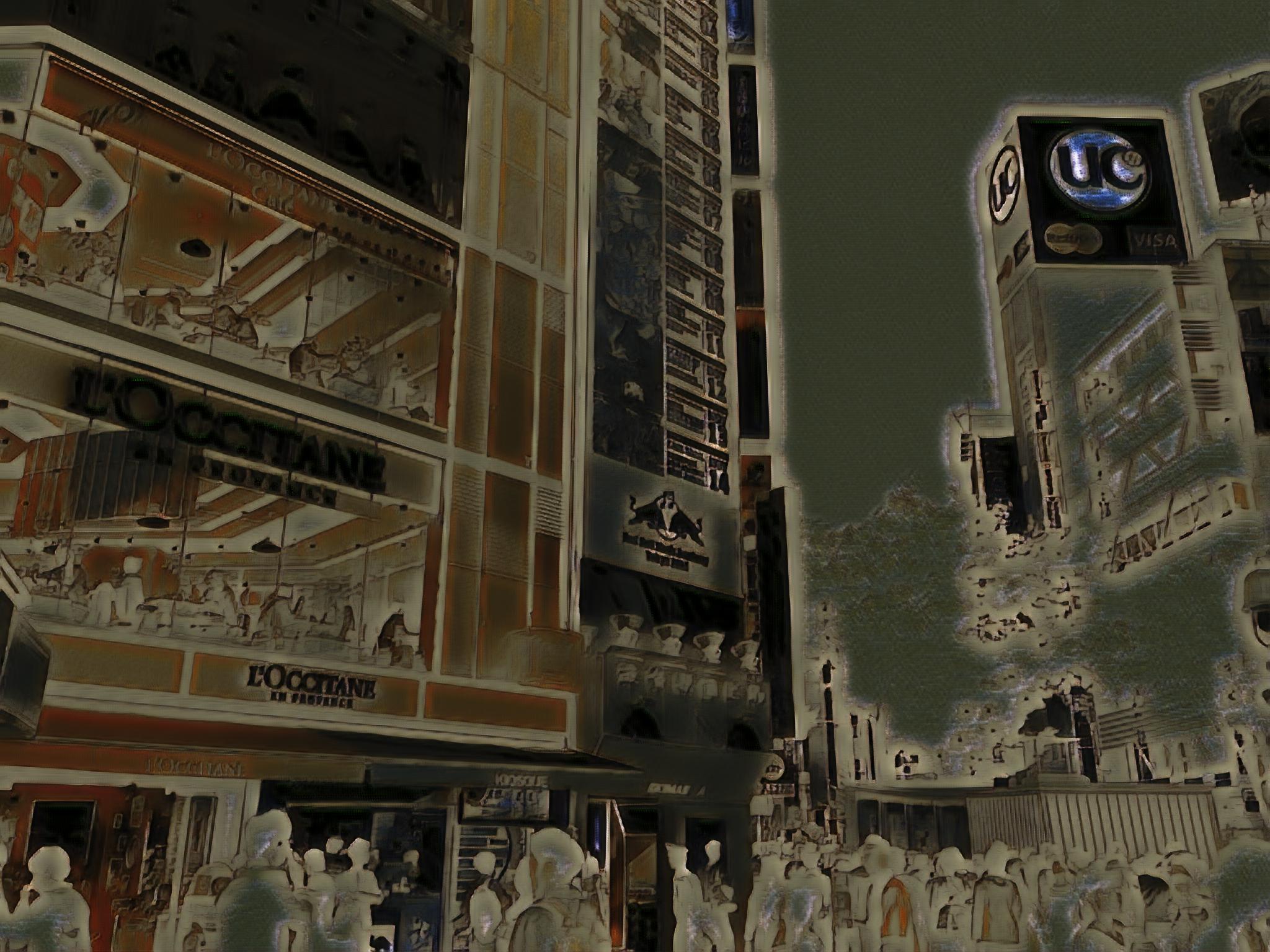}} \,
    \\[0.5em]
    \subfloat{\includegraphics[width=\examplesize\linewidth]{img/night2day/robotcar_1418235223450115.rear_orig.png}} \,
    \subfloat{\includegraphics[width=\examplesize\linewidth]{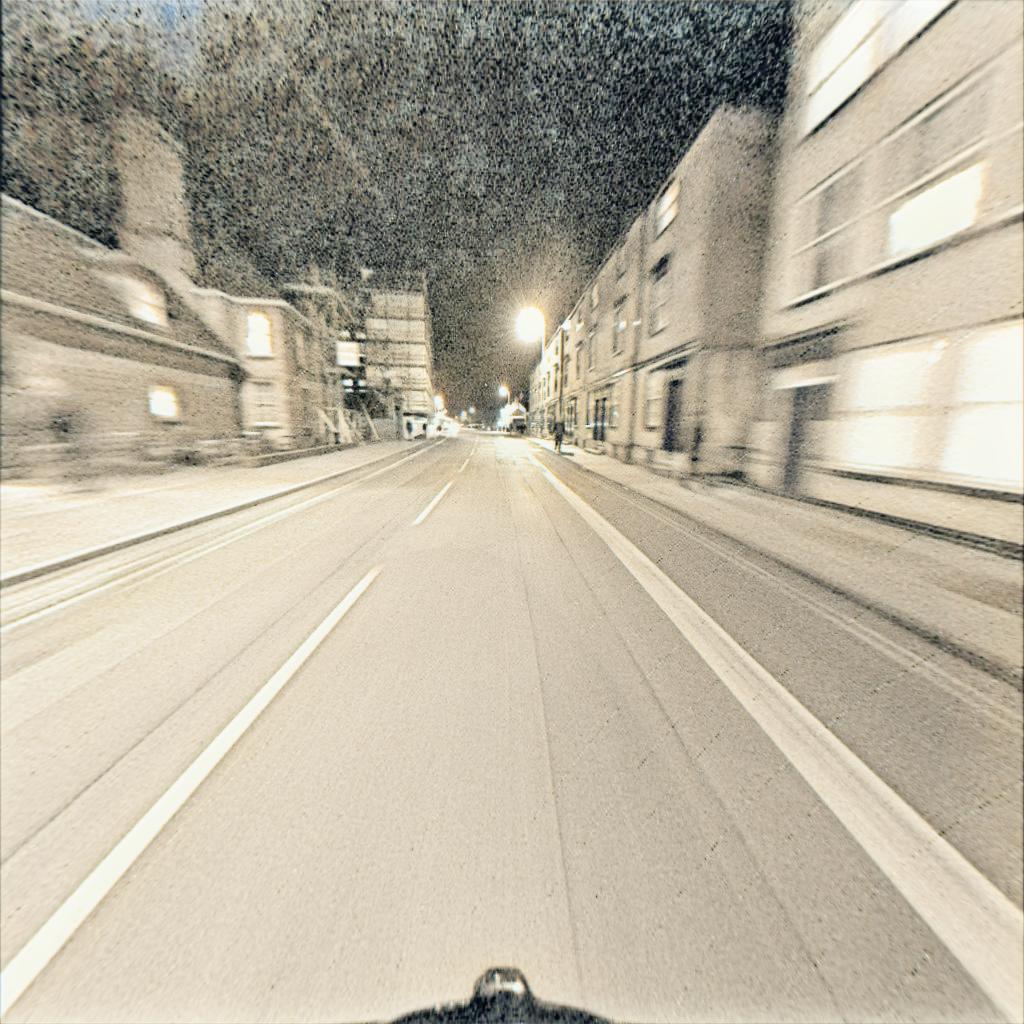}} \,
    \subfloat{\includegraphics[width=\examplesize\linewidth]{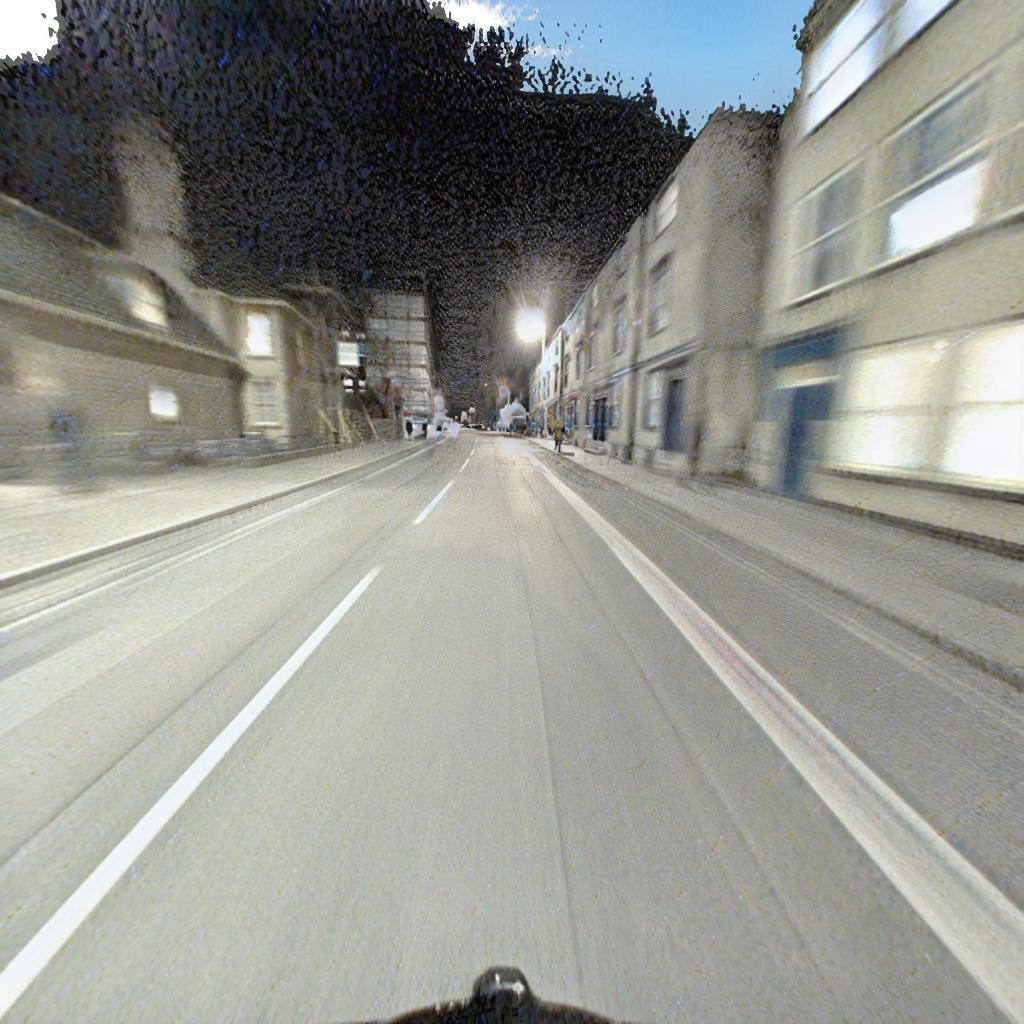}} \,
    \subfloat{\includegraphics[width=\examplesize\linewidth]{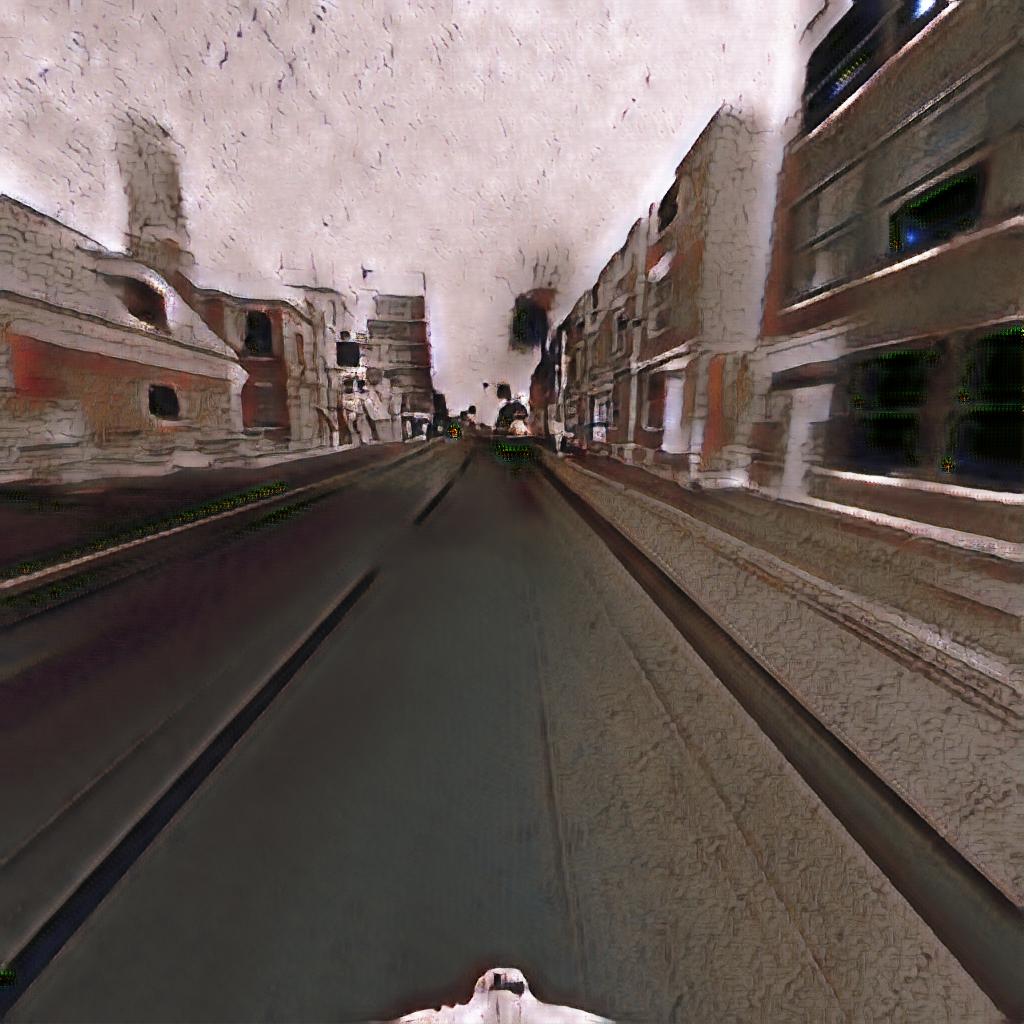}} \,
    \caption{Examples of night images translated into the day domain by different night $\rightarrow$ day generators. The columns correspond to (left-to-right): the original image, ToDayGAN generator~\cite{toDayGAN}, our CycleGAN generator, the original image, CyEDA~\cite{beh2022cyeda} generator trained on BDD dataset~\cite{yu2020bdd100k}, CyEDA generator trained on \textit{SfM}120k~\cite{radenovic2018fine} and tuned by us, and our DRIT generator. The rows show example images from datasets (top-to-bottom) \textit{SfM-N/D}~\cite{jenicek}, \textit{Aachen}~\cite{Sattler2018CVPR,Sattler2012BMVC}, \textit{Tokyo}~\cite{Torii-CVPR2015}, and \textit{RobotCar}~\cite{RobotCarDatasetIJRR}, respectively.}
    \label{fig:night_examples}
\end{figure*}

\clearpage

\section{Appendix C}

Examples of the evolution of light condition invariance.

In Figure~\ref{fig:minedat}, a training data tuple is visualized. The negatives are images of a different landmark that are the most similar to the translated anchor. Negatives are mined (re-computed) in the beginning of each epoch, so as the network trains, the mined negatives change - mined negatives shift from appearance-similar dark images to content-similar images of any domain.

In Figure~\ref{fig:mining}, in each example, there are three images from the \textit{SfM-N/D}~\cite{jenicek} validation set: night anchor, day positive, and mined negative. The distance between the anchor and positive as well as the anchor and negative is plotted - increasing the gap between these two distances translates to easier distinction of the day positive from the night negative.

In epoch 0, the embedding network starts from ImageNet weights~\cite{russakovsky2015imagenet} and has never seen a night image. Therefore the two dark images are deemed similar, measured in L2 distance of their embeddings, while the images of the same objects under different light conditions are less similar. Later in the training with \hedn GAN augmentation, the distance between the images starts reflecting actual visual similarity rather than the similarity of image illumination.

\renewcommand{\thefigure}{C\arabic{figure}}
\setcounter{figure}{1}
\newcommand{\minedsize}{1.0}
\begin{figure*}[t]
    \centering
    \vspace{-5em}
    \includegraphics[width=\minedsize\linewidth]{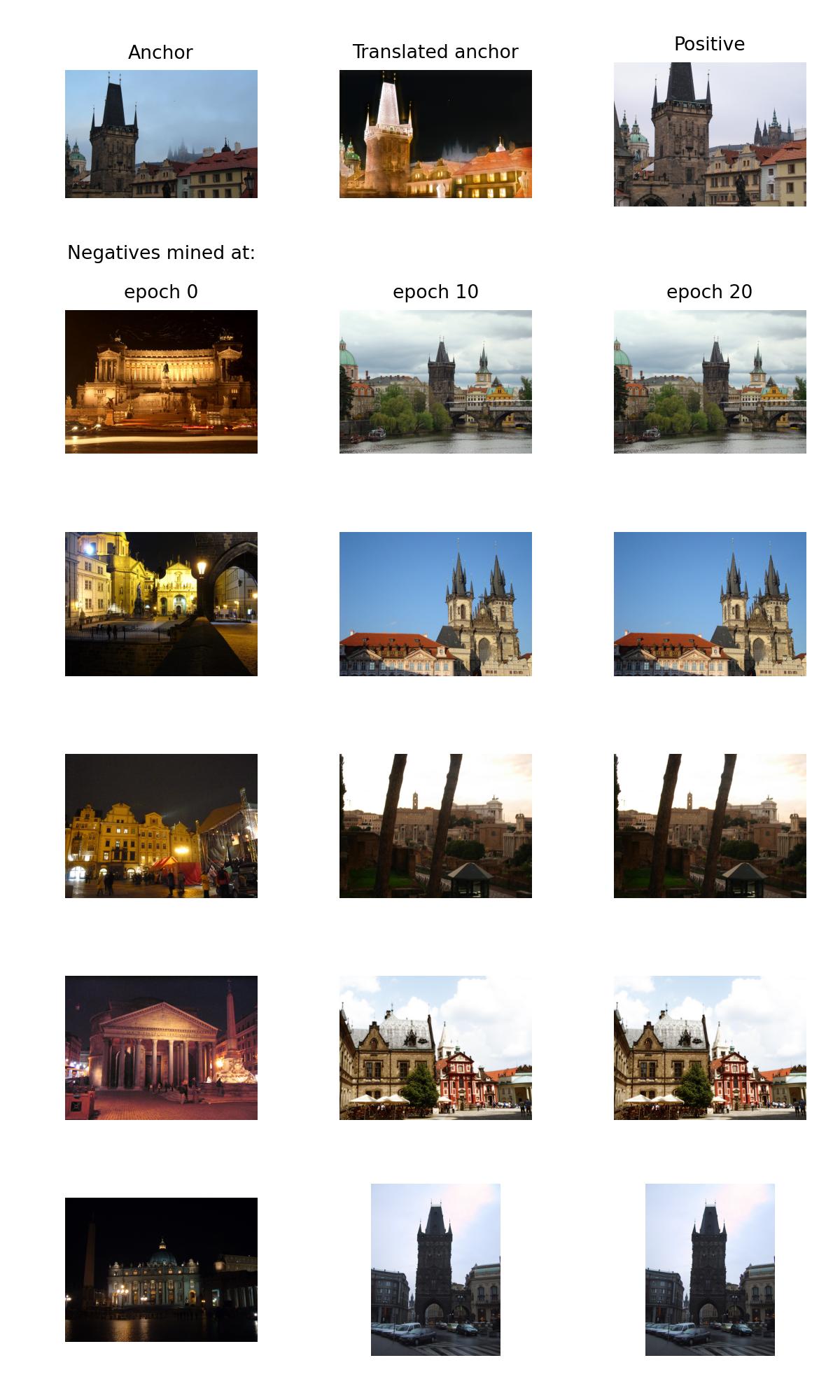}
    \\[1.5em]
    (continues)
\end{figure*}

\begin{figure*}[t]
\ContinuedFloat 
    \centering
    \vspace{-5em}
    \includegraphics[width=\minedsize\linewidth]{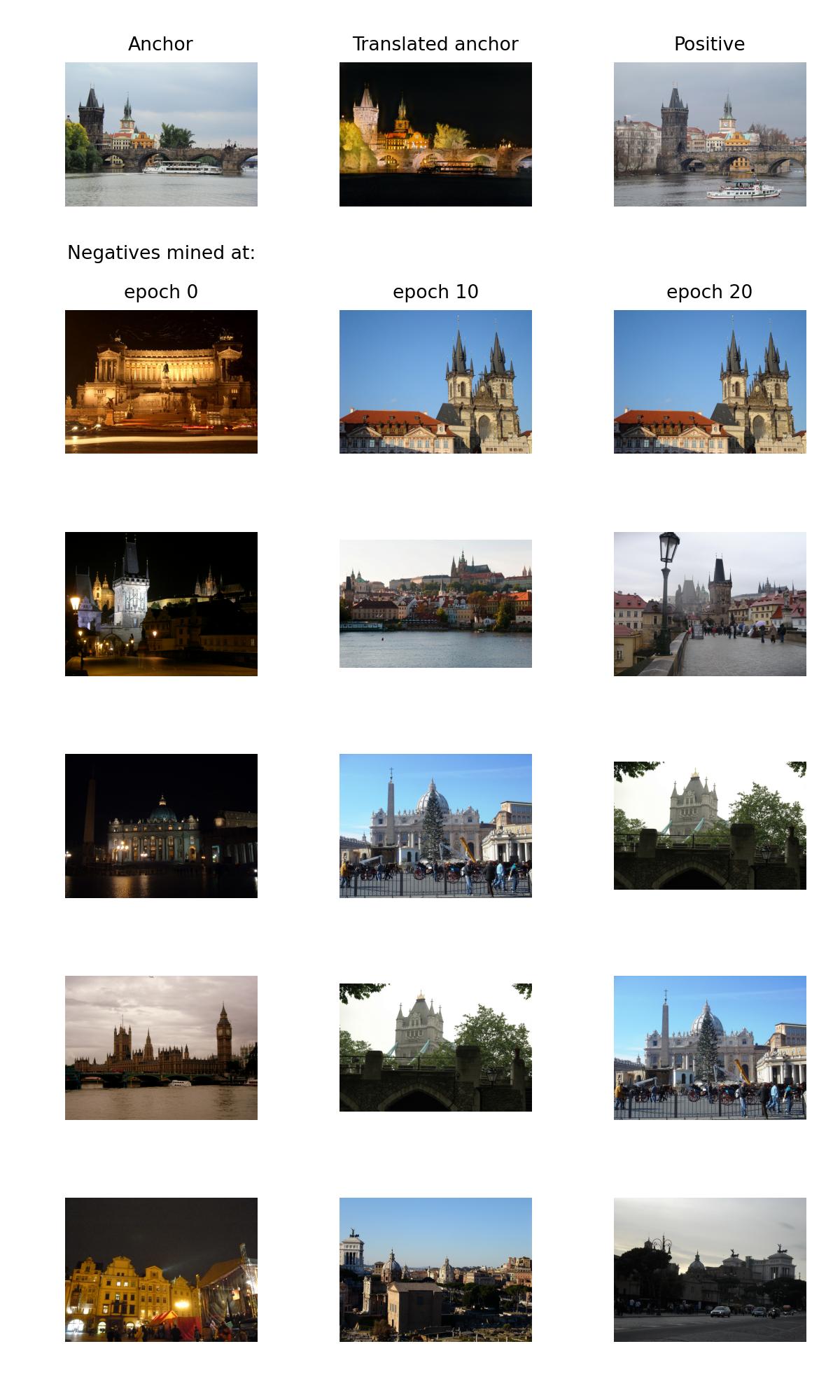}
    \caption{The visualization of hard-negative mining during the training of our method. The anchor image is translated with our \hedn GAN generator into the night domain. Five hardest negative examples are mined for each translated anchor at the beginning of each epoch. We provide mined examples at epoch 0, 10, and 20 for two different anchors. Notice the shift from negatives that are similar to the translated anchor in appearance to negatvies that are similar in content.}
   \label{fig:minedat}
\end{figure*}
\setcounter{figure}{1}
\newcommand{\miningplotsize}{0.33}
\newcommand{\miningsize}{0.18}
\begin{figure*}[t]
    \centering
    \begin{tabularx}{0.8\textwidth}{YYY}
    \hspace{35pt} Anchor & \hspace{35pt} Positive & Negative
    \end{tabularx}
    \subfloat{{\includegraphics[height=\miningsize\linewidth]{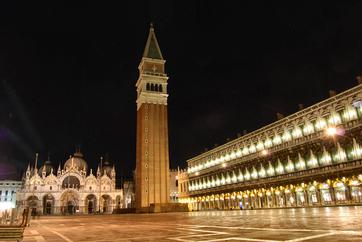}}} \,
    \subfloat{{\includegraphics[height=\miningsize\linewidth]{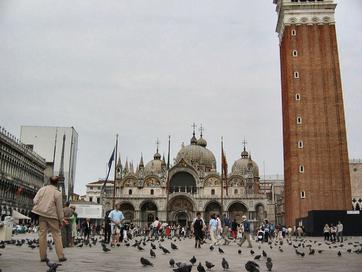}}} \,
    \subfloat{{\includegraphics[height=\miningsize\linewidth]{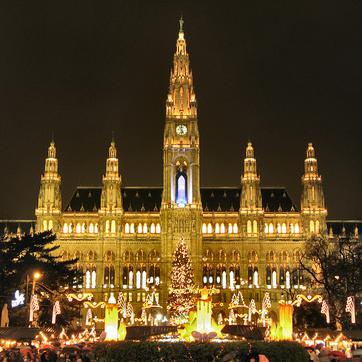}}} \\
    \subfloat{{\includegraphics[height=\miningplotsize\linewidth]{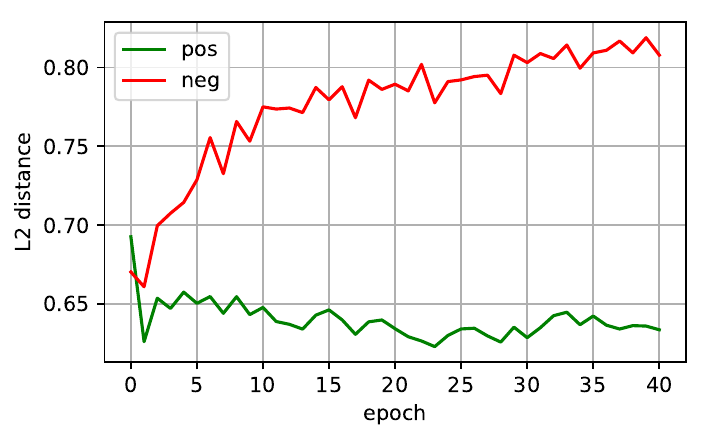}}}\\
    \vspace{0.5em}
    \begin{tabularx}{0.8\textwidth}{YYY}
    Anchor & \hspace{10pt} Positive & \hspace{10pt} Negative
    \end{tabularx}
    \subfloat{{\includegraphics[height=\miningsize\linewidth]{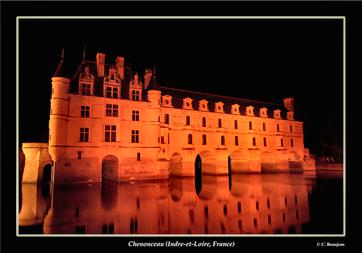}}} \,
    \subfloat{{\includegraphics[height=\miningsize\linewidth]{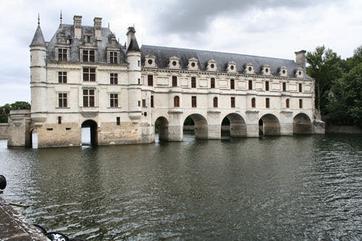}}} \,
    \subfloat{{\includegraphics[height=\miningsize\linewidth]{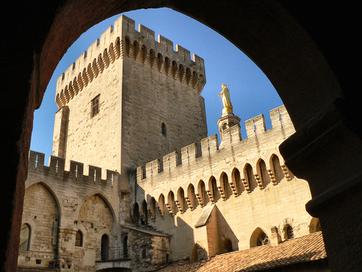}}} \\
    \subfloat{{\includegraphics[height=\miningplotsize\linewidth]{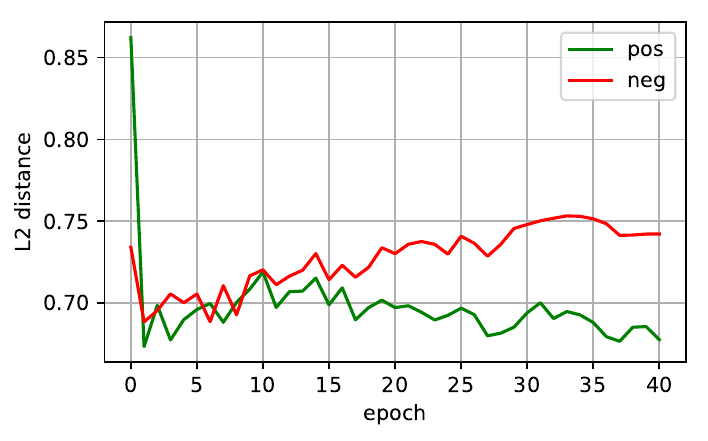}}}\\
    \caption{Two examples of image triplets: night anchor, day positive and mined negative (from left to right). Under each triplet, the L2 distance of anchor-positive (green) and anchor-negative (red) is plotted against the current training epoch. Notice the expanding gap between these two distances as the training progresses.}
   \label{fig:mining}
\end{figure*}

\clearpage

{\small
\bibliographystyle{ieee_fullname}
\bibliography{egbib_supp}
}